\documentclass[lettersize,journal]{IEEEtran}
\usepackage{amsmath,amsfonts}
\usepackage{algorithmic}
\usepackage{algorithm}
\usepackage{setspace}

\usepackage{array}
\usepackage{textcomp}
\usepackage{stfloats}
\usepackage{url}
\usepackage{verbatim}
\usepackage{graphicx}
\usepackage{cite}
\usepackage{color}
\usepackage{bm}
\usepackage{subfloat}
\usepackage{bbding}
\usepackage{makecell}                 
\usepackage{booktabs}                 
\usepackage{tabu}                     
\usepackage{multirow}                 
\usepackage{multicol}                 
\usepackage{multirow}                 
\usepackage[table]{xcolor}  
\usepackage{colortbl}
\usepackage[export]{adjustbox}
\usepackage{arydshln}
\usepackage[inline]{enumitem}
\usepackage{overpic}
\usepackage{pifont}
\usepackage{hyperref}
\usepackage[caption=false,font=small]{subfig}

\begin{document}
\title{ForensicsSAM: Toward Robust and Unified Image Forgery Detection and Localization Resisting to Adversarial Attack}

\author{
	Rongxuan~Peng,~\IEEEmembership{Student Member,~IEEE,}
	Shunquan~Tan,~\IEEEmembership{Senior Member,~IEEE,}
	Chenqi Kong,~\IEEEmembership{Member,~IEEE,}
	Anwei Luo,
	Alex C. Kot,~\IEEEmembership{Life Fellow,~IEEE},
	and~Jiwu~Huang,~\IEEEmembership{Fellow,~IEEE}%
    \thanks{Rongxuan Peng is with the Shenzhen Key Laboratory of Media Security, Faculty of Electronic and Information Engineering, Shenzhen University, China; and the Rapid-Rich Object Search (ROSE) Lab, School of Electrical and Electronic Engineering, Nanyang Technological University, Singapore. Email: 2350433004@email.szu.edu.cn.}%
    
    \thanks{Chenqi Kong, and Anwei Luo are with the Rapid-Rich Object Search Lab, School of Electrical and Electronic Engineering, Nanyang Technological University, Singapore.}%
    
    \thanks{Shunquan Tan, Alex C. Kot, and Jiwu Huang are with the Guangdong Laboratory of Machine Perception and Intelligent Computing, Faculty of Engineering, Shenzhen MSU-BIT University, China.}%
 
    \thanks{Corresponding author: Shunquan Tan. E-mail: tansq@smbu.edu.cn.}%
}

\maketitle
\begin{abstract}
Parameter-efficient fine-tuning (PEFT) has emerged as a popular strategy for adapting large vision foundation models—such as the Segment Anything Model (SAM) and LLaVA—to downstream tasks like image forgery detection and localization (IFDL). However, existing PEFT-based approaches overlook their vulnerability to adversarial attacks. In this paper, we show that highly transferable adversarial images can be crafted solely via the upstream model, without accessing the downstream model or training data, significantly degrading the IFDL performance.
To address this, we propose ForensicsSAM, a unified IFDL framework with built-in adversarial robustness. Our design is guided by three key ideas: 
(1) To compensate for the lack of forgery-relevant knowledge in the frozen image encoder, we inject forgery experts into each transformer block to enhance its ability to capture forgery artifacts. These forgery experts are always activated and shared across any input images.
(2) To detect adversarial images, we design a light-weight adversary detector that learns to capture structured, task-specific artifact in RGB domain, enabling reliable discrimination across various attack methods.
(3) To resist adversarial attacks, we inject adversary experts into the global attention layers and MLP modules to progressively correct feature shifts induced by adversarial noise. These adversary experts are adaptively activated by the adversary detector, thereby avoiding unnecessary interference with clean images.
Extensive experiments across multiple benchmarks demonstrate that ForensicsSAM achieves superior resistance to various adversarial attack methods, while also delivering state-of-the-art performance in image-level forgery detection and pixel-level forgery localization. The resource is available at \url{https://github.com/siriusPRX/ForensicsSAM}.
\end{abstract}
	

\section{Introduction}
\label{section:introduction}
\IEEEPARstart{I}{mage} forgery detection and localization (IFDL)~\cite{kadha2025unravelling} aims to identify whether an image has been tampered with and accurately localize the forged regions. It plays a critical role in digital forensics and media authenticity verification.
\begin{figure}[t]
	\centering 
	\begin{overpic}[width=0.5\textwidth]{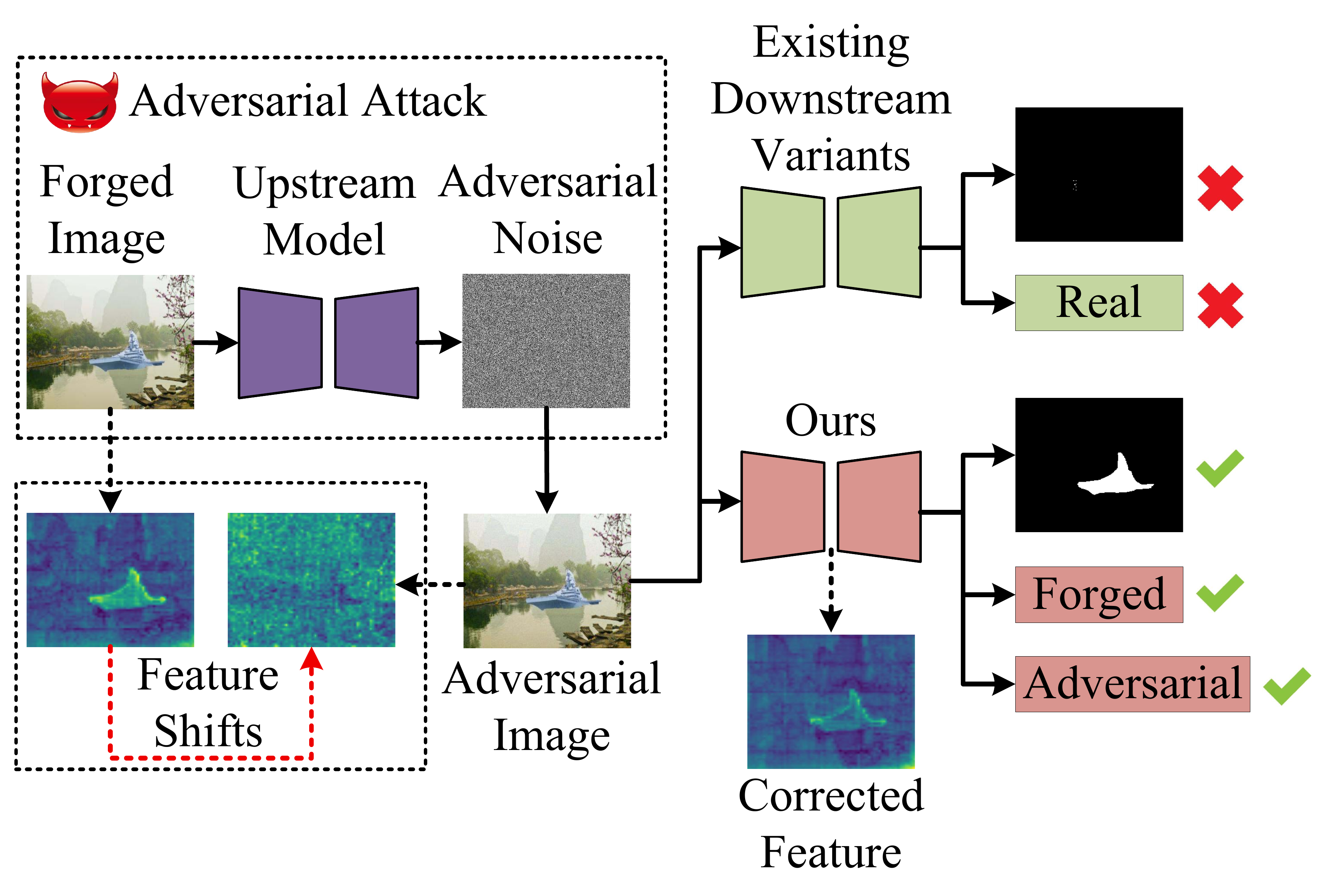} 		
	\end{overpic}
	\caption{Overview of adversarial attack and comparison of adversarial robustness.
		The attacker leverages the upstream model to generate transferable adversarial noise, which introduces feature shifts in the forged image and degrades the performance of downstream variants. In contrast, our method accurately identifies the adversarial image and enables robust image forgery detection and localization under adversarial attacks.}
	\label{fig:intro}
\end{figure}
Recently, deep learning has emerged as the dominant paradigm for IFDL task. Numerous methods~\cite{hu2020span, liu2022pscc, Wang_2022_CVPR, dong2022mvss, kong2022detect, kwon2022learning, wu2022robust, guillaro2023trufor, kong2024moe, peng2024employing, kong2025pixel} have been proposed based on this paradigm, leveraging CNN or transformer architectures to effectively learn to capture subtle forgery traces from images. Building upon this foundation, recent efforts~\cite{su2024novel, xufakeshield, huang2025sida, kwon2025safire} have turned to leveraging large-scale vision foundation models, such as SAM (Segment Anything Model)\cite{kirillov2023segment} and LLaVA\cite{liu2023visual}, for the IFDL task. In addition to transferring rich visual priors from general-purpose models, these approaches also incorporate textual information to assist forgery detection and localization, leading to improved generalization and interpretability. These methods mainly adopt PEFT (parameter-efficient fine-tuning) techniques—such as LoRA~\cite{hu2022lora} or adapter—which update only a small subset of parameters to fine-tune the large-scale vision foundation models for IFDL.

However, recent studies~\cite{peng2025active} have pointed out that existing IFDL methods remain highly vulnerable to adversarial attacks, posing serious risks in security-sensitive applications. Motivated by PATA++~\cite{zheng2024black} and UMI-GRAT~\cite{xiatransferable}, in our work, we further reveal that IFDL methods based on large-scale vision foundation models also suffer from such vulnerability: adversarial images crafted solely via the upstream model—without any access to their downstream model parameters or training data—can significantly degrade their forgery detection and localization performance.

In this paper, we propose ForensicsSAM, a novel and unified IFDL framework with built-in adversarial robustness, as illustrated in Fig.~\ref{fig:intro}. To achieve this goal, we identify two core challenges:
(1) Existing models is hard to effectively capture universal forgery-relevant knowledge that remains stable across real, forged, and adversarial images;
(2) Adversarial noise crafted from the upstream model can induce significant feature shifts in the downstream model, leading to incorrect predictions. 

Building upon these challenges, we correspondingly develop our solution. 
First, in order to fully utilize the rich image semantic knowledge of SAM and compensate for its lack of forgery-relevant knowledge, we inject forgery experts into each transformer block of its image encoder. These forgery experts are always activated and shared across any input images. Second, for potential adversarial attacks, we provide multi-granularity resistance. At the image level, unlike random noise, adversarial noise introduces structured, task-specific artifacts in the RGB domain that deviate from natural image statistics. To leverage this, we introduce a light-weight adversary detector that learns to identify such pattern and outputs an adversary score. At the feature level, we further inject adversary experts into the global attention layers and MLP (Multi-Layer Perceptron) modules of SAM's image encoder. These experts are adaptively activated based on the adversary score and are designed to progressively correct feature shifts. This design ensures that corrupted features are restored while avoiding unnecessary interference with clean images. Finally, we introduce a forgery detector and a mask decoder to use the robust feature generated from our elaborately designed image encoder to perform image-level forgery detection and pixel-level forgery localization, respectively.

Our primary contributions can be summarized as follows:
\begin{itemize} 
	\item We propose a robust and unified IFDL framework, ForensicsSAM, which can simultaneously perform image-level detection for real, forged, and adversarial images, while also achieving pixel-level forgery localization for both forged and adversarial images.
	\item We propose a three-stage training strategy with dedicated objectives for clean and adversarial images, respectively. This design effectively decouples the learning process of ForensicsSAM on clean and adversarial images, allowing the two objectives to be optimized independently yet collaboratively.
	\item Comprehensive experimental results demonstrate that the proposed ForensicsSAM has superior adversarial robustness under various adversarial attack methods, and achieves the state-of-the-art IFDL performance compared to existing methods.
\end{itemize}

The remainder of this paper is structured as follows. Section~\ref{sect:related_work} presents the related works about image forgery detection and localization, and adversarial attack on PEFT-based SAM variants. Section~\ref{sect:methodology} presents our proposed IFDL method. Section~\ref{sect:experiments} presents the details of settings and comprehensive experimental results with various datasets and adversarial attack algorithms. Section~\ref{sect:conclusion} concludes the paper and gives the prospect of our future work.

\section{Related Work}
\label{sect:related_work}
\subsection{Image Forgery Detection and Localization}
Image forgeries often leave subtle traces that can be used for detection and localization. Early IFDL methods typically rely on statistical priors, such as CFA (Color Filter Array) artifacts~\cite{popescu2005exposing, ferrara2012image}, compression inconsistencies~\cite{farid2009exposing, lin2009fast, bianchi2012image}, PRNU pattern~\cite{lukas2006digital, chierchia2011prnu}, or edge artifacts~\cite{dong2009run}, but often suffer from poor generalization in increasingly complex and post-processed scenes. 

Recent learning-based methods, built upon CNN and transformer architectures~\cite{kadha2025unravelling}, have significantly improved IFDL performance by effectively capturing subtle forgery traces. For instance, MVSS-Net++~\cite{dong2022mvss} introduces multi-view feature learning and multi-scale supervision to capture semantic-agnostic forgery traces by jointly leveraging RGB, noise, and edge views. CAT-Net v2~\cite{kwon2022learning} learns compression artifacts from DCT coefficient distributions and utilizes them to localize forged regions. IF-OSN~\cite{wu2022robust} enhances robustness under real-world transmission distortions by modeling both predictable and unseen degradations introduced by online social networks. TruFor~\cite{guillaro2023trufor} combines RGB information and a learned noise-sensitive fingerprint to improve the generalization. CoDE~\cite{peng2024employing} formulates image forgery localization as a Markov Decision Process and uses reinforcement learning to iteratively refine pixel-level forgery mask with improved robustness.

However, Peng et. al. \cite{peng2025active} have revealed that existing IFDL methods are highly vulnerable to adversarial attacks. To address this challenge, we propose a unified framework that not only achieves state-of-the-art performance in IFDL task, but also achieves superior adversarial robustness against various adversarial attack methods.

\subsection{Adversarial Attack on PEFT-based SAM Variants}
PEFT (Parameter-efficient Fine-tuning)~\cite{han2024parameterefficient} has emerged as a practical approach for adapting large models like SAM~\cite{kirillov2023segment} to downstream IFDL task. Prior works such as AutoSAM~\cite{su2024novel} and SAFIRE~\cite{kwon2025safire} incorporate adapters into SAM's image encoder to enhance its ability to capture forgery traces. FakeShield~\cite{xufakeshield} and SIDA~\cite{huang2025sida} apply LoRA (Low-rank Adaption)~\cite{hu2022lora} to fine-tune LLaVa~\cite{liu2023visual} for explanatory text generation, while simultaneously keeping SAM's image encoder frozen and tuning only its mask decoder for IFDL. However, these PEFT-based SAM variants primarily be optimized on clean images, largely overlooking their vulnerability to adversarial attacks.

Recently, growing attention has been paid to the adversarial vulnerability of SAM and its variants. PATA++~\cite{zheng2024black} achieves prompt-agnostic targeted attack on SAM by perturbing only the image encoder, improving transferability through feature dominance regularization. DarkSAM~\cite{zhou2024darksam} proposes a universal prompt-free attack that disrupts both spatial semantics and frequency-domain features, causing SAM to fail across varied prompts. UMI-GRAT~\cite{xiatransferable} crafts transferable adversarial images by extracting SAM’s intrinsic vulnerabilities via meta-initialization, enhancing the attack transferability to medical image segmentation, shadow segmentation and camouflaged object segmentation.

\section{Methodology}
\label{sect:methodology}
\begin{figure*}[t]
	\centering 
	\begin{overpic}[width=1.0\textwidth]{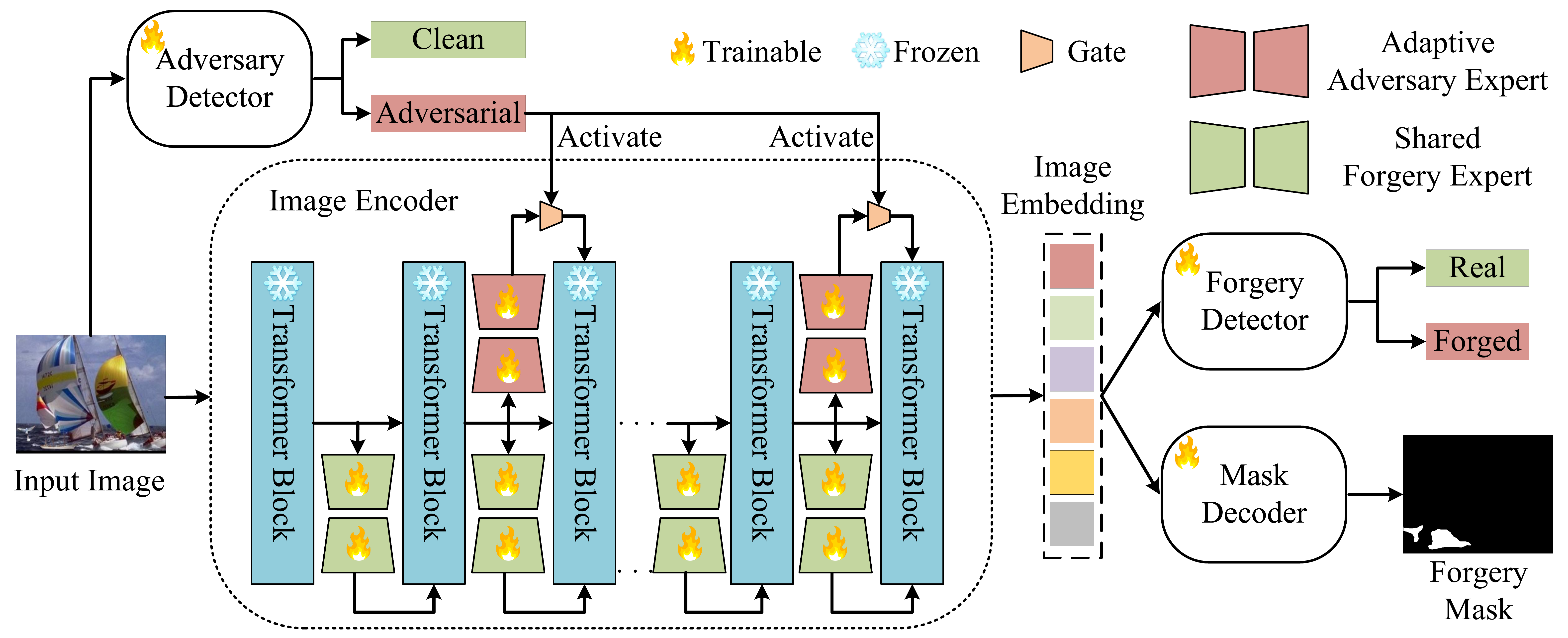} 
	\end{overpic}
	\caption{Overview of the proposed ForensicsSAM framework. Given an input image, an adversary detector first predicts whether the input is clean or adversarial. Within the image encoder, shared forgery experts and adaptive adversary experts cooperatively refine intermediate features into robust image embedding. The forgery detector and mask decoder use the robust image embedding to generate image-level  detection result (real or forged) and a pixel-level forgery mask.}
	\label{fig:ForensicsSAM}
\end{figure*}
\subsection{Problem Statement}
\label{sect:problem_statement}
Given an input image $X \in \mathbb{R}^{C \times H \times W}$, the objective of image forgery detection and localization is twofold: 
\textbf{(1) to classify $X$ as either real or forged image; and 
	(2) to localize the forged region}. Formally, it requires to learn a mapping:
\begin{equation}
	\mathcal{F}: X \mapsto (S_{f}, M_{f})
\end{equation}
where $S_{f} \in (0,1)$ denotes the image-level forgery score. $M_{f} \in (0,1)^{H \times W}$ denotes the pixel-level forgery mask. $H$ and $W$ denote the height and width of the input image, respectively. Both $S_{f}$ and $M_{f}$ use 0.5 as the forgery threshold.

However, through empirical analysis, we find that existing IFDL methods built upon large-scale pretrained model via PEFT technique are often vulnerable to transferable adversarial attacks. Specifically, effective adversarial images can be generated solely via the upstream model, without accessing the downstream model or corresponding training data.
\begin{equation}
	X_{a} = A(X) = \underset{X':\ \|X' - X\|_p \leq \epsilon}{\arg\max} \ d\left(E_{\theta}(X'), E_{\theta}(X)\right)
\end{equation}
where $A(\cdot)$ denotes any adversarial attack method, $d(\cdot, \cdot)$ denotes any distance metric, and $E_{\theta}(\cdot)$ denotes the image encoder of upstream model. In PEFT, the $E_{\theta}(\cdot)$ is typically kept frozen. When $X_{a}$ is given as input, the downstream task-specific modules receive and process the corrupted features produced by $E_{\theta}(X_{a})$, ultimately leading to incorrect predictions for both forgery detection and localization. To ensure robustness under such threats, a robust IFDL model must satisfy two additional properties: 
\textbf{(3) to classify $X$ as either clean or adversarial images; and (4) to correct feature shifts caused by adversarial noise.}

\subsection{Overview}
\label{overview}
In this section, we present an overview of the proposed model, ForensicsSAM, which is built upon SAM~\cite{kirillov2023segment} and incorporates LoRA~\cite{hu2022lora} for adaption in IFDL task.
As illustrated in Fig.~\ref{fig:ForensicsSAM}, ForensicsSAM first injects shared forgery experts into the image encoder to enhance its capability in capturing universal forgery-relevant features among real, forged, and adversarial images. To mitigate the feature shifts caused by adversarial noise, adaptive adversary experts are selectively activated to correct corrupted features. These enhanced features are then passed to a forgery detector and a mask decoder to perform image-level real-forged detection and produce the pixel-level forgery mask, respectively. Meanwhile, an adversary detector identifies whether the image is adversarial and dynamically controls the activation of the adversary experts accordingly. The following sections detail our solution to the aforementioned four objectives.

\subsection{Shared Forgery Experts}
Although the original SAM's image encoder $E_\theta$ can capture rich semantic and structural information of input image, it lacks intrinsic knowledge of forgery artifacts, which are essential for image forgery detection and localization. To address this, we insert forgery experts via LoRA into the image encoder and fine-tune them to extract universal forgery-relevant feature. Besides, although real, forged, and adversarial images differ in appearance, they all contain universal discriminative cues. Thus, these forgery experts are always activated and shared by any input image. For ease of understanding, we provide the formulation below. Specifically, the shared forgery experts are inserted into each attention block's qkv projection layer to guide attention toward forged region, and into the first linear layer of the MLP (Multi-Layer Perceptron) module to adapt token-wise processing to the modified attention patterns.
\begin{equation}
	\begin{aligned}
		q &= W_{q}^{ori} x_{a},\quad k = W_{k}^{ori} x_{a}, \\
		v &= W_{v}^{ori} x_{a},\quad z = W_{z}^{ori} x_{m}
	\end{aligned}
\end{equation}
This represents the frozen original qkv (query, key, value) projection and linear projection with weights $W_{*}^{ori} \in \mathcal{R}^{D\times d}$. And $x_{a},x_{m} \in \mathcal{R}^{N\times D}$ denote the respective input patch embeddings. $N$, $D$, and $d$ are the token number, embedding dimension, and hidden dimension, respectively.
\begin{equation}
	\begin{aligned}
		q' &= q + W_{q'}^{u} W_{q'}^{d} x_{a},\quad k' = k + W_{k'}^{u} W_{k'}^{d} x_{a},\\
		v' &= v + W_{v'}^{u} W_{v'}^{d} x_{a},\quad z' = z + W_{z'}^{u} W_{z'}^{d} x_{m}
	\end{aligned}
	\label{eq:update_forgery_experts}
\end{equation}
This represents the low-rank updates to the qkv projection and linear projection. $W_{*}^{u} \in \mathcal{R}^{D\times r}$ and $ W_{*}^{d} \in \mathcal{R}^{r\times d}$ are the weights of LoRA up-projection and down-projection matrices. Based on empirical studies, we set the rank to $r=8 \ll d$ in our implementation. In the end, we denote the weights of all shared forgery experts as $\phi$, and the new image encoder is denoted as $E_{\theta, \phi}$.

\subsection{Adversary Detector}
\begin{figure}[t]
	\centering
	\def\fig_height{3.5cm}
	\begin{overpic}[height=\fig_height]{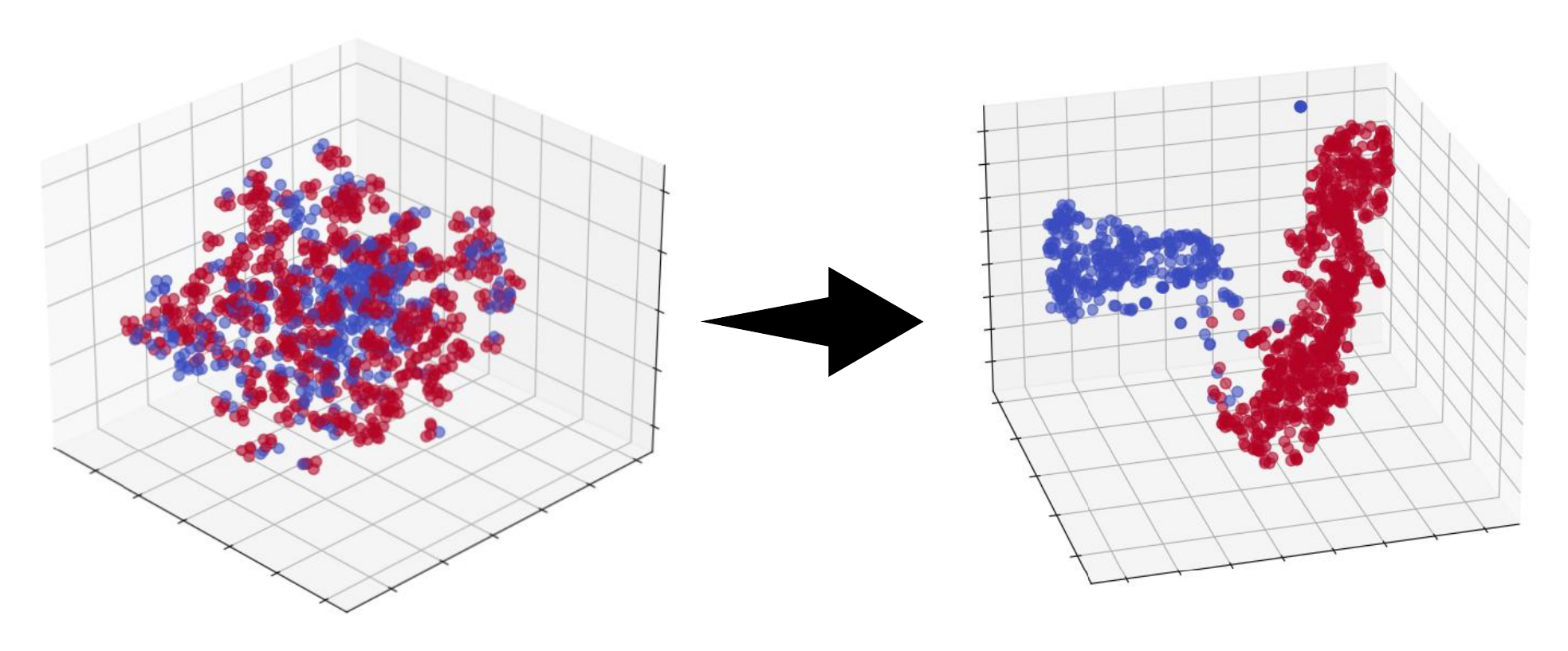}
	\end{overpic}
	\caption{Feature distributions of real/forged (blue) and adversarial (red) images before and after adversary detector optimization.}
	\label{fig:ad_feature} 
\end{figure}
Adversarial noise, unlike random noise, is deliberately crafted to induce misbehavior in the model. As a result, it tend to introduce structured, task-dependent artifacts in the RGB domain that deviate from natural image statistics. Such artifacts can be exploited as detectable signals. To capture such artifacts, we introduce a light-weight adversary detector $D_a$, which uses ResNet-18~\cite{he2016deep} as the backbone, followed by a projection layer and a classification head. Given an input image $X$, $D_a$ produces an adversary score $S_{a} \in (0,1)$, indicating the likelihood that $X$ is an adversarial image.
\begin{equation}
	S_{a} = D_{a}(X)\\
	\label{eq:output_adversary_detector}
\end{equation}

To enhance its discriminative capability, we adopt a combination of BCE loss $\mathcal{L}^{a}_{det}$ and contrastive learning via triplet loss $\mathcal{L}_{tri}$~\cite{schroff2015facenet}. The $\mathcal{L}_{bce}^{a}$ supervises the adversary score $S_{a}$. And the $\mathcal{L}_{tri}$ explicitly encourages the learned feature to be clustered by type (clean or adversarial) in the latent space. As shown in Fig.~\ref{fig:ad_feature}, all images are mixed and inseparable before optimization. While after optimization, the real and forged images (clean images, denoted in blue) are clustered together, while the adversarial images (denoted in red) form a separate cluster.
\begin{equation}
	\mathcal{L}_{tri} = \max\left(0, \left\| f_a - f_p \right\|_2^2 - \left\| f_a - f_n \right\|_2^2 + \alpha \right)
\end{equation}
where $f_a$, $f_p$, and $f_n$ are the features from the projection layer for anchor, positive, and negative images, respectively. $\alpha$ is a predefined margin. $f_a$ is a feature of a given input image, $f_p$ is drawn from another image of the same type (e.g., both clean or both adversarial), and $f_n$ is taken from a image of the opposite type. The final loss function combines both terms, where the weight of $\mathcal{L}_{tri}$ is set to $\lambda=0.5$.
\begin{equation}
	\mathcal{L}_{1} = \mathcal{L}_{bce}^{a} + \lambda \mathcal{L}_{tri}
	\label{eq:loss_adversary_detector}
\end{equation}

\subsection{Adaptive Adversary Experts}
\begin{figure}[t]
	\centering
	\def\fig_height{2cm}
	\setlength\tabcolsep{2mm}
	\subfloat[Before Attack]{
	\begin{tabular}{c c c c c c}
		Clean & Attention & Image
		\\
		Image & Map & Embedding
		\\
		\begin{overpic}[height=\fig_height]{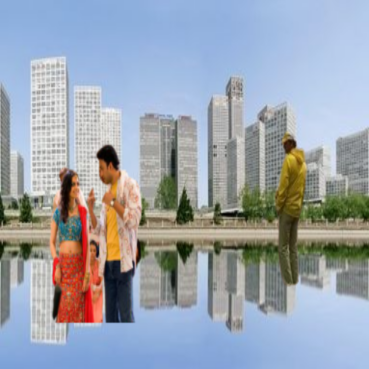}
		\end{overpic}&
		\begin{overpic}[height=\fig_height]{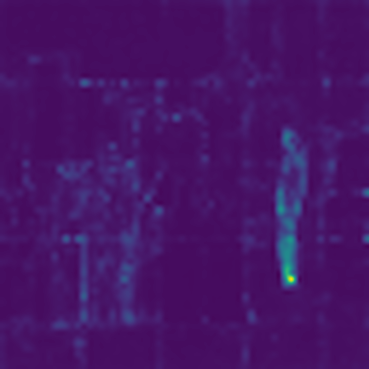}
		\end{overpic}&
		\begin{overpic}[height=\fig_height]{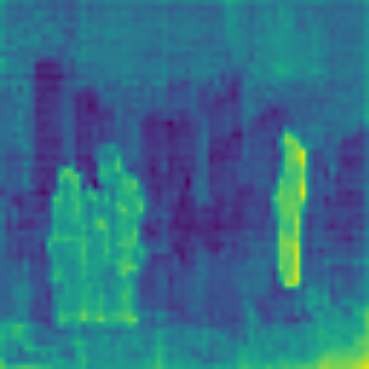}
		\end{overpic}
		\\
		\begin{overpic}[height=\fig_height]{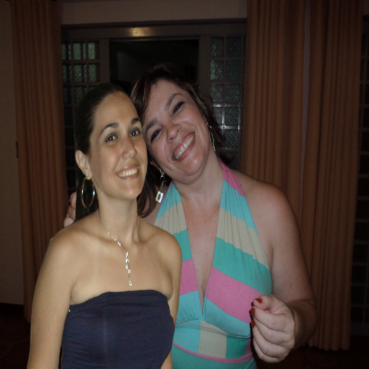}
		\end{overpic}&
		\begin{overpic}[height=\fig_height]{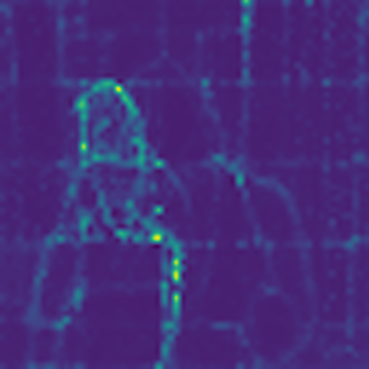}
		\end{overpic}&
		\begin{overpic}[height=\fig_height]{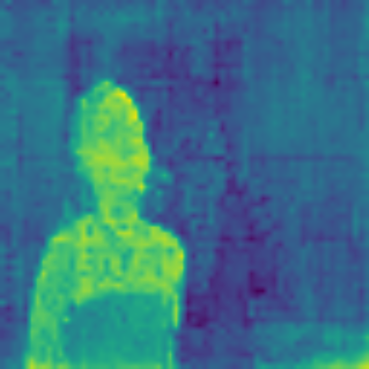}
		\end{overpic}
		\\
	\end{tabular}
	}

	\subfloat[After Attack]{
		\begin{tabular}{c c c}
			Adversarial & Attention & Image
			\\
			Image & Map & Embedding
			\\
			\begin{overpic}[height=\fig_height]{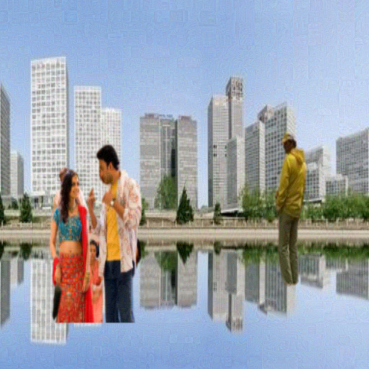}
			\end{overpic}&
			\begin{overpic}[height=\fig_height]{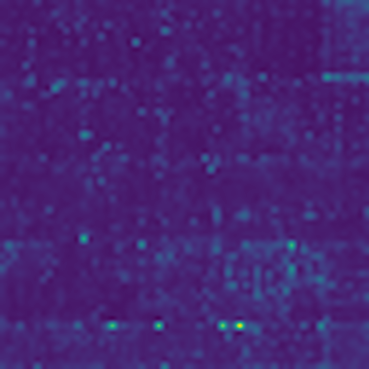}
			\end{overpic}&
			\begin{overpic}[height=\fig_height]{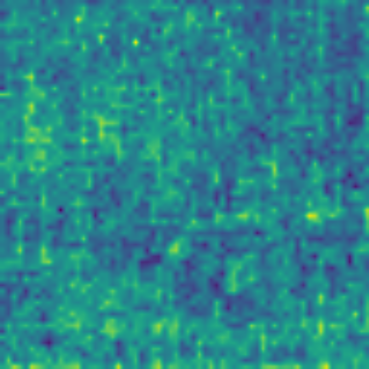}
			\end{overpic}
			\\
			\begin{overpic}[height=\fig_height]{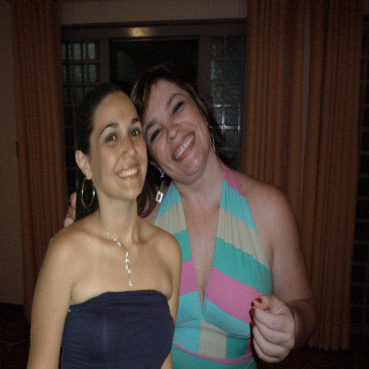}
			\end{overpic}&
			\begin{overpic}[height=\fig_height]{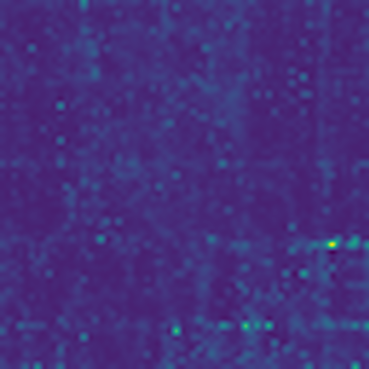}
			\end{overpic}&
			\begin{overpic}[height=\fig_height]{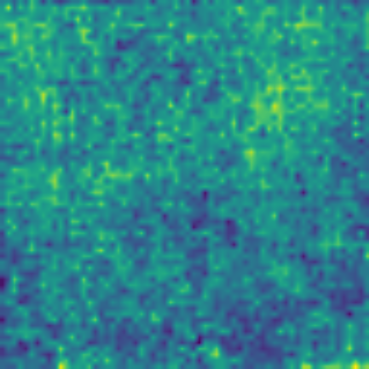}
			\end{overpic}
			\\		
		\end{tabular}
	}
	\caption{Comparison of global attention maps and image embeddings between clean and adversarial images.}
	\label{fig:attn_feature} 
\end{figure}
While the adversary detector provides a global, image-level detection of whether an input image is adversarial, it does not address the robust issue of forgery localization under adversarial attacks. The effectiveness of adversarial attacks targeting upstream SAM stems from their ability to introduce significant feature shifts within the upstream image encoder $E_{\theta}$. The corrupted feature can then propagate to $E_{\theta, \phi}$, so as to degrade the performance of downstream task. Therefore, we propose to insert adaptive adversary experts into $E_{\theta, \phi}$ to correct such adversarially induced feature shifts to ensure robust forgery localization. 

Since adversarial noise is often crafted to achieve pixel-level misbehavior, it tends to be globally distributed across the entire image, thereby affecting a broad spatial context. As illustrated in Fig.~\ref{fig:attn_feature}, compared to clean images, the global attention maps of adversarial images lose clear object boundaries and semantic structure. Moreover, the corresponding image embeddings derived from adversarial images become spatially noisy and less discriminative, confirming that the global attention flow has been disrupted. Thus, we propose to inject the adversary experts into the global transformer blocks, specifically into the global attention layers and their subsequent MLP modules. The update process is like Eq.\eqref{eq:update_forgery_experts}:
\begin{equation}
	\begin{aligned}
		q'' &= q' + G_{a}W_{q''}^{u} W_{q''}^{d} x_{a}, \quad
		k'' = k' + G_{a}W_{k''}^{u} W_{k''}^{d} x_{a},\\
		v'' &= v' + G_{a}W_{v''}^{u} W_{v''}^{d} x_{a}, \quad
		z'' = z' + G_{a}W_{z''}^{u} W_{z''}^{d} x_{m}
	\end{aligned}
	\label{eq:update_adversary_experts}
\end{equation}
where $G_{a} \in \{0, 1\}$ is a binary gate signal derived from the adversary score $S_{a} \in (0,1)$, such that $G_a = \mathbb{I}(S_a > 0.5)$. And $\mathbb{I}(\cdot)$ denotes the indicator function. \(G_a = 0\) disables the adversary experts, while \(G_a = 1\) enables them. Next, we denote the weights of adversary experts as $\psi$, and the new image encoder is denoted as $E_{\theta, \phi, \psi}$.

After updating, we obtain four intermediate embeddings from the global transformer blocks where the adversary experts inserted, and one image embedding from the final layer that reflect progressively corrected features. Let these features be denoted as $\{F_i^{a}\}_{i=1}^5$ for adversarial images and $\{ F_i^{c}\}_{i=1}^5$ for the corresponding clean images. The objective is to align each $F_i^{a}$ with corresponding $F_i^{c}$, thereby progressively correcting the feature shifts caused by adversarial noise.
\begin{equation}
	\mathcal{L}_2 = \sum_{i=1}^{5} 
	\left(\left\| F_i^{a} - F_i^{c} \right\|_2^2 + 1 - \text{cos}(\hat{F}_i^{a}, \hat{F}_i^{c})\right)
	\label{eq:loss_adversary_experts}
\end{equation}
where $\mathcal{L}_2$ consists of an $\ell_2$ loss term and a cosine similarity loss term, and $\hat{F}$ denotes the $\ell_2$-normalized feature vector. The joint loss corrects feature magnitude while enforcing directional alignment.
\begin{algorithm}[t]
	\setstretch{1.0}
	\caption{Pseudo-code of Training Algorithm} 		
	\begin{algorithmic}[1]
		\label{al:training_algorithm}
		\REQUIRE Clean dataset $\mathcal{D}_{c}$ and Adversarial dataset $\mathcal{D}_{a}$; image encoder $E_{\theta, \phi, \psi}$; adversary detector $D_{a}$; forgery detector $D_{f}$; mask decoder $D_{m}$; Training iterations $e_1$, $e_2$, and $e_3$;
		\ENSURE	Well-trained ForensicsSAM;\\
		\textit{\textbf{First Stage: Inject Forgery-specific Knowledge}}\\
		\STATE Initialize $E_{\theta, \phi, \psi}$, $D_{f}$, and $D_{m}$; 
		\STATE Freeze $\theta$ and $\psi$;
		\FOR{epoch in $1$ to $e_1$}
		\FOR{$X_{c}$ $\subset$ $\mathcal{D}_{c}$}
		\STATE $F_{X_{c}} = E_{\theta, \phi, \psi}(X_{c}, 0)$;\hfill Eq.~\eqref{eq:output_image_encoder}\\
		\STATE $S_{f} = D_{f}(X_{c})$;\hfill Eq.~\eqref{eq:output_forgery_detector}\\
		\STATE $M_{f} = M_{f}(X_{c})$;\hfill Eq.~\eqref{eq:output_mask_decoder}\\
		\STATE Update $E_{\theta, \phi, \psi}$, $D_{f}$, and $D_{m}$ via $\mathcal{L}_{3}$;\hfill Eq.~\eqref{eq:loss_shared_forgery_experts}\\
		\ENDFOR
		\ENDFOR
		\STATE $E_{\theta, \phi^{'}, \psi} = E_{\theta, \phi, \psi}$; $D_{f}^{'} = D_{f}$; $D_{m}^{'} = D_{m}$;\\
		\textit{\textbf{Second Stage: Identify Adversarial Images}} \\
		\STATE Initialize $D_{a}$;
		\FOR{epoch in $1$ to $e_2$}
		\FOR{$X$ $\subset$ $(\mathcal{D}_{c}, \mathcal{D}_{a})$}
		\STATE $S_{a} = D_{a}(X)$;\hfill Eq.~\eqref{eq:output_adversary_detector}\\
		\STATE Update $D_{a}$ via $\mathcal{L}_{2}$;\hfill Eq.~\eqref{eq:loss_adversary_detector}\\
		\ENDFOR
		\ENDFOR
		\STATE $D_{a}^{'} = D_{a}$;\\
		\textit{\textbf{Third Stage: Correct Feature Shifts}} \\
		\STATE Freeze $\theta$, $\phi^{'}$, $D_{a}^{'}$, $D_{f}^{'}$, and $D_{m}^{'}$;
		\FOR{epoch in $1$ to $e_3$}
		\FOR{$(X_{c}, X_{a})$ $\subset$ $(\mathcal{D}_{c}, \mathcal{D}_{a})$}
		\STATE Get $\{F_i^{c}\}_{i=1}^5$ from $E_{\theta, \phi^{'}, \psi}(X_{c}, \mathbb{I}(D_{a}^{'}(X_{c})))$;\hfill Eq.~\eqref{eq:output_image_encoder}\\
		\STATE Get $\{F_i^{a}\}_{i=1}^5$ from $E_{\theta, \phi^{'}, \psi}(X_{a}, \mathbb{I}(D_{a}^{'}(X_{a})))$;\\
		\STATE Update $E_{\theta,\phi^{'},\psi}$ via $\mathcal{L}_{1}$;\hfill Eq.~\eqref{eq:loss_adversary_experts}\\
		\ENDFOR
		\ENDFOR
		\STATE $E_{\theta, \phi^{'}, \psi^{'}} = E_{\theta, \phi^{'}, \psi}$;
		\RETURN $E_{\theta, \phi^{'}, \psi^{'}}$, $D_{a}^{'}$, $D_{f}^{'}$, $D_{m}^{'}$;
	\end{algorithmic} 
	\label{algo:training_algorithm}
\end{algorithm}
\subsection{Forgery Detection and Localization Module}
After redefining the image encoder via shared forgery experts and adaptive adversary experts, we can obtain the forgery-relevant feature (image embedding) $F_{X}$ from $E_{\theta, \phi, \psi}$. 
\begin{equation}
	F_{X} = E_{\theta, \phi, \psi}(X, G_a)\\
	\label{eq:output_image_encoder}
\end{equation}
Then $F_{X}$ is utilized to perform image-level detection and pixel-level localization. 

For detection, we design a light-weight forgery detector $D_{f}$. Given the $F_{X}$, we first apply adaptive average pooling to obtain a global descriptor. This descriptor is then passed through two linear layers with a ReLU activation. Finally, a sigmoid function is applied to produce a forgery score $S_{f}$.
\begin{equation}
	S_{f} = D_{f}(F_{X})\\
	\label{eq:output_forgery_detector}
\end{equation}
where higher $S_{f}$ indicates higher likelihood of a forged image. The detection loss $\mathcal{L}_{bce}^{f}$ is computed using the BCE (Binary Cross-Entropy) loss. 

For localization, we fully fine-tune SAM's mask decoder $D_{m}$ to predict a forgery mask $M_{f}$ from given $F_{X}$. 
\begin{equation}
	M_{f} = D_{m}(F_{X})\\
	\label{eq:output_mask_decoder}
\end{equation}
Since forgery localization does not involve any user interaction, the SAM's prompt encoder is not used during training or inference. As a result, prompt-based attacks are not applicable to our ForensicsSAM. The localization loss is defined as a weighted combination of dice loss $\mathcal{L}_{dice}^{loc}$ and pixel-level BCE loss $\mathcal{L}_{bce}^{loc}$. In the end, the detection and localization are trained joinly using the following loss function, where $a, b, c$ are set to 1.0, 0.7, 0.3, respectively in our implementation.
\begin{equation}
	\mathcal{L}_{3} = a\cdot\mathcal{L}_{bce}^{f} + (b\cdot\mathcal{L}_{dice}^{loc} + c\cdot\mathcal{L}_{bce}^{loc})
	\label{eq:loss_shared_forgery_experts}
\end{equation}

\subsection{Training Pipeline}
In this section, we present the detailed training algorithm of ForensicsSAM, which involves three stage. The pseudo-code is presented in Algorithm~\ref{algo:training_algorithm}.
\subsubsection{First Stage} In this stage, only the shared forgery experts, forgery detector, and mask decoder are trained to predict an image-level forgery score $S_{f}$ and a pixel-level forgery mask $M_{f}$. These trainable components are optimized by Eq.~\eqref{eq:loss_shared_forgery_experts}. Notably, adversarial images are not involved at this stage.
\subsubsection{Second Stage} This stage uses both adversarial images and corresponding clean images. The adversary detector is trained to produce an image-level adversary score $S_{a}$, optimized using Eq.\eqref{eq:loss_adversary_detector}.
\subsubsection{Third Stage} In this stage, only the adaptive adversary experts are activated, while all other components remain frozen. Both adversarial and clean images are fed into the image encoder to obtain paired features $\{F_i^{a}\}_{i=1}^5$ and $\{F_i^{c}\}_{i=1}^5$, which are then aligned via Eq.~\eqref{eq:loss_adversary_experts}. It is worth emphasizing that both the second and third stages require only a small subset of the training data used in the first stage. Furthermore, the adversarial images are generated exclusively using MI-FGSM method~\cite{dong2018boosting} to achieve superior adversarial robustness.
\begin{table}[t]
	\centering
	\caption{Training and test datasets with the number of real and forged images, and the manipulation types: splicing (SP), copy-move (CM), and inpainting (INP).}
	\tabcolsep=0.2cm
	\begin{tabular}{c r | c c c c c c c}
		& \multirow{2}*{Dataset} & \multicolumn{2}{c}{Number} & \multicolumn{3}{c}{Forgery types}\\
		& & Real & Forged & SP & CM & INP\\
		\hline
		\multirow{4}{*}{\rotatebox{90}{Train}} & CASIAv2 & 7491 & 5098 & \checkmark & \checkmark \\
		& IMD20 & 414 & 2000 & \checkmark & \checkmark \\
		& FantasticReality & 16592 & 19423 & \checkmark \\
		& TamperedCR & 24462 & 23981 & \checkmark & \checkmark \\
		\hline
		\multirow{10}{*}{\rotatebox{90}{Test}} & CASIAv1+ & 800 & 920 & \checkmark & \checkmark \\
		& MISD & 620 & 296 & \checkmark \\
		& Columbia & 183 & 180 & \checkmark\\
		& DSO-1 & 100 & 100 & \checkmark\\
		& Coverage & 100 & 100 & & \checkmark \\
		& NIST & 875 & 564 & \checkmark & \checkmark & \checkmark\\
		& CocoGlide & 512 & 512 &  &  & \checkmark\\
		& IPM15k & - & 15000 & \checkmark & \checkmark\\
		& ACDSee & 364 & 337 & \checkmark & \checkmark & \checkmark \\
		& In-the-wild & - & 201 & \checkmark\\
		\hline        
	\end{tabular}
	\label{tab:datasets}
\end{table}

\subsection{Inference Pipeline}
The well-trained ForensicsSAM comprises four components: image encoder $E_{\theta, \phi^{\prime}, \psi^{\prime}}$, adversary detector $D_{a}^{\prime}$, forgery detector $D_{f}^{\prime}$, and mask decoder $D_{m}^{\prime}$. Given an input image $X$, $D_{a}^{\prime}$ first predicts an adversary score $S_a$, which is then binarized using a threshold of $0.5$ to obtain a gating signal $G_a$, indicating whether the adversary experts should be activated. Then $E_{\theta, \phi^{\prime}, \psi^{\prime}}$ then takes $X$ and $G_a$ as input and produces the image embedding $F_X$. Finally, $F_X$ is fed into both $D_{f}^{\prime}$ and $D_{m}^{\prime}$ to generate the image-level forgery score $S_f$ and the pixel-level forgery mask $M_f$, respectively. Both $S_f$ and $M_f$ are also binarized using a threshold of $0.5$ to obtain binary predictions.

\section{Experiments}
\begin{figure}[!t]
	\centering
	\def\fig_height{1.9cm}
	\setlength\tabcolsep{1.5mm}
	\begin{tabular}{c c c c >{\centering\arraybackslash}m{1.5cm}}
		\multirow{2}*{Image} & \multirow{2}*{GT} & Forgery & Forgery
		\\
		& & Mask & Score
		\\
		\begin{overpic}[width=\fig_height]{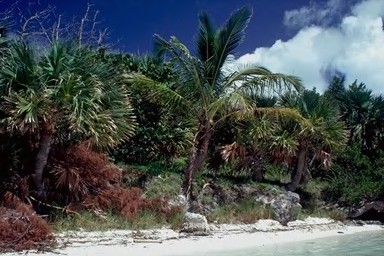}
		\end{overpic}&
		\begin{overpic}[width=\fig_height]{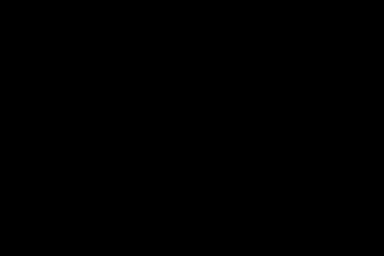}
		\end{overpic}&
		\begin{overpic}[width=\fig_height]{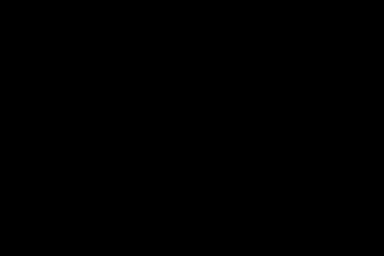}
			\put(120, 30){0.0022}
			\put(120, -50){0.0057}
			\put(120, -125){0.9922}
			\put(120, -220){0.9532}
		\end{overpic}&
		\\
		\begin{overpic}[width=\fig_height]{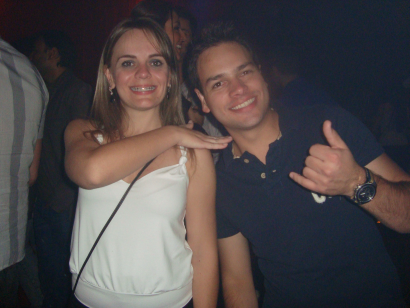}
		\end{overpic}&
		\begin{overpic}[width=\fig_height]{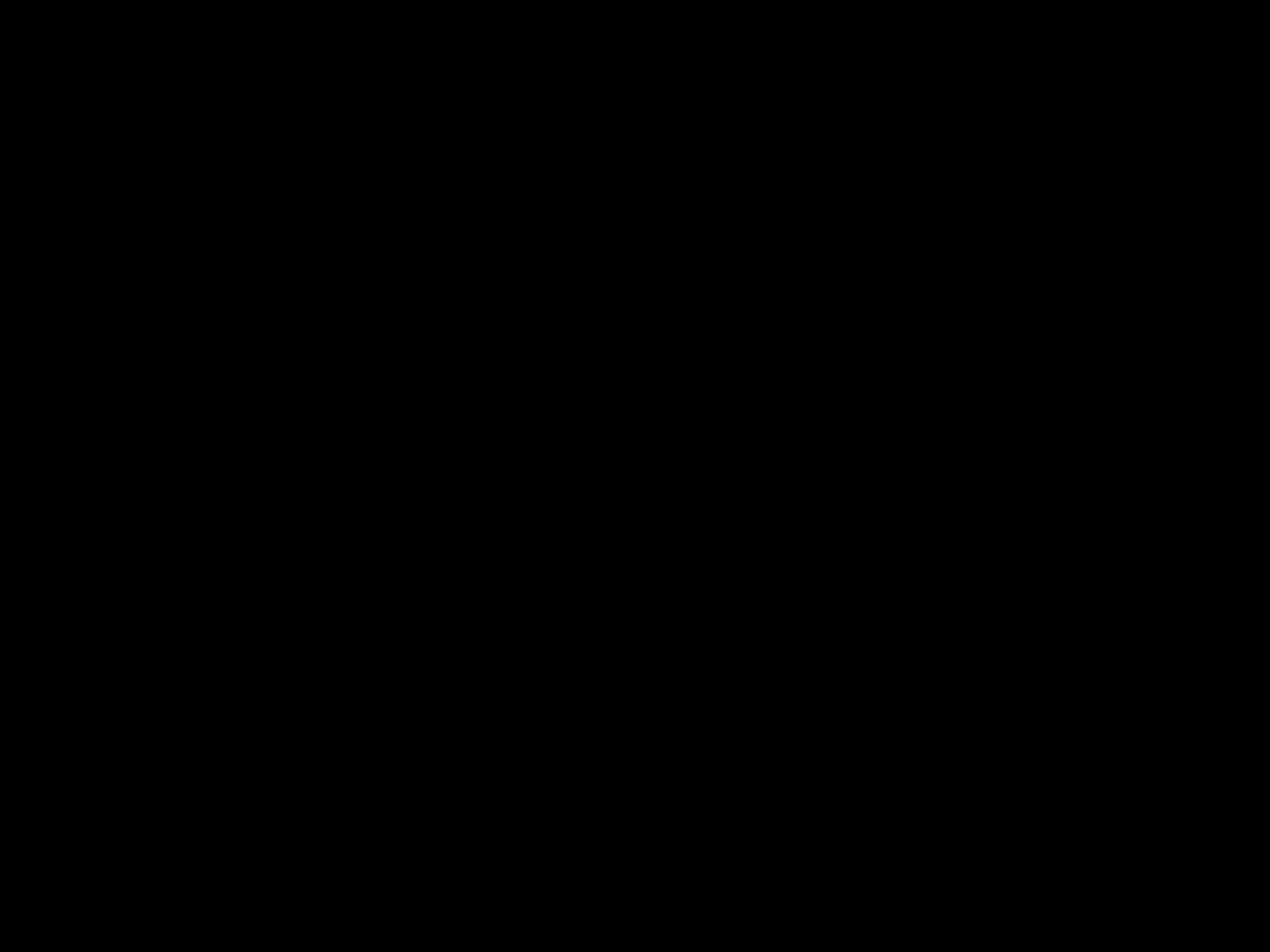}
		\end{overpic}&
		\begin{overpic}[width=\fig_height]{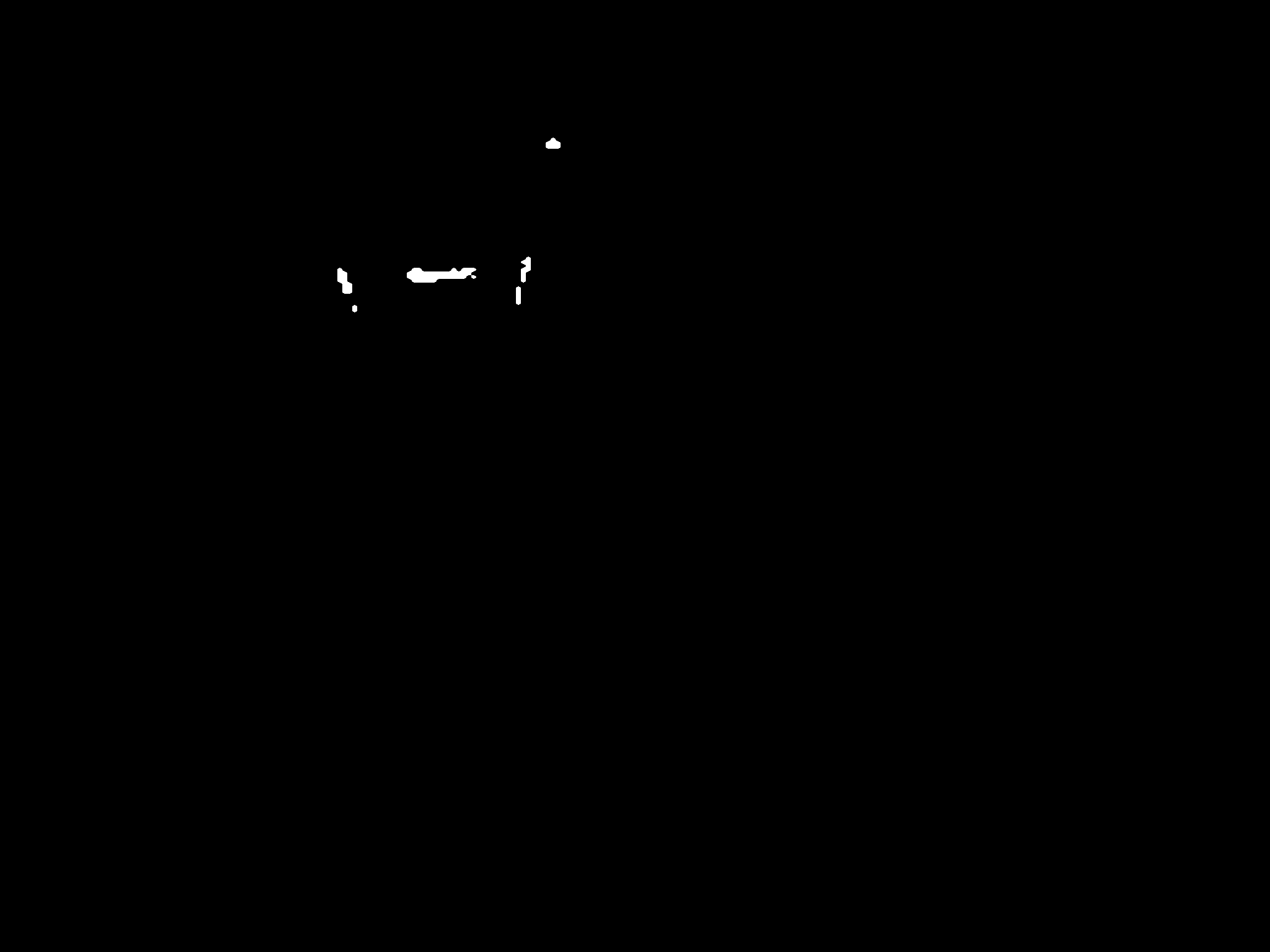}
		\end{overpic}&
		\\
		\begin{overpic}[width=\fig_height]{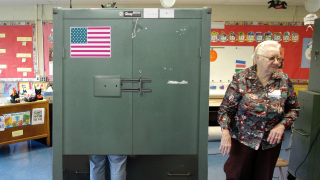}
		\end{overpic}&
		\begin{overpic}[width=\fig_height]{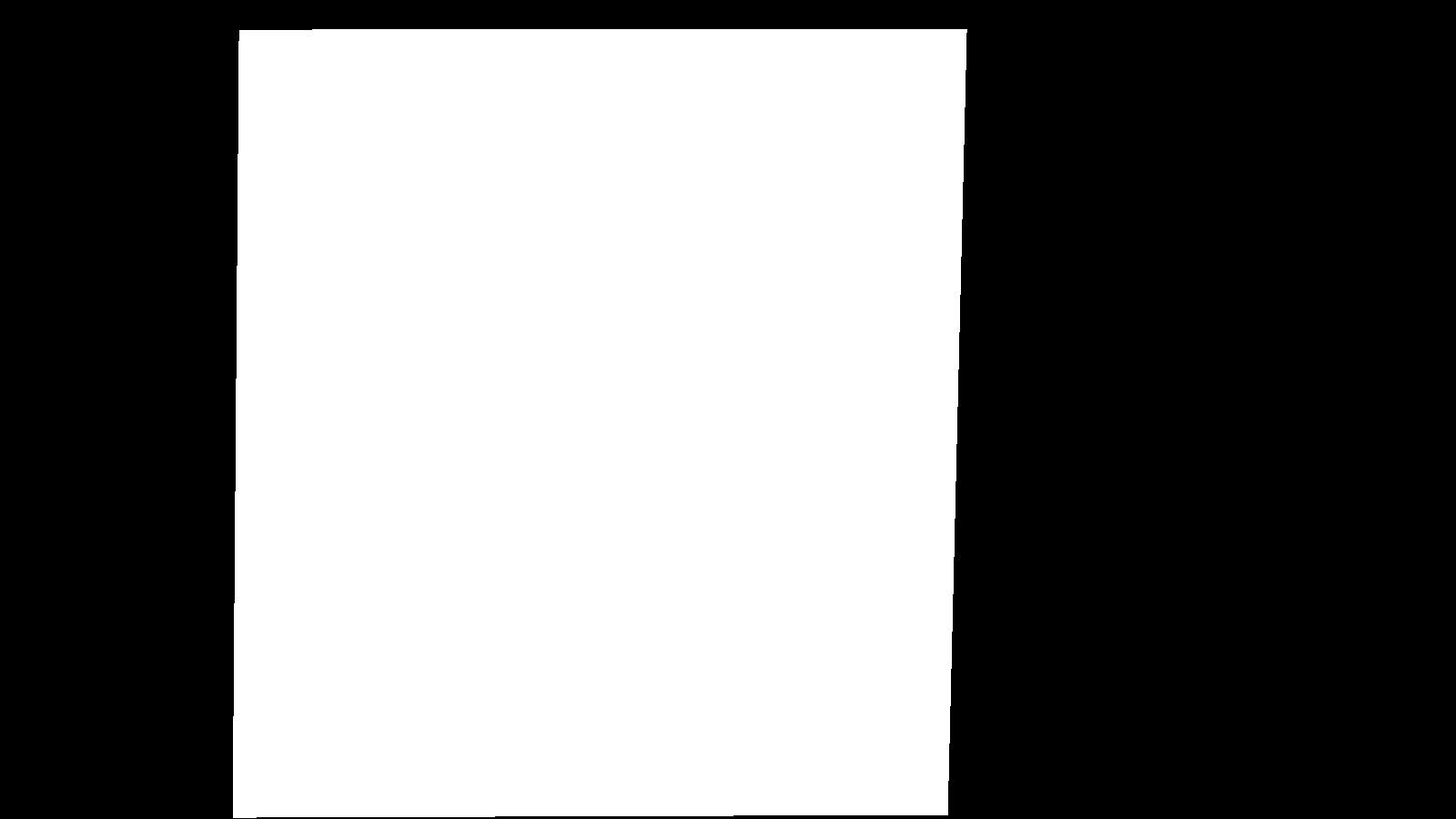}
		\end{overpic}&
		\begin{overpic}[width=\fig_height]{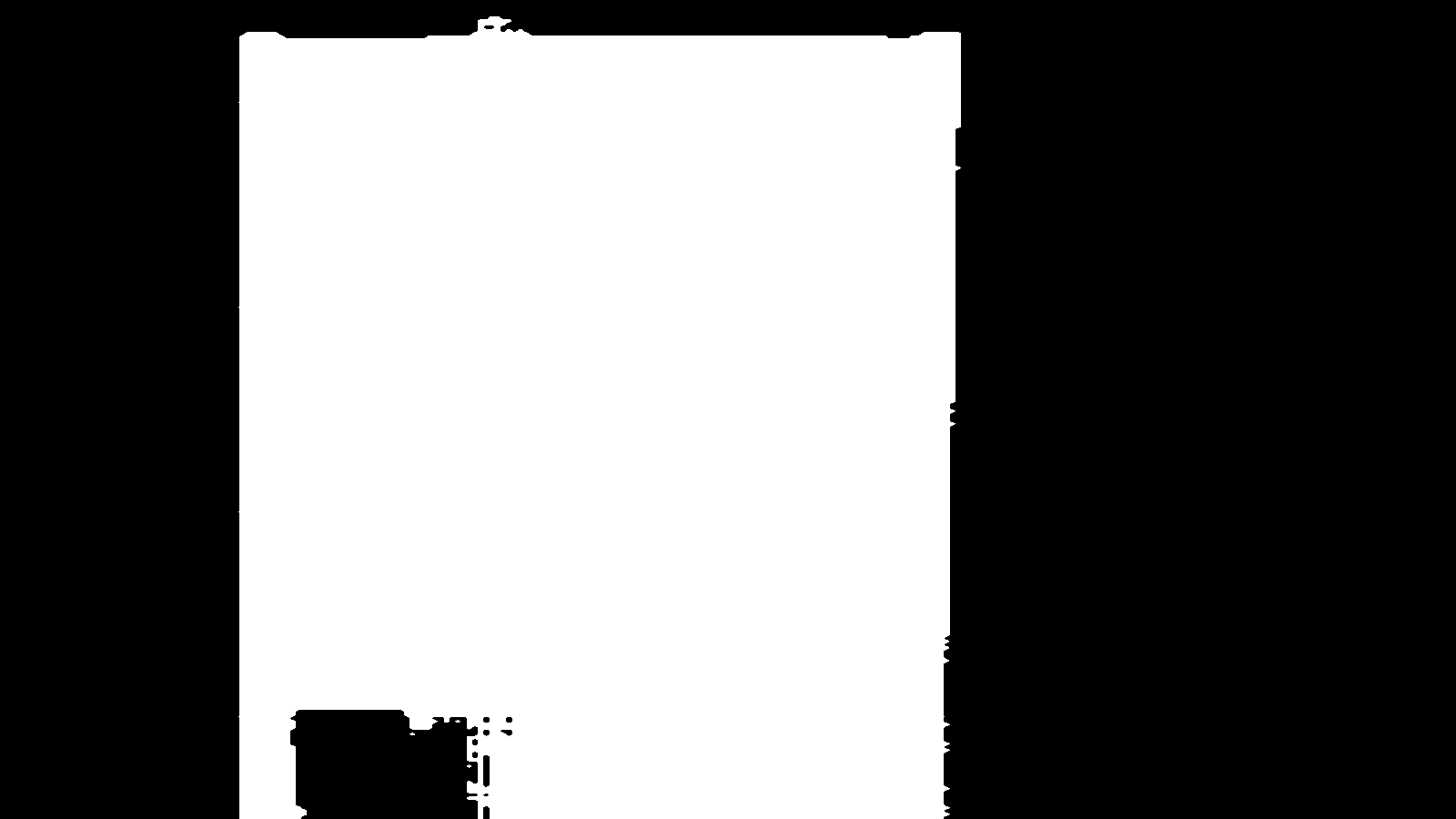}
		\end{overpic}&
		\\
		\begin{overpic}[width=\fig_height]{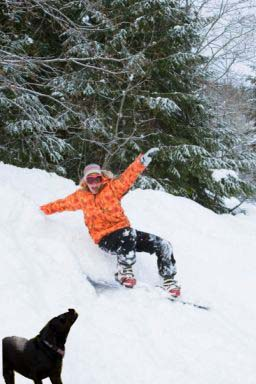}
		\end{overpic}&
		\begin{overpic}[width=\fig_height]{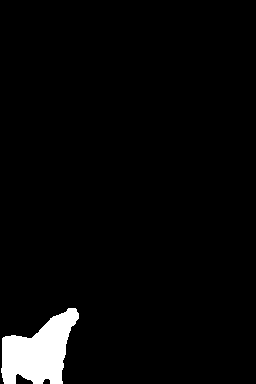}
		\end{overpic}&
		\begin{overpic}[width=\fig_height]{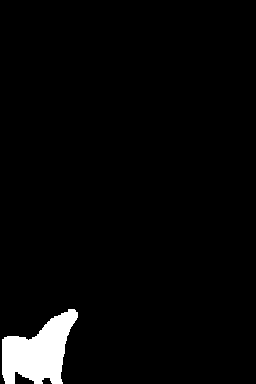}
		\end{overpic}&
		\\
	\end{tabular}
	\caption{The forgery detection and localization results of real and forged images.}
	\label{fig:detection_results} 
\end{figure}
\label{sect:experiments}
\subsection{Experimental Setup}
\subsubsection{IFDL Baseline}
We compare our proposed ForensicsSAM with the SOTA (state-of-the-art) methods, including MVSS-Net++~\cite{dong2022mvss}, IF-OSN~\cite{wu2022robust}, CAT-Net v2~\cite{kwon2022learning}, TruFor~\cite{guillaro2023trufor}, CoDE~\cite{peng2024employing}, AutoSAM~\cite{su2024novel}, FakeShield~\cite{xufakeshield}, and SAFIRE~\cite{kwon2025safire}. For fairness, we directly adopt their official pre-trained weights, and evaluate the performance using the same metrics.

\subsubsection{Adversarial Attack Methods}
We evaluate the model's resistance to adversarial attacks with recent adversarial attack methods, including gradient-based MI-FGSM~\cite{dong2018boosting} and PGN~\cite{ge2023boosting}, input augmentation-based BSR~\cite{wang2024boosting}, and transfer-based UMI-GRAT~\cite{xiatransferable}. Following common practice, for all attack methods, we set the perturbation bound $\varphi \in \{8, 12\}$, step size $\alpha =2$, and the attack update iterations $T=\varphi$. 

Note that all adversarial images are crafted only using the upstream SAM's image encoder, without accessing the downstream PEFT-based SAM variants. And the competitors for comparison of adversarial robustness include PEFT-based AutoSAM, FakeShield, and SAFIRE.

\subsubsection{Datasets} Table~\ref{tab:datasets} shows the number of real and forged images, as well
as the tampering manipulations for each dataset. Following the protocols of SOTA IFDL methods, we use CASIAv2~\cite{dong2013casia}, FantasiticReality~\cite{kniaz2019point}, IMD20~\cite{novozamsky2020imd2020} and TamperedCR~\cite{kwon2022learning} as the training set. And the test set includes CASIAv1+~\cite{dong2022mvss}, MISD~\cite{kadam2021multiple}, Columbia~\cite{ng2004data}, DSO-1~\cite{de2013exposing}, Coverage~\cite{wen2016coverage}, NIST~\cite{guan2019mfc}, CocoGlide~\cite{guillaro2023trufor}, IPM15k~\cite{ren2023mfi}, ACDSee, and In-the-wild~\cite{huh2018fighting}.

\subsubsection{Evaluation Metrics}
Following common evaluation protocols~\cite{kwon2022learning, guillaro2023trufor, su2024novel, kwon2025safire}, we adopt Accuracy (ACC) for image-level forgery detection and permute F1-score (hereinafter referred to as F1) for pixel-level forgery localization. Higher ACC and F1 indicate better overall performance.

\subsubsection{Implementation Details} 
\begin{figure*}[t]\footnotesize
	\centering
	\def\fig_width{1.5cm}
	\setlength\tabcolsep{0.7mm}
	\begin{tabular}{c c c c c c c c c c c }
		\begin{overpic}[width=\fig_width]{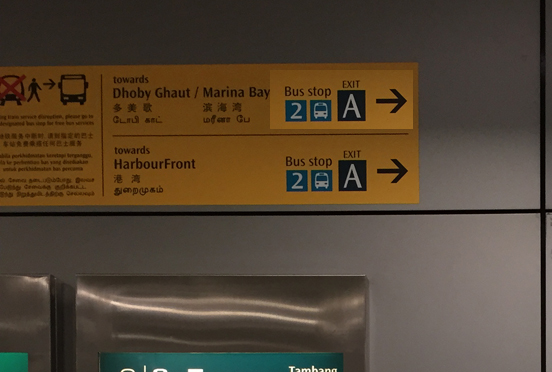}
		\end{overpic}&
		\begin{overpic}[width=\fig_width]{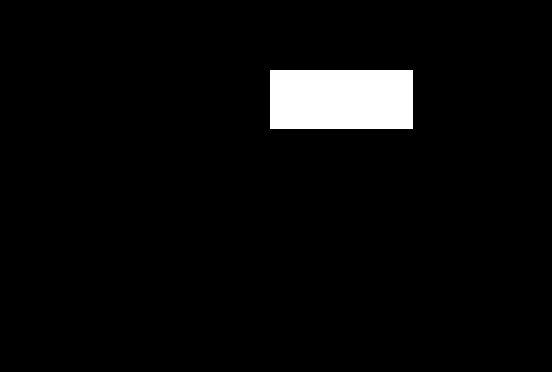}
		\end{overpic}&
		\begin{overpic}[width=\fig_width]{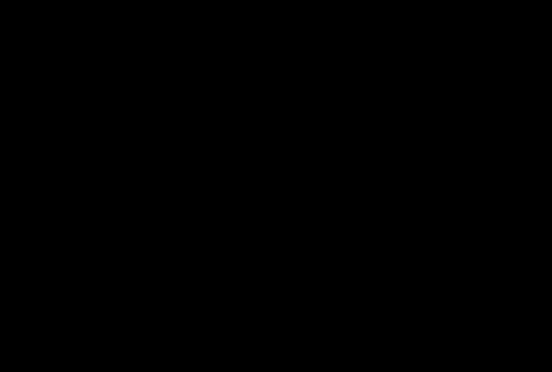}
		\end{overpic}&
		\begin{overpic}[width=\fig_width]{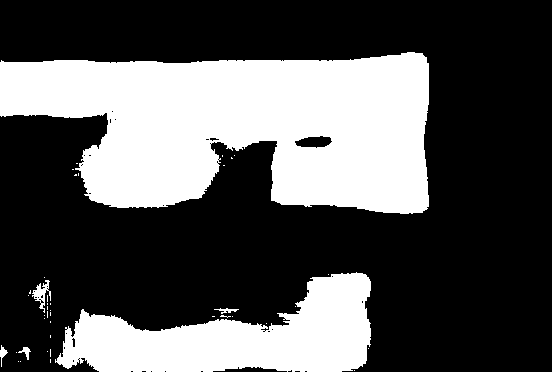}
		\end{overpic}&
		\begin{overpic}[width=\fig_width]{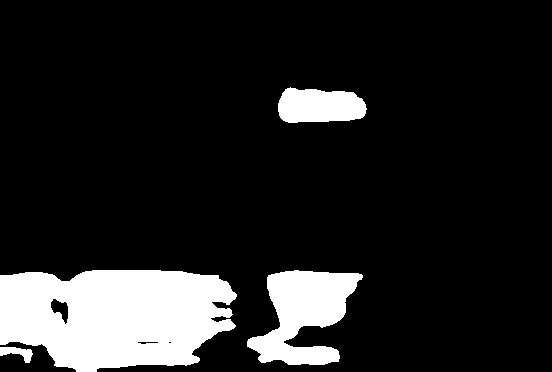}
		\end{overpic}&
		\begin{overpic}[width=\fig_width]{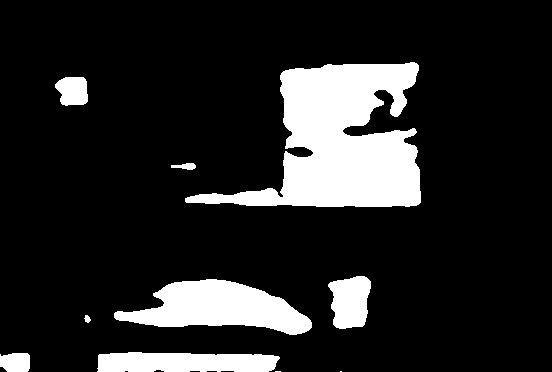}
		\end{overpic}&
		\begin{overpic}[width=\fig_width]{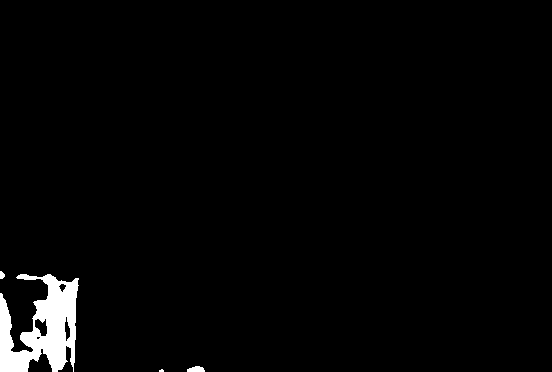}
		\end{overpic}&
		\begin{overpic}[width=\fig_width]{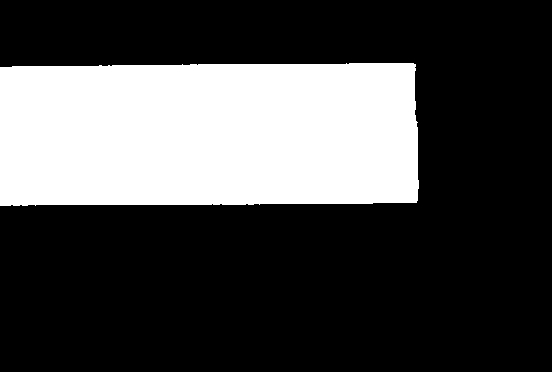}
		\end{overpic}&
		\begin{overpic}[width=\fig_width]{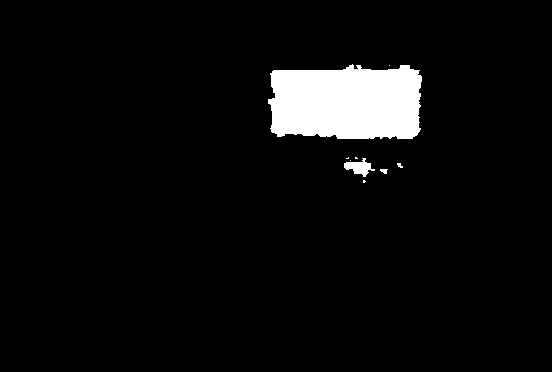}
		\end{overpic}&
		\begin{overpic}[width=\fig_width]{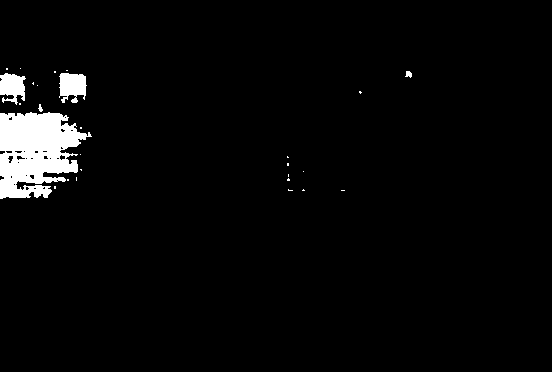}
		\end{overpic}&
		\begin{overpic}[width=\fig_width]{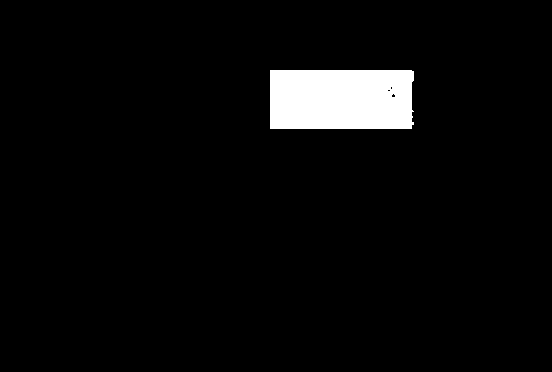}
		\end{overpic}    
		\\
		\begin{overpic}[width=\fig_width]{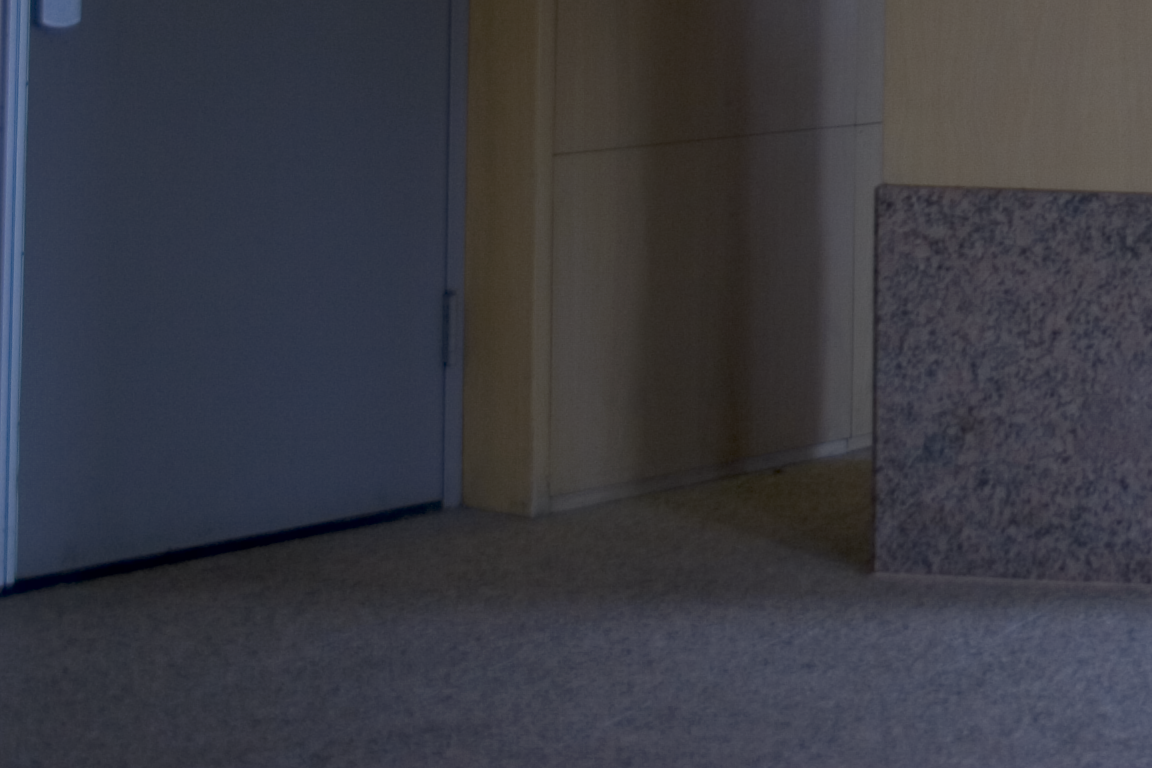}
		\end{overpic}&
		\begin{overpic}[width=\fig_width]{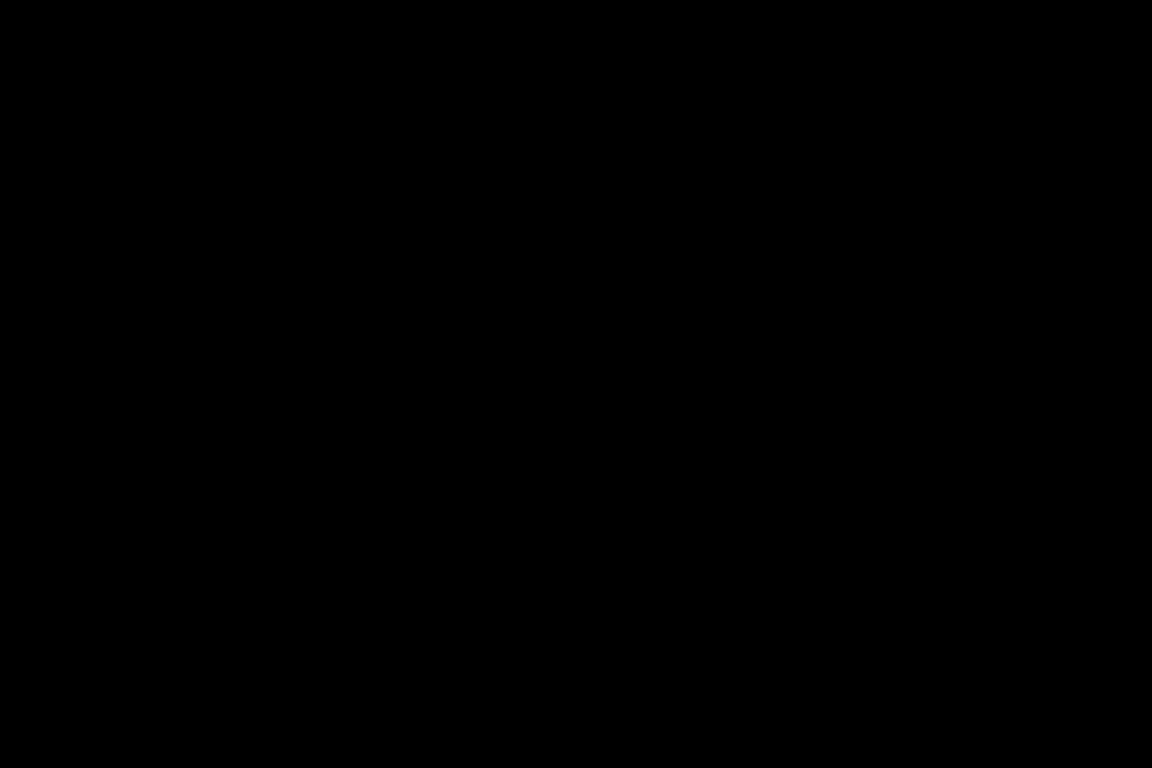}
		\end{overpic}&
		\begin{overpic}[width=\fig_width]{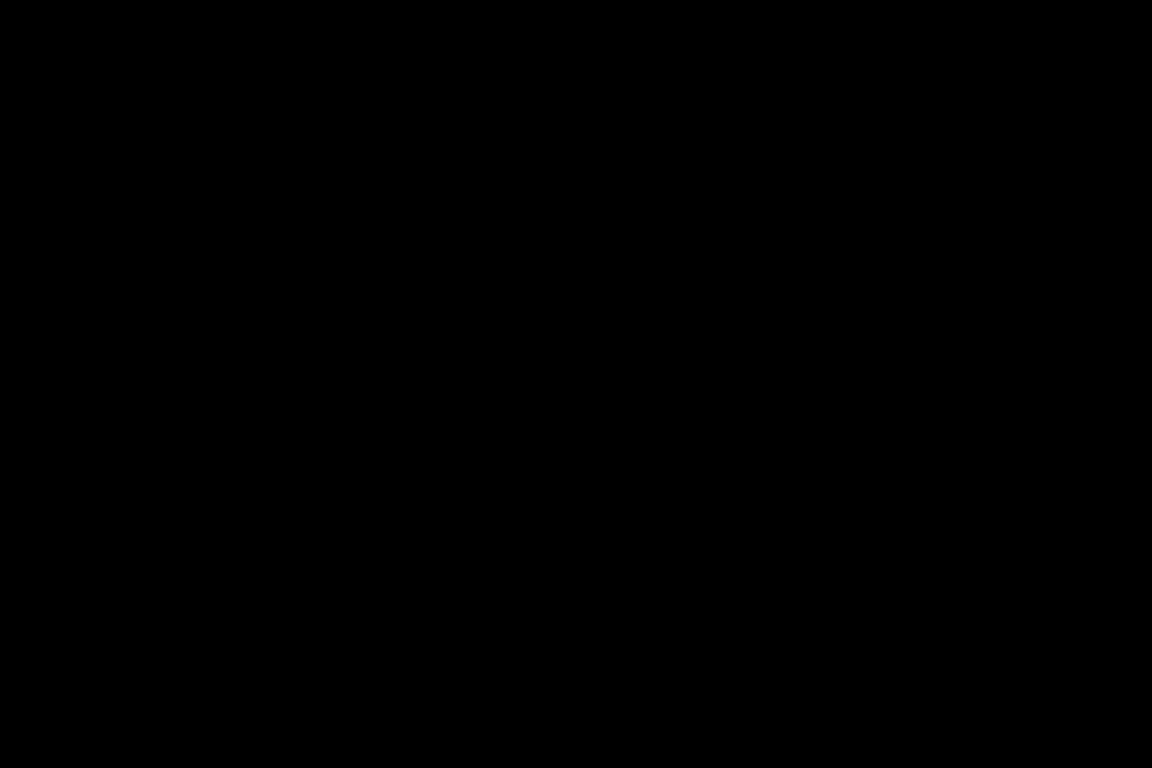}
		\end{overpic}&
		\begin{overpic}[width=\fig_width]{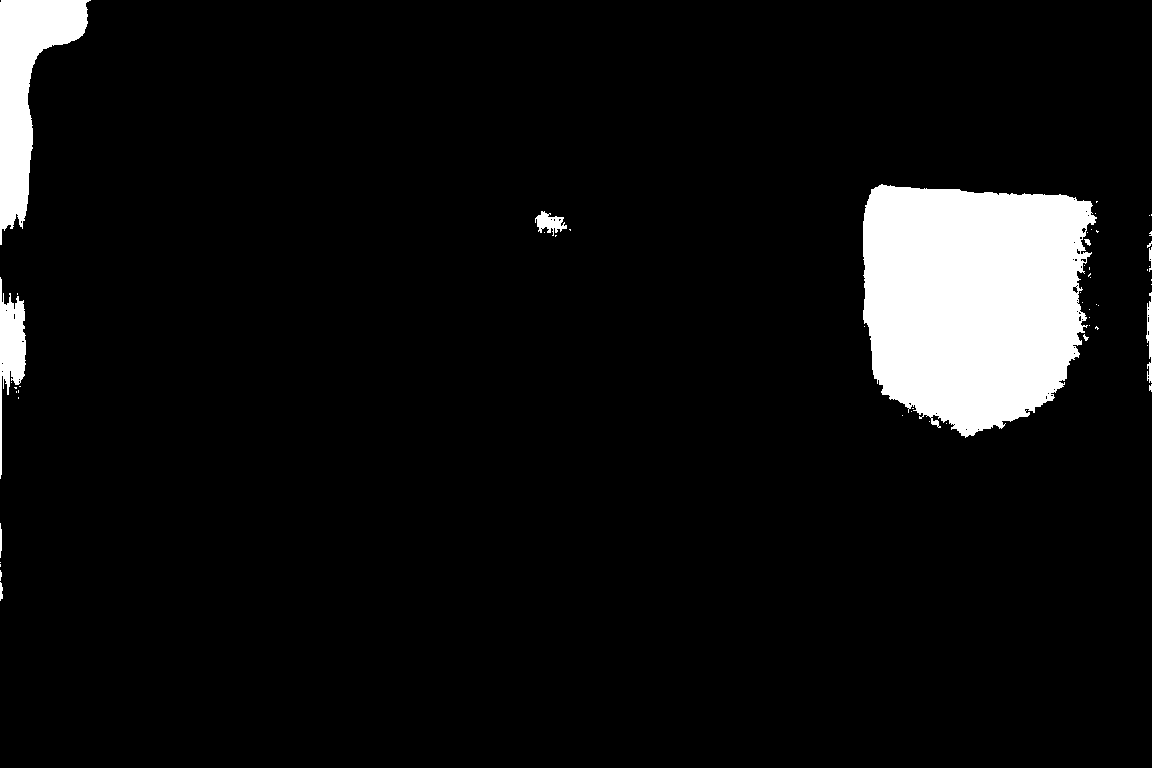}
		\end{overpic}&
		\begin{overpic}[width=\fig_width]{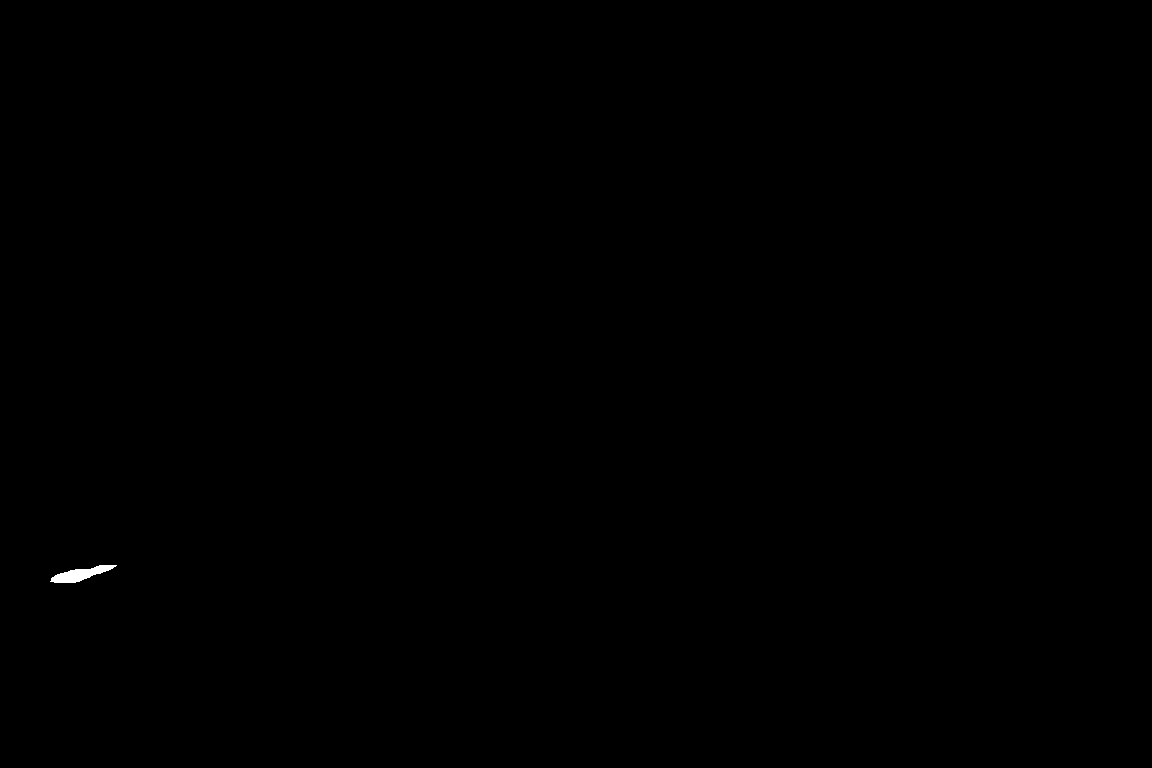}
		\end{overpic}&
		\begin{overpic}[width=\fig_width]{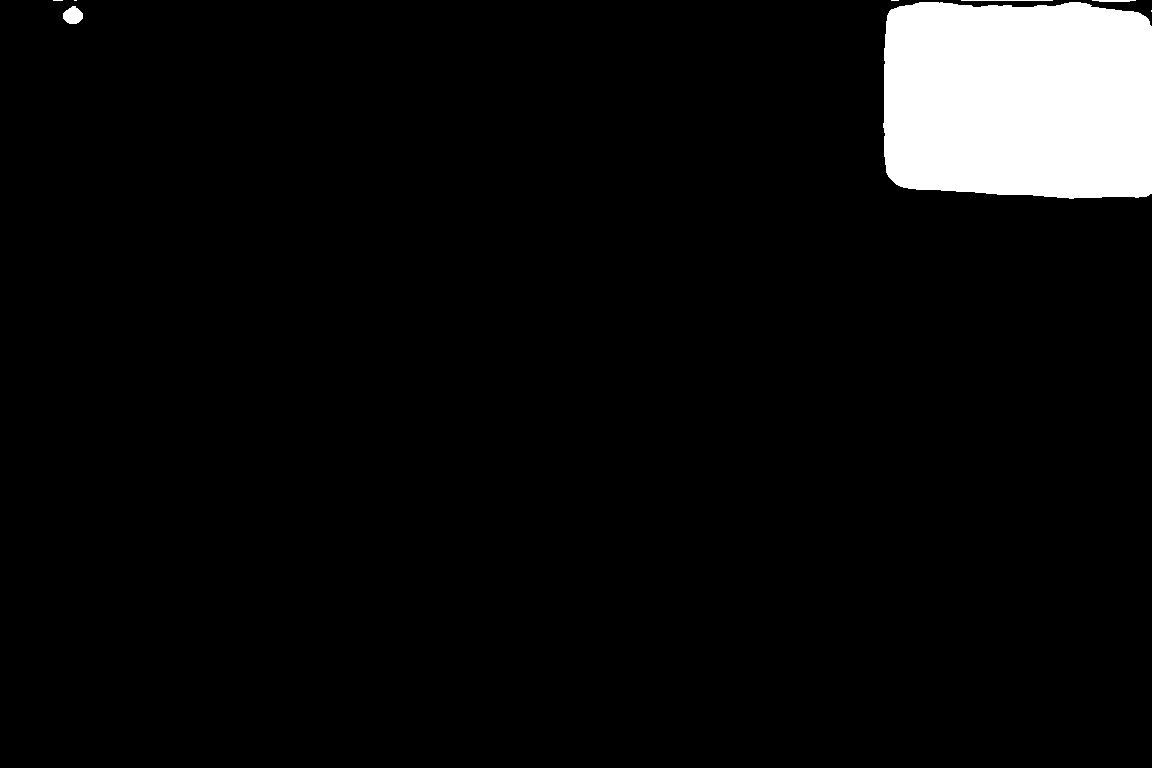}
		\end{overpic}&
		\begin{overpic}[width=\fig_width]{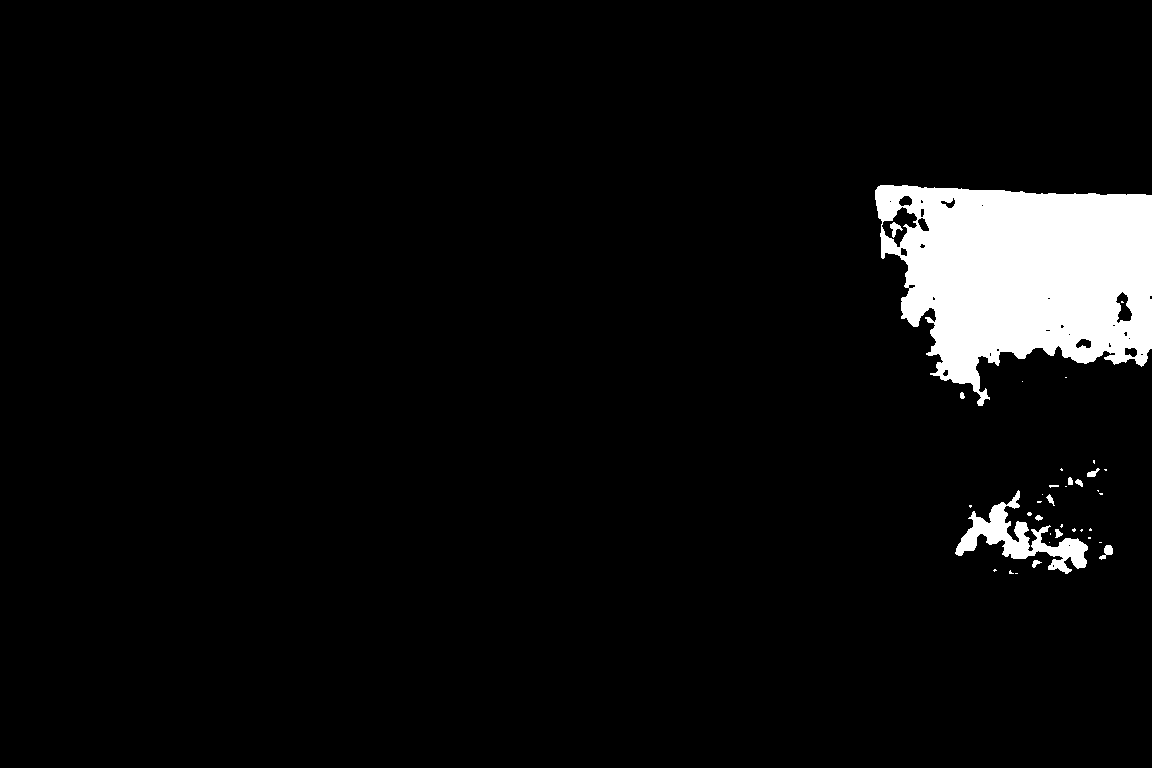}
		\end{overpic}&
		\begin{overpic}[width=\fig_width]{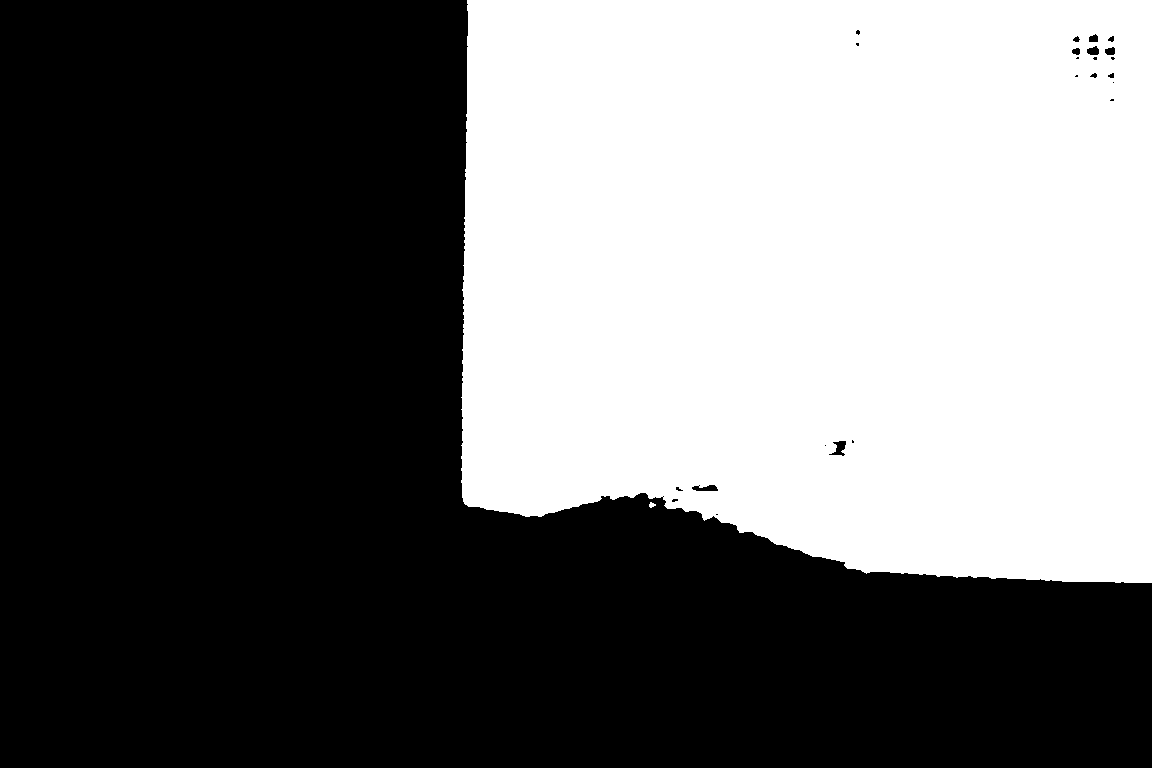}
		\end{overpic}&
		\begin{overpic}[width=\fig_width]{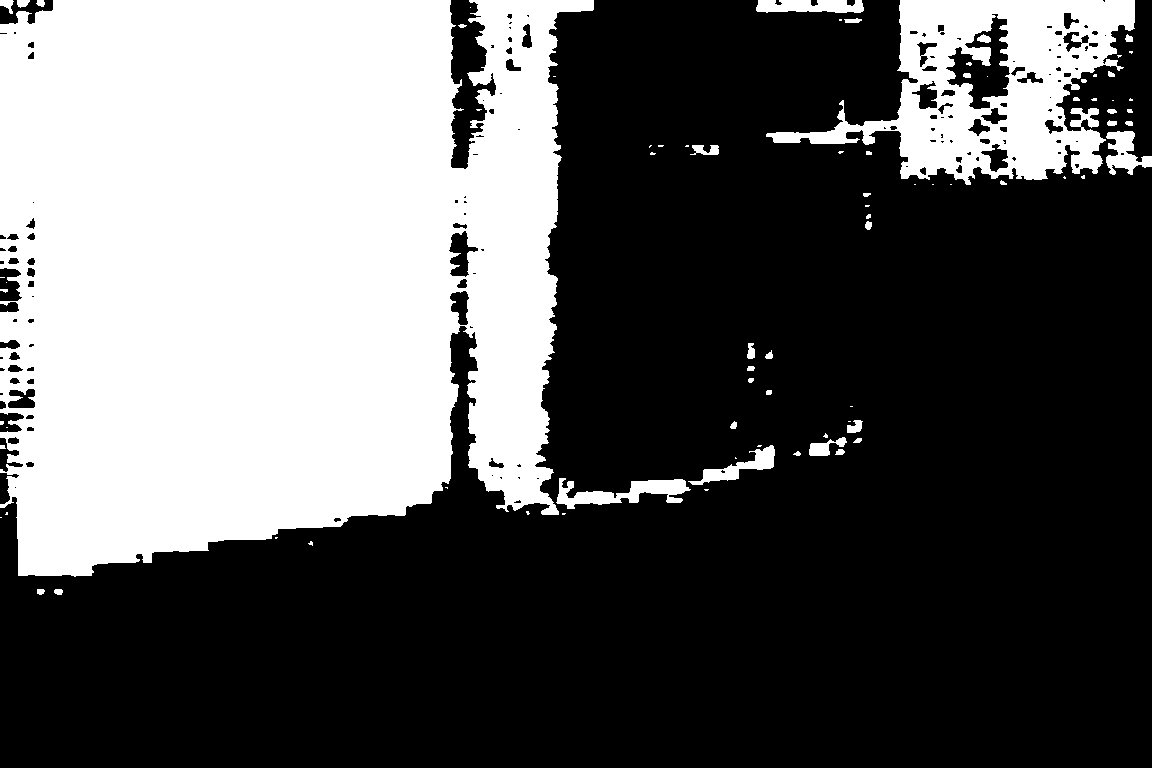}
		\end{overpic}&
		\begin{overpic}[width=\fig_width]{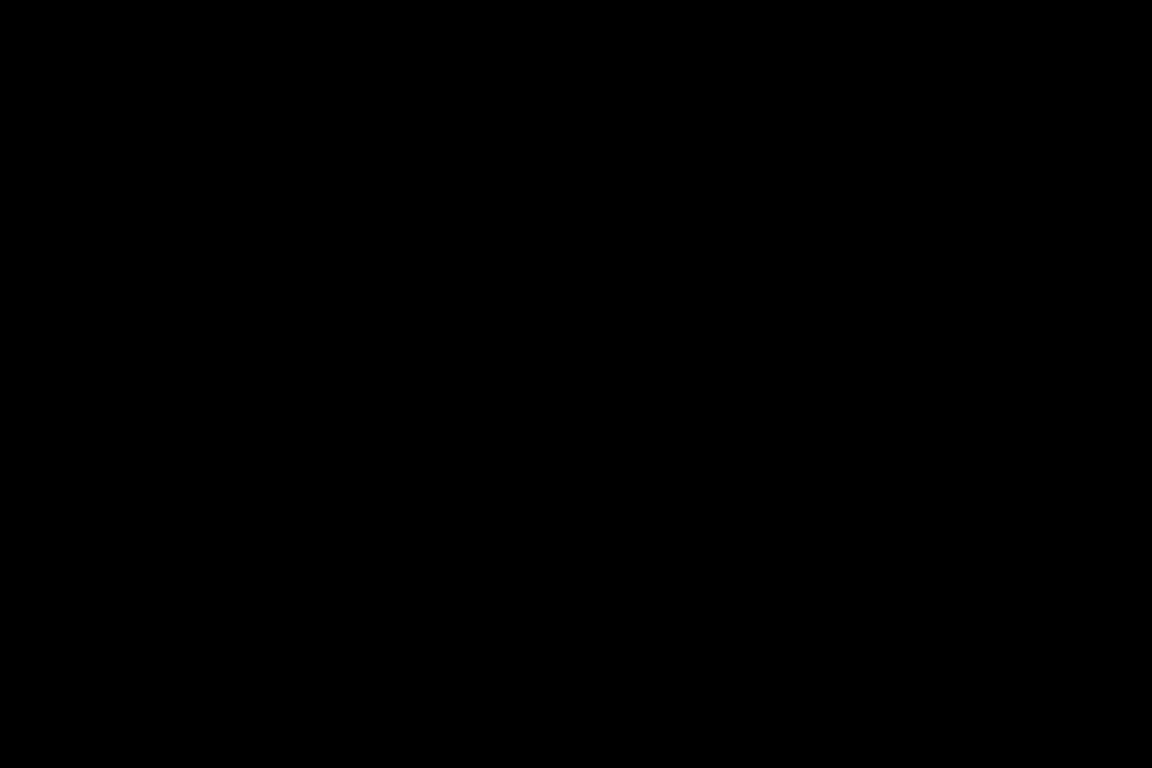}
		\end{overpic}&
		\begin{overpic}[width=\fig_width]{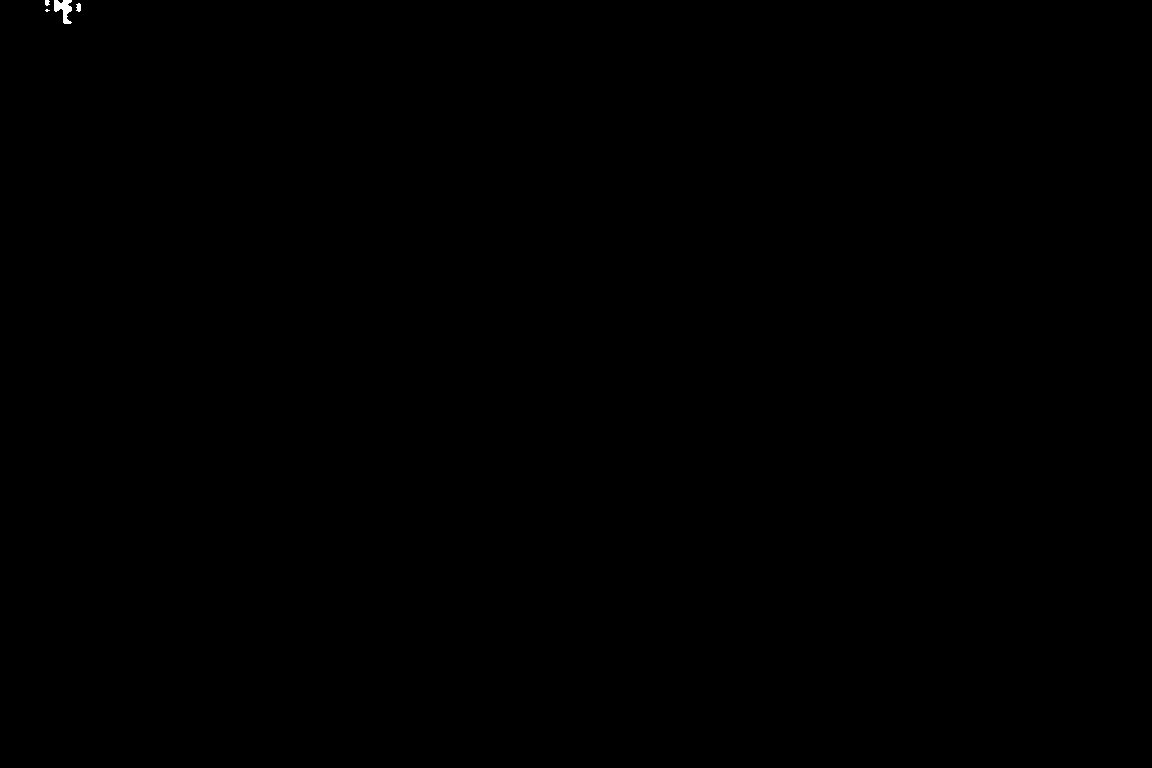}
		\end{overpic}    
		\\
		\begin{overpic}[width=\fig_width]{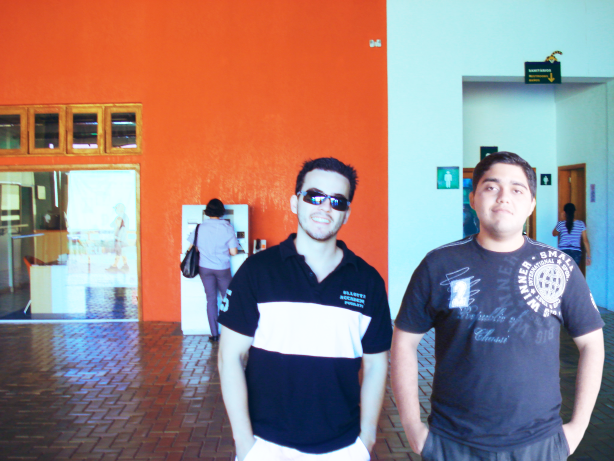}
		\end{overpic}&
		\begin{overpic}[width=\fig_width]{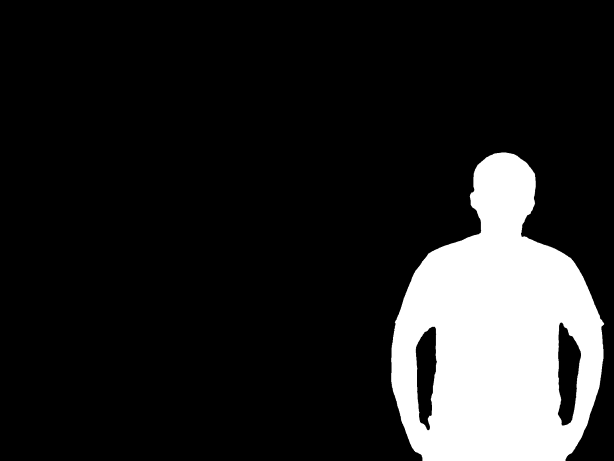}
		\end{overpic}&
		\begin{overpic}[width=\fig_width]{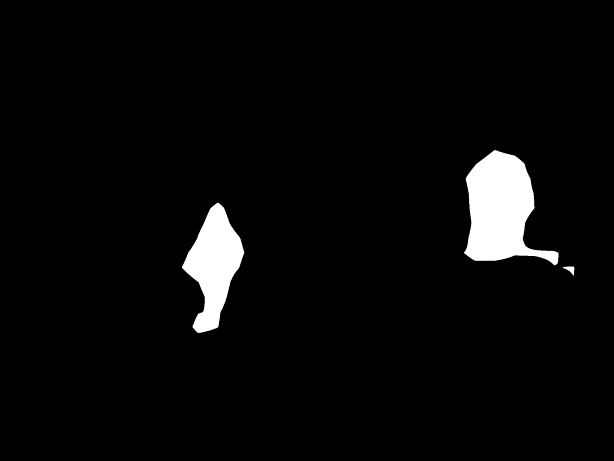}
		\end{overpic}&
		\begin{overpic}[width=\fig_width]{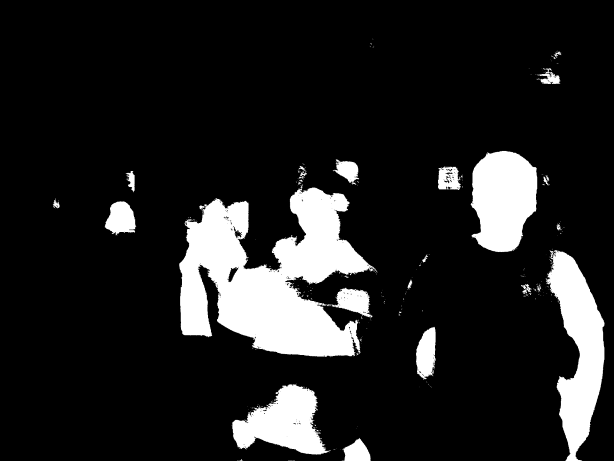}
		\end{overpic}&
		\begin{overpic}[width=\fig_width]{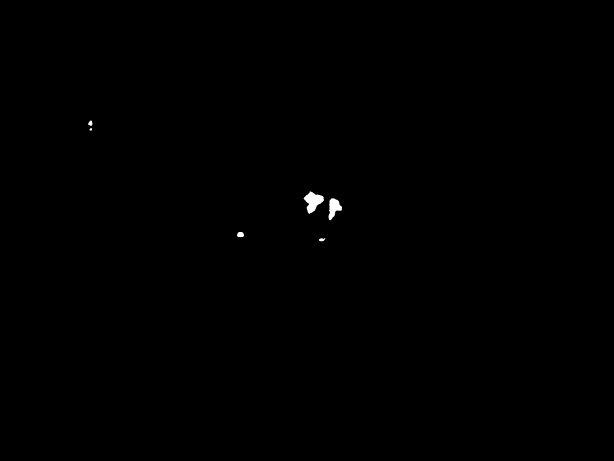}
		\end{overpic}&
		\begin{overpic}[width=\fig_width]{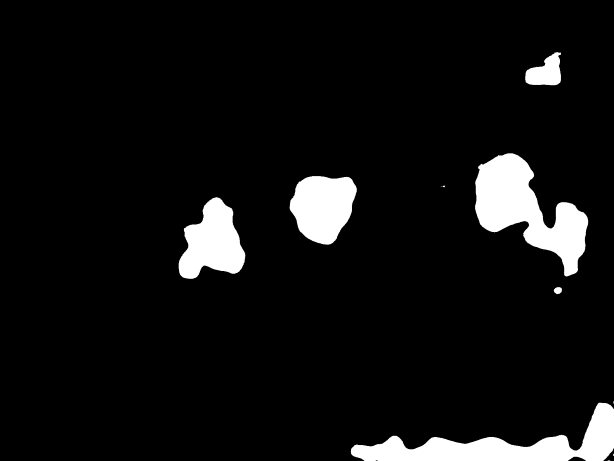}
		\end{overpic}&
		\begin{overpic}[width=\fig_width]{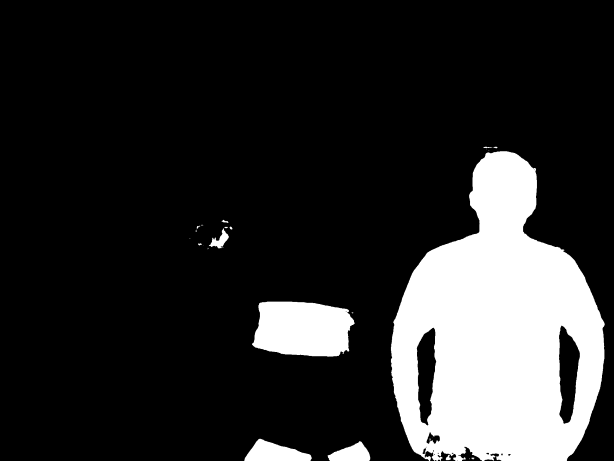}
		\end{overpic}&
		\begin{overpic}[width=\fig_width]{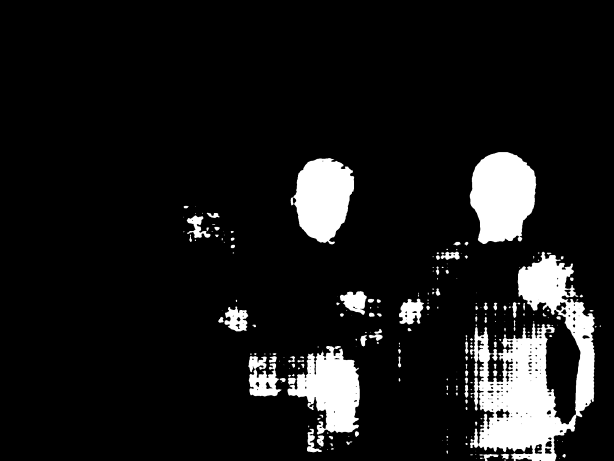}
		\end{overpic}&
		\begin{overpic}[width=\fig_width]{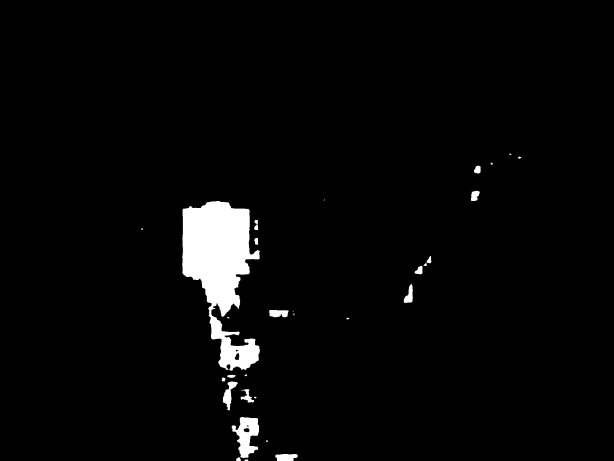}
		\end{overpic}&
		\begin{overpic}[width=\fig_width]{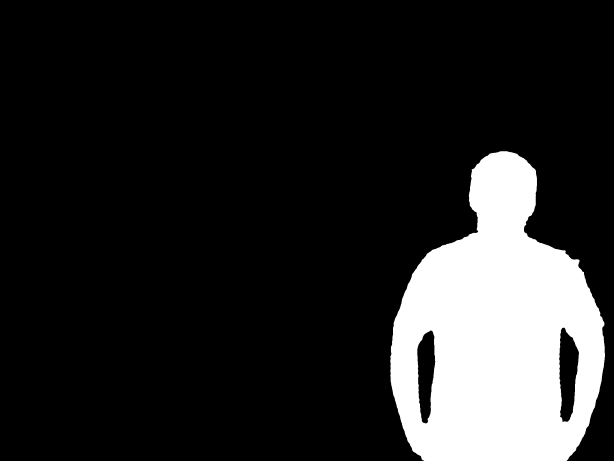}
		\end{overpic}&
		\begin{overpic}[width=\fig_width]{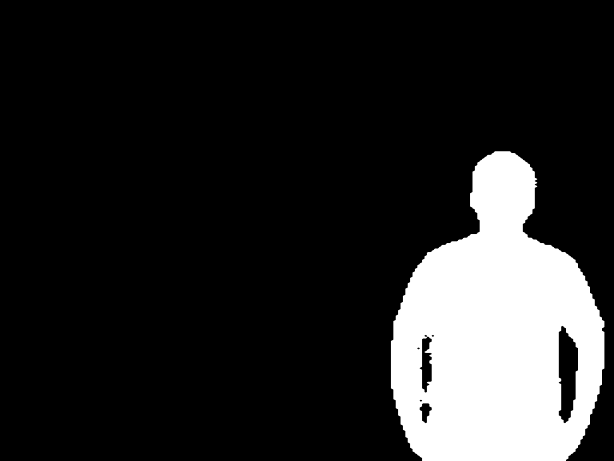}
		\end{overpic}
		\\
		\begin{overpic}[width=\fig_width]{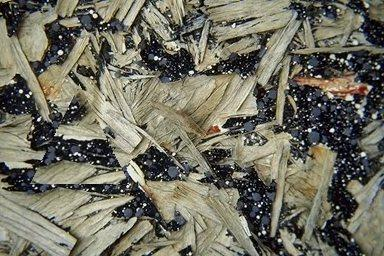}
		\end{overpic}&
		\begin{overpic}[width=\fig_width]{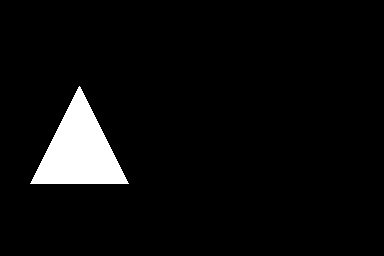}
		\end{overpic}&
		\begin{overpic}[width=\fig_width]{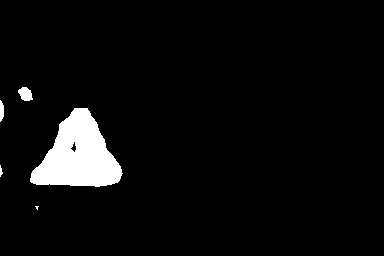}
		\end{overpic}&
		\begin{overpic}[width=\fig_width]{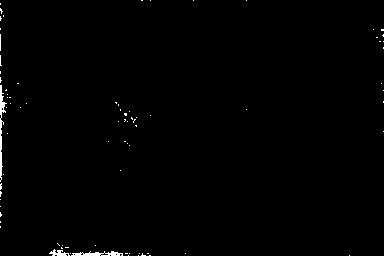}
		\end{overpic}&
		\begin{overpic}[width=\fig_width]{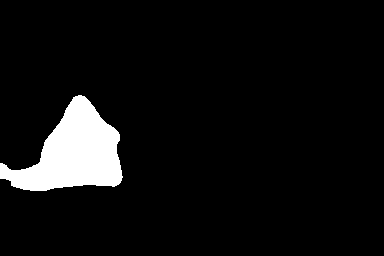}
		\end{overpic}&
		\begin{overpic}[width=\fig_width]{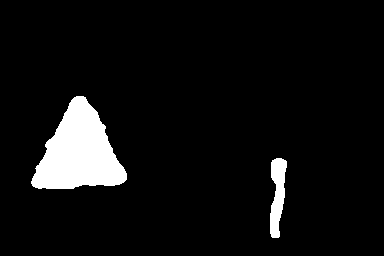}
		\end{overpic}&
		\begin{overpic}[width=\fig_width]{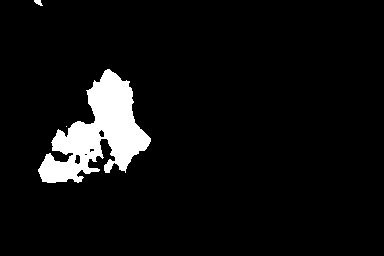}
		\end{overpic}&
		\begin{overpic}[width=\fig_width]{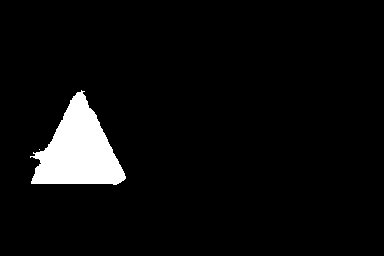}
		\end{overpic}&
		\begin{overpic}[width=\fig_width]{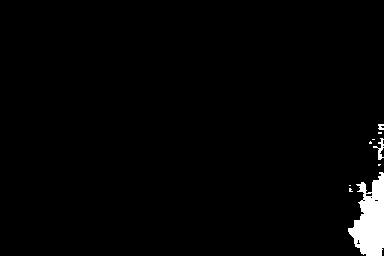}
		\end{overpic}&
		\begin{overpic}[width=\fig_width]{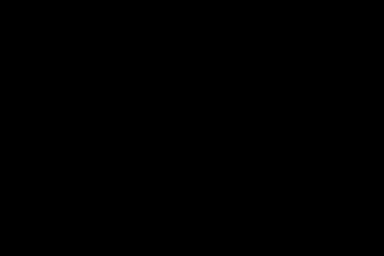}
		\end{overpic}&
		\begin{overpic}[width=\fig_width]{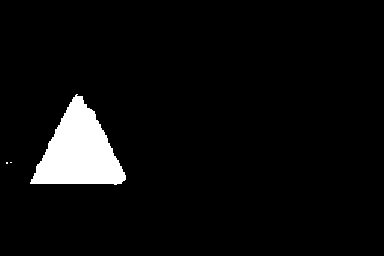}
		\end{overpic}    
		
		\\
		\begin{overpic}[width=\fig_width]{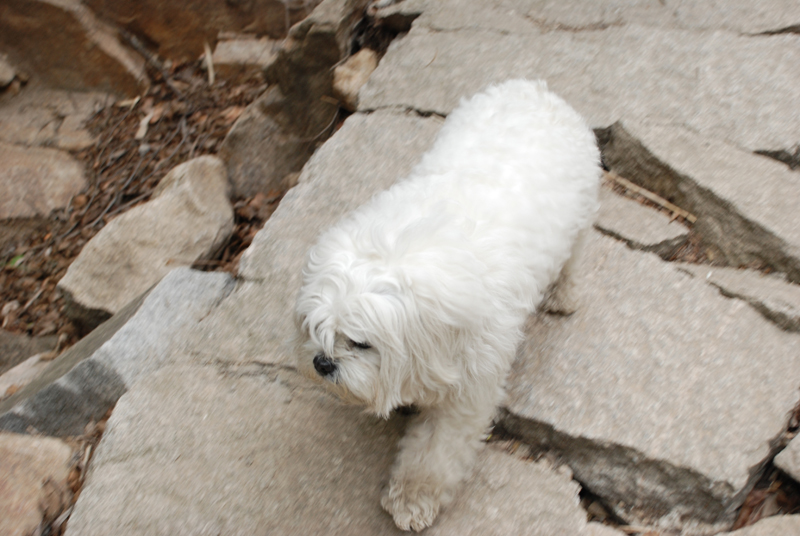}
		\end{overpic}&
		\begin{overpic}[width=\fig_width]{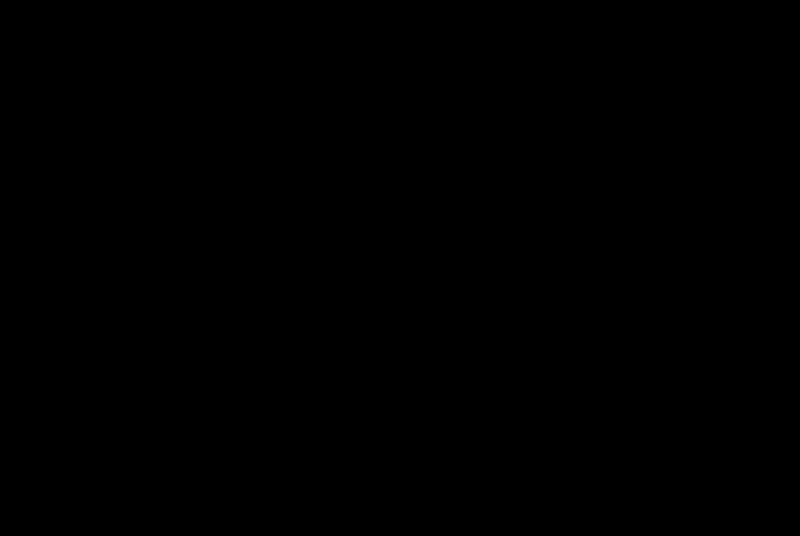}
		\end{overpic}&
		\begin{overpic}[width=\fig_width]{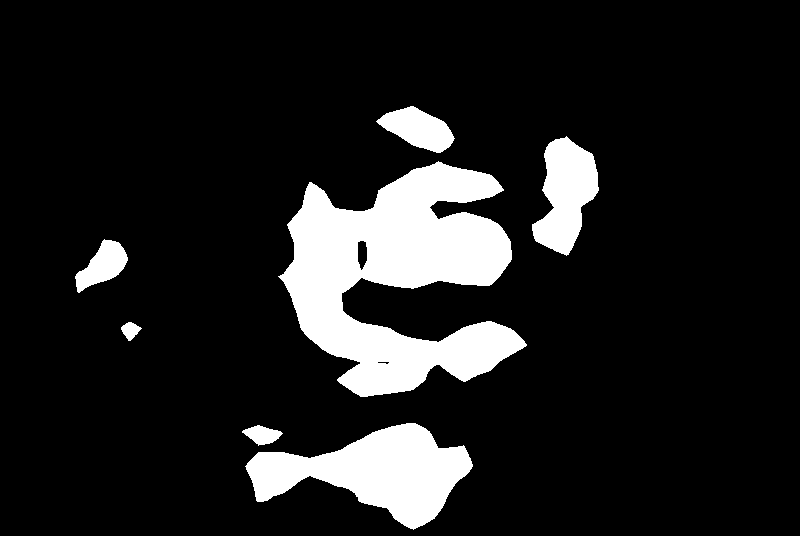}
		\end{overpic}&
		\begin{overpic}[width=\fig_width]{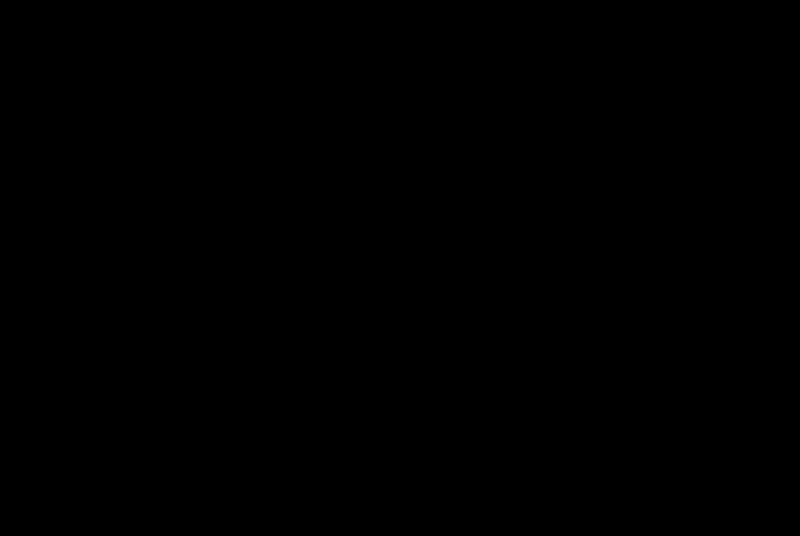}
		\end{overpic}&
		\begin{overpic}[width=\fig_width]{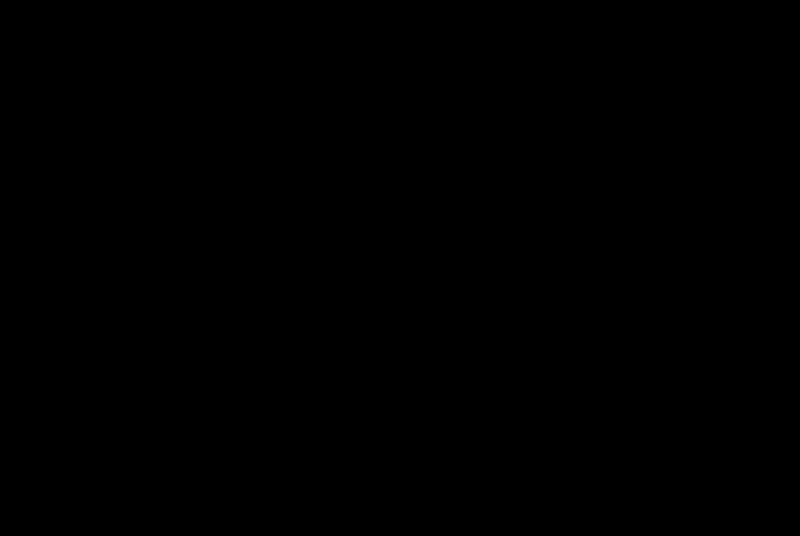}
		\end{overpic}&
		\begin{overpic}[width=\fig_width]{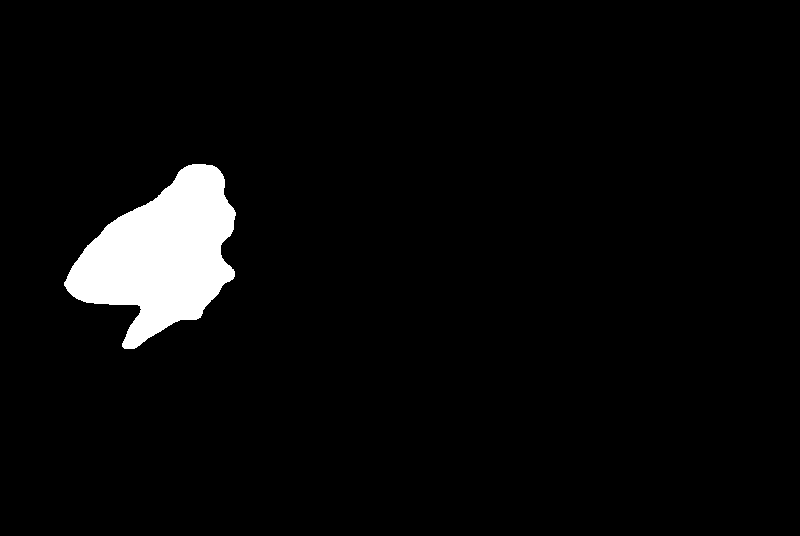}
		\end{overpic}&
		\begin{overpic}[width=\fig_width]{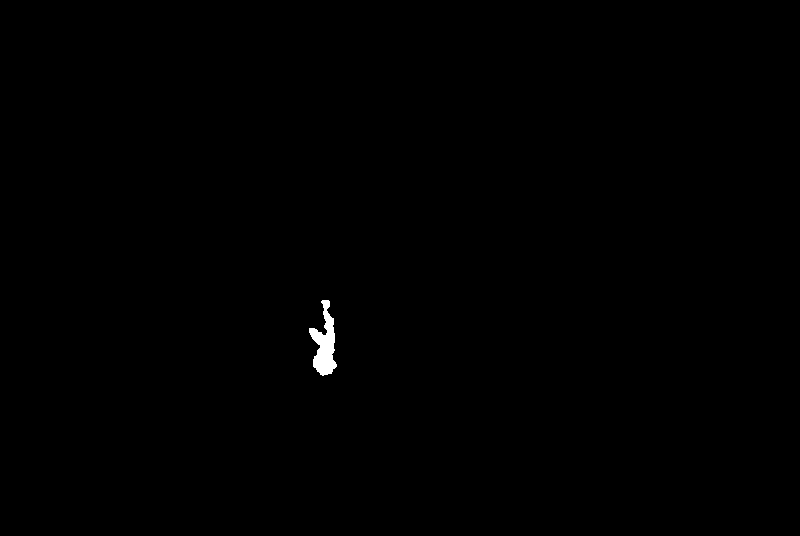}
		\end{overpic}&
		\begin{overpic}[width=\fig_width]{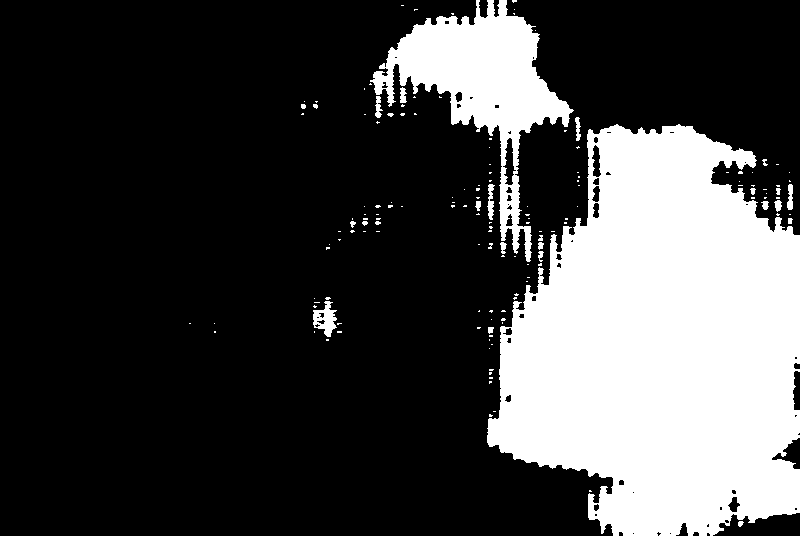}
		\end{overpic}&
		\begin{overpic}[width=\fig_width]{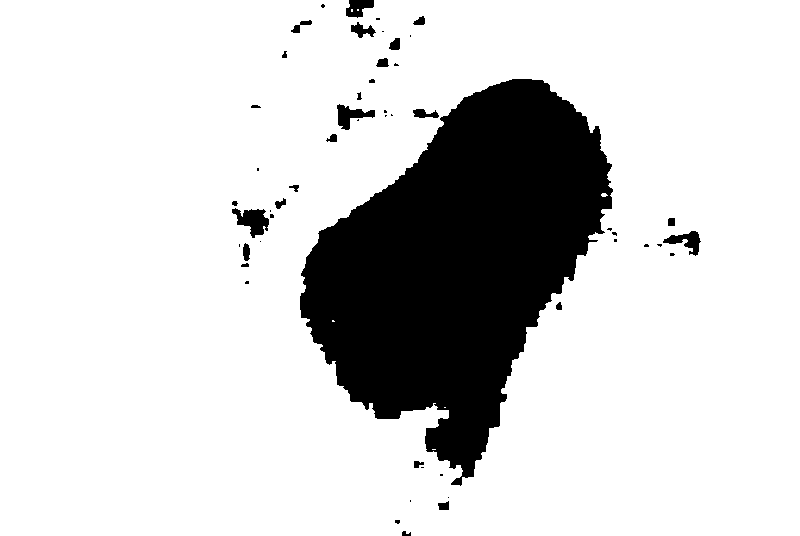}
		\end{overpic}&
		\begin{overpic}[width=\fig_width]{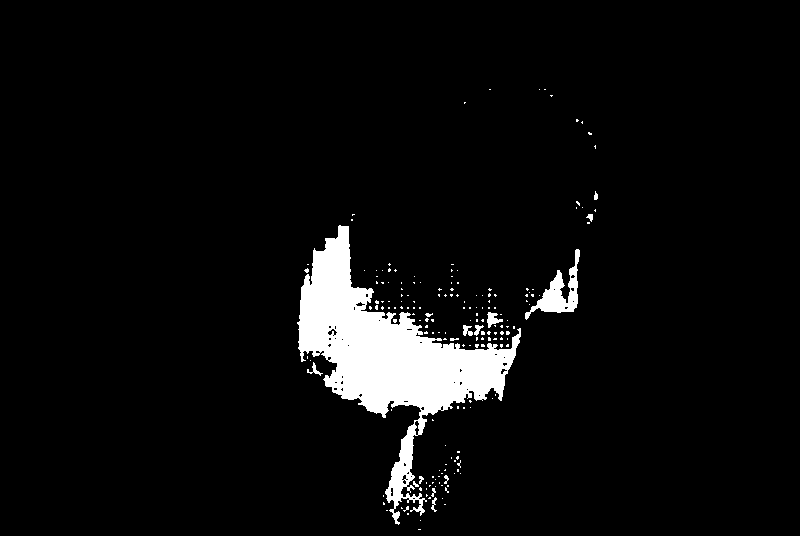}
		\end{overpic}&
		\begin{overpic}[width=\fig_width]{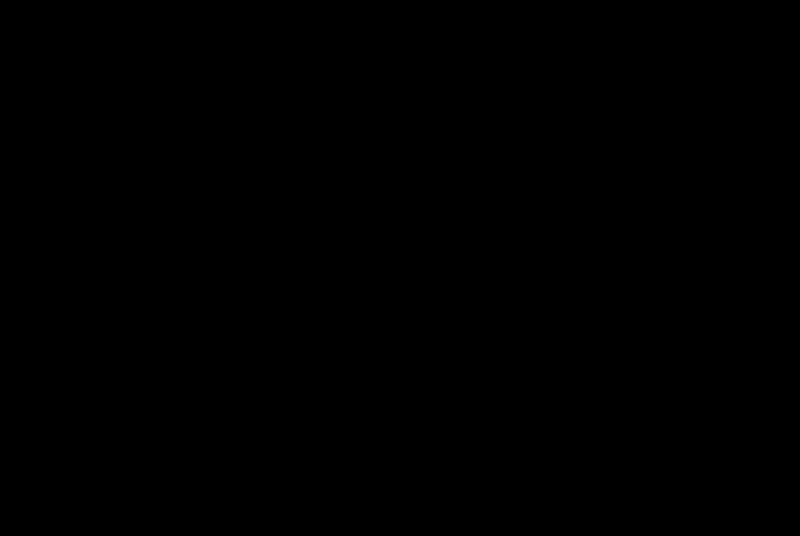}
		\end{overpic}   
		\\
		Image & GT & MVSS-Net++ & IF-OSN & CAT-Net v2 & CoDE & TruFor & AutoSAM & SAFIRE & FakeShield & \textbf{Ours*}\\
	\end{tabular}
	\caption{Comparison of predicted forgery mask on real and forged images. White color denotes the forged regions and black color denotes the authentic regions.}
	\label{fig:image_forgery_localization} 
\end{figure*}
\begin{table*}[t]
	\centering
	\caption{Image-level forgery detection performance (ACC). First and second ranking are highlighted in bold and underline, respectively.}
	\tabcolsep=0.09cm
	\begin{tabular}{r | c c c c c c c c c c | c}
		\hline
		IFDL Method & CASIAv1+ & MISD & Columbia & DSO-1 & Coverage & NIST & CocoGlide & IPM15k & ACDSee & In-the-wild & Average \\
		\hline
		MVSS-Net++ & 0.655 & 0.590 & 0.725 & 0.525 & 0.585 & 0.479 & 0.552 & 0.817 & 0.514 & 0.915 & 0.635\\
		IF-OSN & 0.647 & 0.667 & 0.521 & 0.500 & 0.510 & 0.418 & 0.567 & 0.920 & 0.484 & 1.000 & 0.623 \\
		CAT-Net v2 & 0.843 & 0.796 & 0.805 & 0.525 & 0.645 & 0.500 & \underline{0.586} & 0.735 & 0.691 & 0.811 & 0.694\\
		CoDE & 0.551 & 0.355 & 0.507 & 0.500 & 0.505 & 0.401 & 0.505 & \underline{0.994} & 0.489 & \textbf{1.000} & 0.581\\
		TruFor & 0.811 & \textbf{0.957} & \underline{0.983} & \textbf{0.930} & \underline{0.685} & \underline{0.670} & \textbf{0.639} & 0.528 & \underline{0.730} & 0.657 & \underline{0.759} \\
		AutoSAM & 0.588 & 0.450 & 0.606 & 0.500 & 0.500 & 0.445 & 0.508 & 0.963 & 0.482 & 0.995 & 0.604\\
		SAFIRE & 0.535 & 0.335 & 0.498 & 0.500 & 0.505 & 0.392 & 0.501 & \textbf{0.997} & 0.483 & \underline{0.998} & 0.574 \\
		FakeShield & \textbf{0.925} & 0.853 & 0.846 & 0.610 & 0.500 & 0.629 & 0.502 & 0.587 & 0.522 & 0.662 & 0.664\\
		\rowcolor{gray!20}  
		$\textbf{Ours}^{\star}$ & \underline{0.894} & \underline{0.936} & \textbf{0.989} & \underline{0.820} & \textbf{0.730} & \textbf{0.776} & 0.585 & 0.768 & \textbf{0.770} & 0.905 & \textbf{0.817}\\
		\hline     
	\end{tabular}
	\label{tab:image_forgery_detection}
\end{table*}
\begin{table*}[t]
	\centering
	\caption{Pixel-level forgery localization performance (F1). First and second ranking are highlighted in bold and underline, respectively.}
	\tabcolsep=0.09cm
	\begin{tabular}{r | c c c c c c c c c c | c}
		\hline
		IFDL Method & CASIAv1+ & MISD & Columbia & DSO-1 & Coverage & NIST & CocoGlide & IPM15k & ACDSee & In-the-wild & Average \\
		\hline
		MVSS-Net++ & 0.532 & 0.692 & 0.737 & 0.334 & 0.516 & 0.366 & 0.543 & 0.421 & 0.368 & 0.421 & 0.486 \\
		IF-OSN & 0.554 & 0.732 & 0.748 & 0.443 & 0.339 & 0.311 & 0.459 & 0.465 & 0.377 & 0.569 & 0.500 \\
		CAT-Net v2 & 0.728 & 0.522 & 0.849 & 0.576 & 0.386 & 0.370 & 0.467 & 0.403 & \textbf{0.719} & 0.506 & 0.553 \\
		CoDE & 0.734 & \underline{0.780} & 0.920 & 0.442 & 0.502 & 0.464 & 0.580 & 0.592 & 0.501 & 0.603 & 0.612 \\
		TruFor & 0.713 & 0.691 & 0.848 & \textbf{0.910} & 0.576 & 0.441 & 0.518 & 0.598 & 0.628 & \underline{0.667} & \underline{0.659} \\
		AutoSAM & \underline{0.740} & 0.763 & 0.916 & 0.464 & 0.628 & 0.436 & 0.545 & 0.640 & 0.479 & 0.651 & 0.626\\
		SAFIRE & 0.394 & 0.664 & \textbf{0.967} & 0.558 & \underline{0.635} & \underline{0.489} & \textbf{0.629} & 0.410 & 0.411 & 0.566 & 0.572 \\
		FakeShield & 0.617 & 0.543 & 0.810 & 0.569 & 0.431 & 0.414 & \underline{0.592} & \underline{0.739} & 0.456 & 0.707 & 0.587 \\
		\rowcolor{gray!20}  
		$\textbf{Ours}^{\star}$ & \textbf{0.815} & \textbf{0.792} & \underline{0.959} & \underline{0.851} & \textbf{0.781} & \textbf{0.613} & 0.588 & \textbf{0.755} & \underline{0.645} & \textbf{0.734} & \textbf{0.753}\\
		\hline    
	\end{tabular}
	\label{tab:image_forgery_localization}
\end{table*}
The proposed ForensicsSAM is implemented using PyTorch. All input images are resized to 1024$\times$1024 for both training and inference. We adopt SAM-H as the backbone of ForensicsSAM, and use the AdamW~\cite{loshchilov2017decoupled} optimizer throughout all training stages. All training stages are conducted on a Tesla A100 GPU cluster using PyTorch DistributedDataParallel (DDP) for multi-GPU acceleration.

In the first training stage, we use CASIAv2, IMD20, FantasticReality, and TamperedCR as the training set, including only clean images (both real and forged). The learning rate and batch size are set to 1e-4 and 24, respectively. This stage is trained for $e_1 = 5$ epochs using 6 Tesla A100 GPUs, which takes approximately 17 hours.

In the second and third stages, we only use CASIAv2 and IMD20 as the training set. During these stages, adversarial examples are generated using the MI-FGSM method, with perturbation bounds $\varphi \in \{2, 4, 8\}$. The learning rate remains 1e-4 for both stages. The batch sizes are set to 64 and 24 for the second and third stages, respectively. Stage two is trained for $e_2 = 10$ epochs using 2 Tesla A100 GPUs, requiring about 2 hours. Stage three runs for $e_3 = 20$ epochs using 6 Tesla A100 GPUs, taking approximately 9 hours.

\subsection{Comparison Results}
This section evaluates the performance of image-level forgery detection and pixel-level forgery localization of our method under non-adversarial settings and compares it with existing state-of-the-art methods.

\begin{figure*}[t]
	\centering
	\def\fig_width{2.cm}
	\setlength\tabcolsep{0.7mm}
	\subfloat[Before Adversarial Attack]{
		\begin{tabular}{c c c c}
			AutoSAM & SAFIRE & FakeShield & \textbf{Ours*}
			\\
			\begin{overpic}[width=\fig_width]{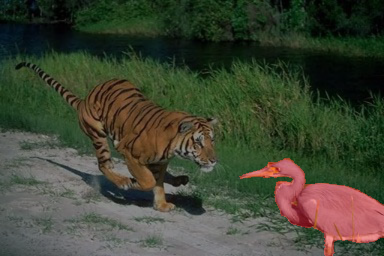}
				\put(-20, -40){\rotatebox{90}{MI-FGSM}}
				\put(-20, -205){\rotatebox{90}{PGN}}
				\put(-20, -380){\rotatebox{90}{BSR}}
				\put(-20, -570){\rotatebox{90}{UMI-GRAT}}
			\end{overpic}&
			\begin{overpic}[width=\fig_width]{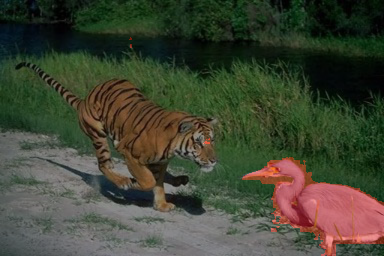}
			\end{overpic}&
			\begin{overpic}[width=\fig_width]{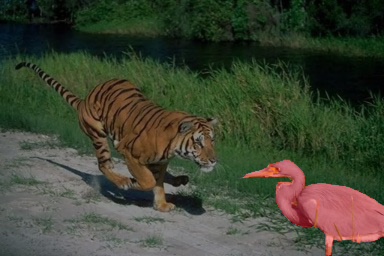}
			\end{overpic}&
			\begin{overpic}[width=\fig_width]{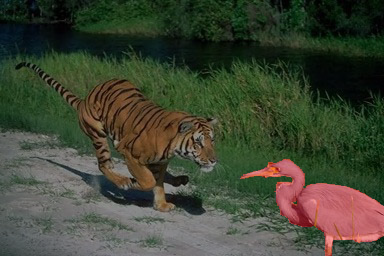}
			\end{overpic}
			\\
			\begin{overpic}[width=\fig_width]{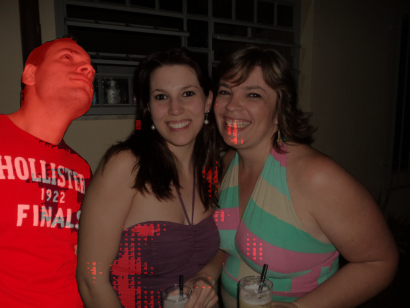}
			\end{overpic}&
			\begin{overpic}[width=\fig_width]{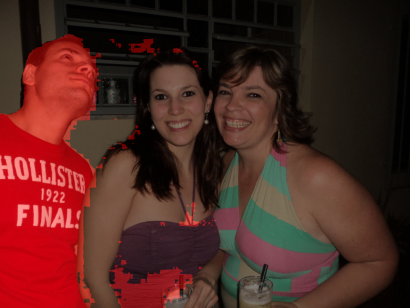}
			\end{overpic}&
			\begin{overpic}[width=\fig_width]{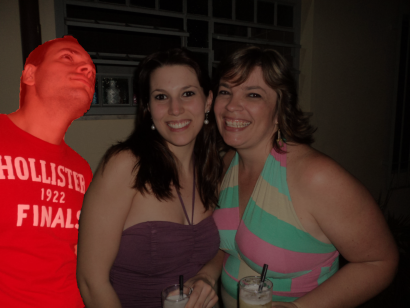}
			\end{overpic}&
			\begin{overpic}[width=\fig_width]{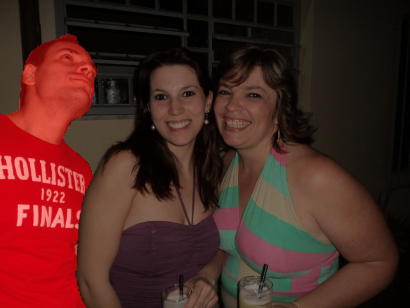}
			\end{overpic}
			\\
			\\
			\begin{overpic}[width=\fig_width]{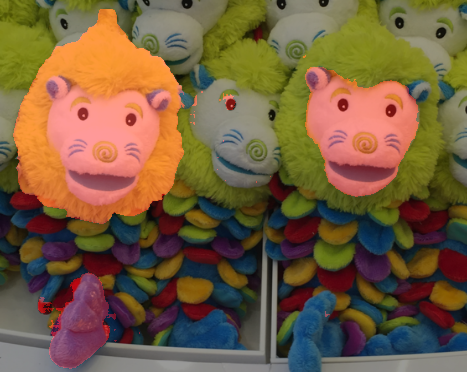}
			\end{overpic}&
			\begin{overpic}[width=\fig_width]{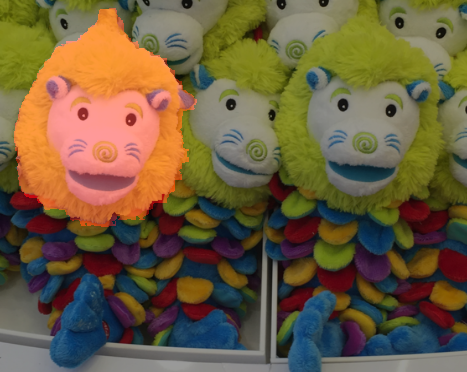}
			\end{overpic}&
			\begin{overpic}[width=\fig_width]{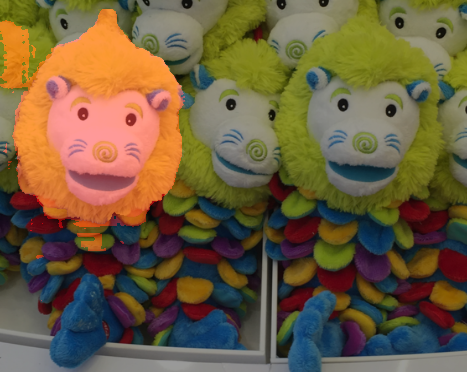}
			\end{overpic}&
			\begin{overpic}[width=\fig_width]{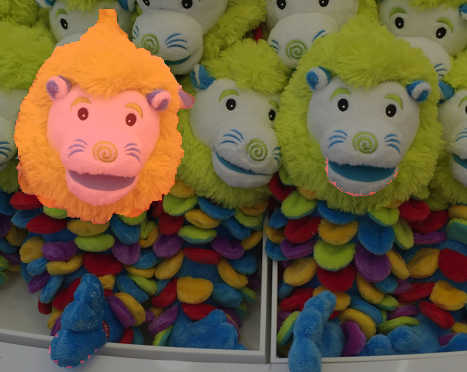}
			\end{overpic}
			\\
			\begin{overpic}[width=\fig_width]{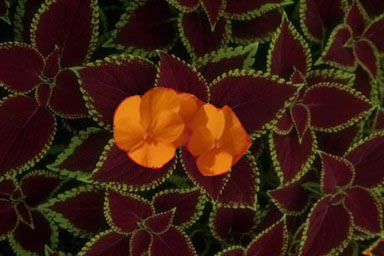}
			\end{overpic}&
			\begin{overpic}[width=\fig_width]{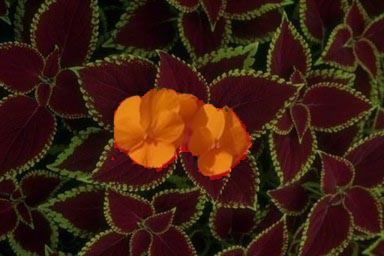}
			\end{overpic}&
			\begin{overpic}[width=\fig_width]{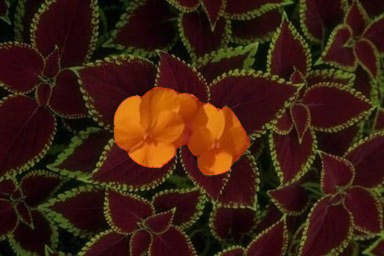}
			\end{overpic}&
			\begin{overpic}[width=\fig_width]{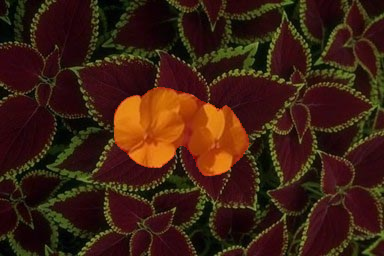}
			\end{overpic}
			\\
			\\
			\begin{overpic}[width=\fig_width]{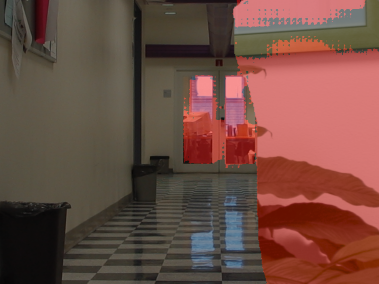}
			\end{overpic}&
			\begin{overpic}[width=\fig_width]{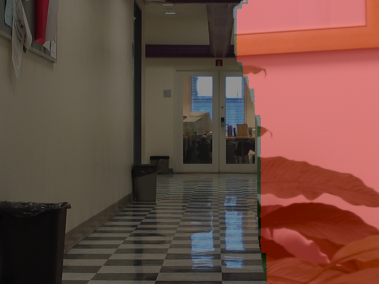}
			\end{overpic}&
			\begin{overpic}[width=\fig_width]{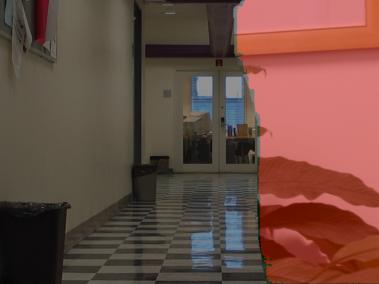}
			\end{overpic}&
			\begin{overpic}[width=\fig_width]{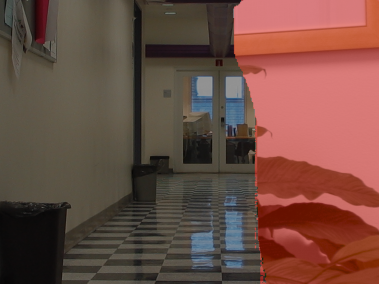}
			\end{overpic}
			\\
			\begin{overpic}[width=\fig_width]{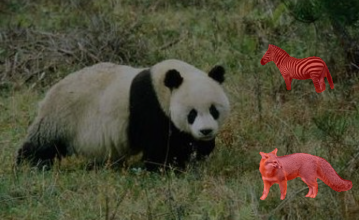}
			\end{overpic}&
			\begin{overpic}[width=\fig_width]{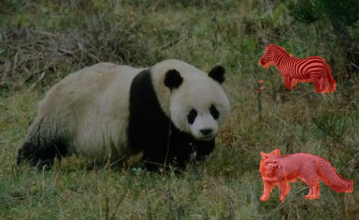}
			\end{overpic}&
			\begin{overpic}[width=\fig_width]{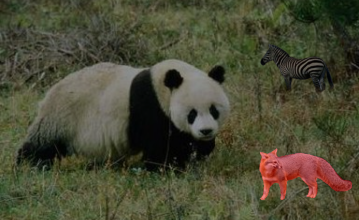}
			\end{overpic}&
			\begin{overpic}[width=\fig_width]{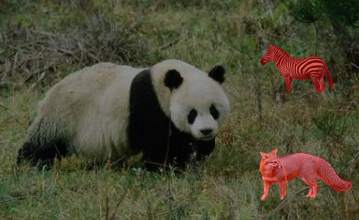}
			\end{overpic}
			\\
			\\
			\begin{overpic}[width=\fig_width]{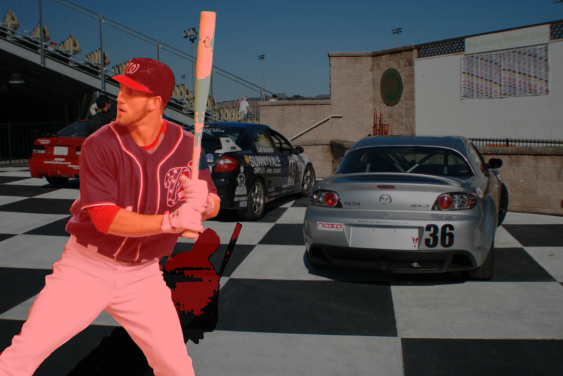}
			\end{overpic}&
			\begin{overpic}[width=\fig_width]{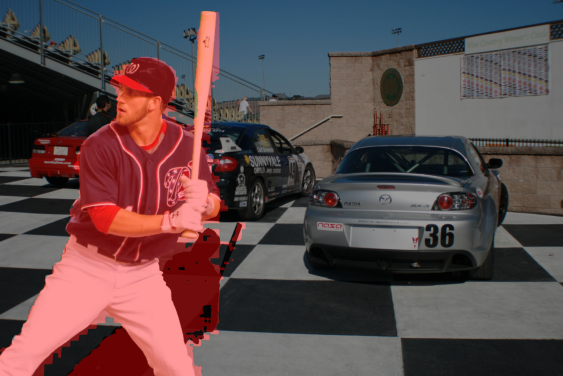}
			\end{overpic}&
			\begin{overpic}[width=\fig_width]{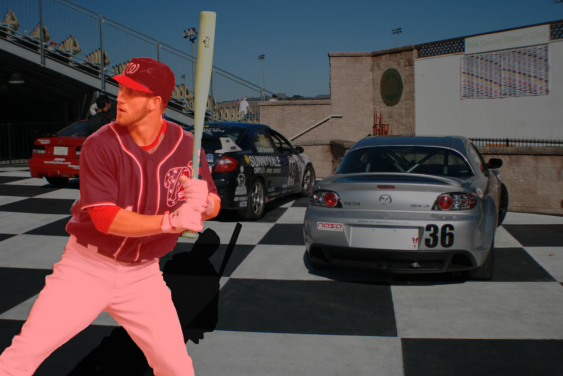}
			\end{overpic}&
			\begin{overpic}[width=\fig_width]{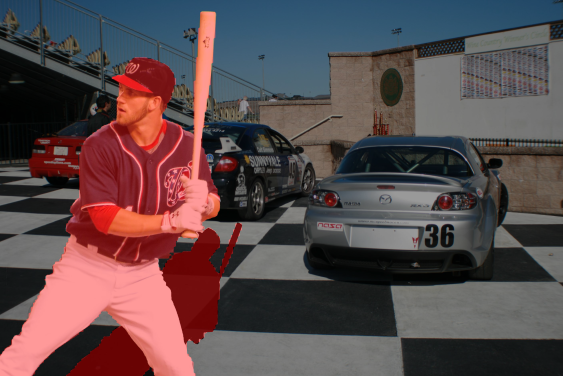}
			\end{overpic}
			\\
			\begin{overpic}[width=\fig_width]{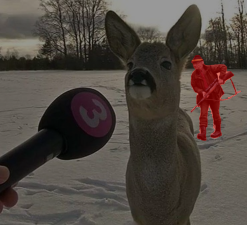}
			\end{overpic}&
			\begin{overpic}[width=\fig_width]{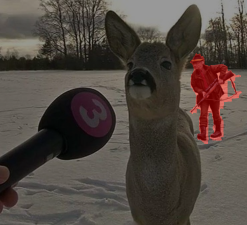}
			\end{overpic}&
			\begin{overpic}[width=\fig_width]{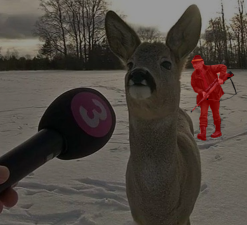}
			\end{overpic}&
			\begin{overpic}[width=\fig_width]{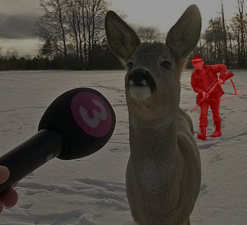}
			\end{overpic}
			\\
		\end{tabular}
	}
	\subfloat[After Adversarial Attack]{
		\begin{tabular}{c c c c}
			AutoSAM & SAFIRE & FakeShield & \textbf{Ours*}
			\\
			\begin{overpic}[width=\fig_width]{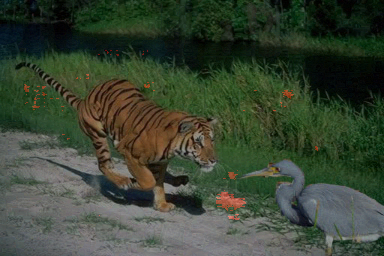}
			\end{overpic}&
			\begin{overpic}[width=\fig_width]{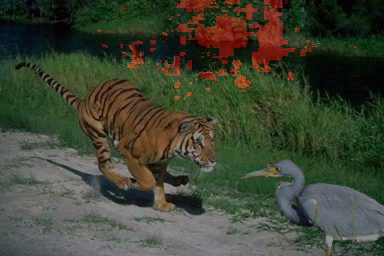}
			\end{overpic}&
			\begin{overpic}[width=\fig_width]{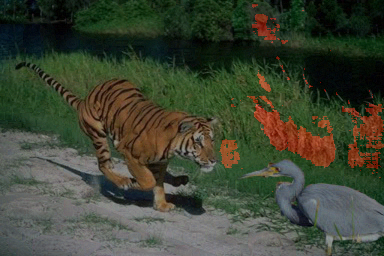}
			\end{overpic}&
			\begin{overpic}[width=\fig_width]{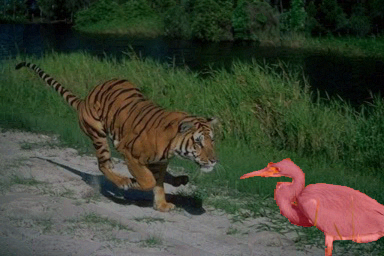}
			\end{overpic}
			\\
			\begin{overpic}[width=\fig_width]{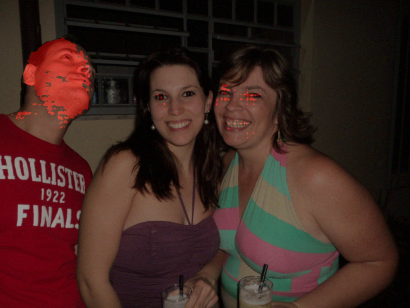}
			\end{overpic}&
			\begin{overpic}[width=\fig_width]{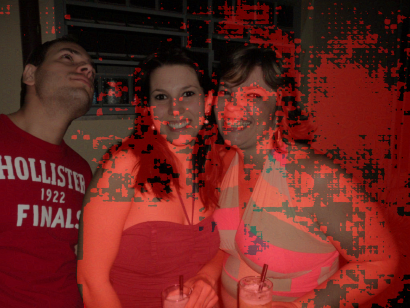}
			\end{overpic}&
			\begin{overpic}[width=\fig_width]{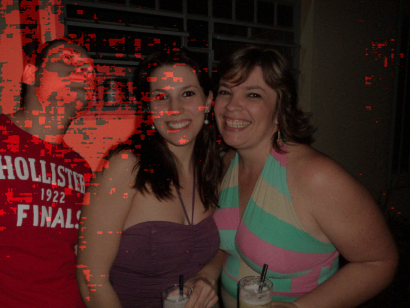}
			\end{overpic}&
			\begin{overpic}[width=\fig_width]{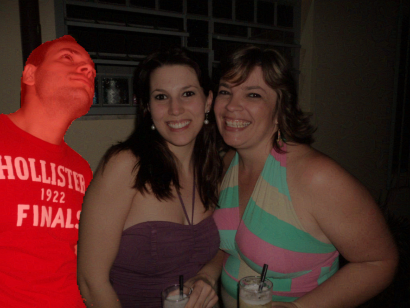}
			\end{overpic}
			\\
			\\
			\begin{overpic}[width=\fig_width]{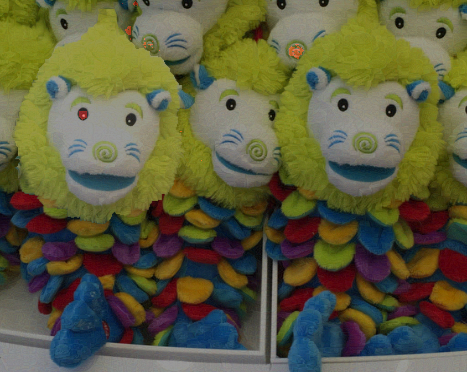}
			\end{overpic}&
			\begin{overpic}[width=\fig_width]{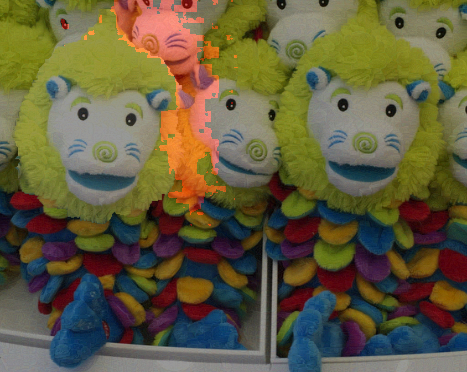}
			\end{overpic}&
			\begin{overpic}[width=\fig_width]{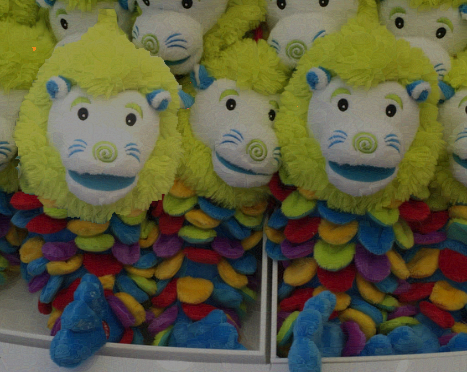}
			\end{overpic}&
			\begin{overpic}[width=\fig_width]{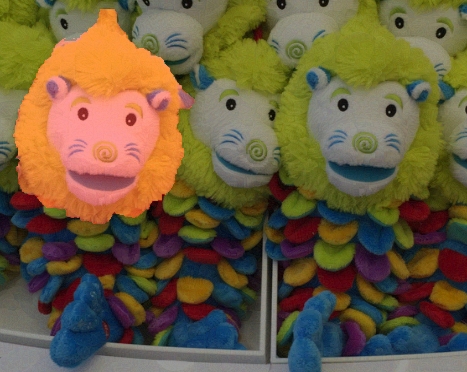}
			\end{overpic}
			\\
			\begin{overpic}[width=\fig_width]{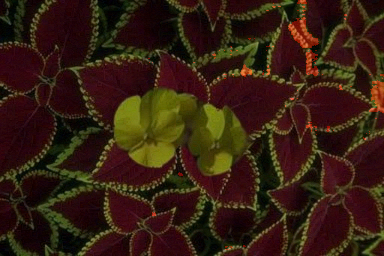}
			\end{overpic}&
			\begin{overpic}[width=\fig_width]{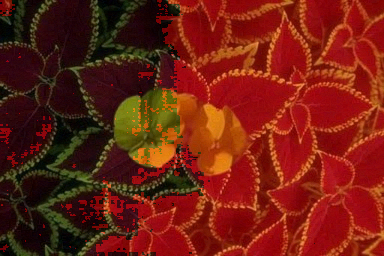}
			\end{overpic}&
			\begin{overpic}[width=\fig_width]{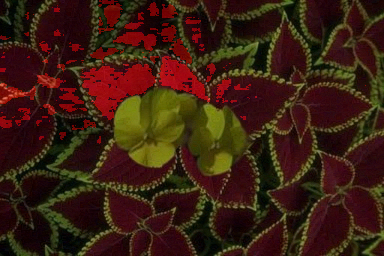}
			\end{overpic}&
			\begin{overpic}[width=\fig_width]{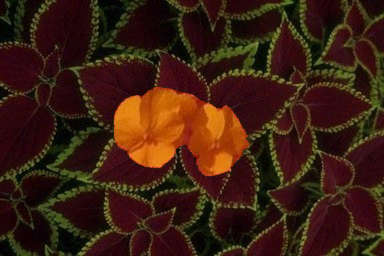}
			\end{overpic}
			\\
			\\
			\begin{overpic}[width=\fig_width]{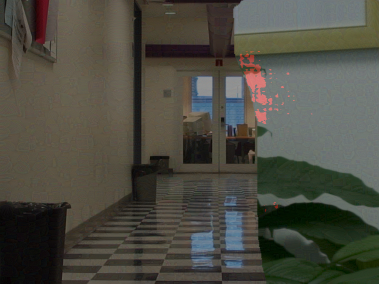}
			\end{overpic}&
			\begin{overpic}[width=\fig_width]{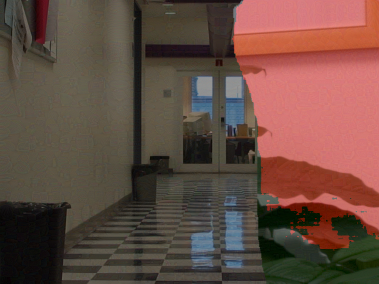}
			\end{overpic}&
			\begin{overpic}[width=\fig_width]{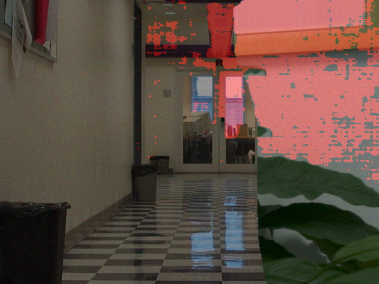}
			\end{overpic}&
			\begin{overpic}[width=\fig_width]{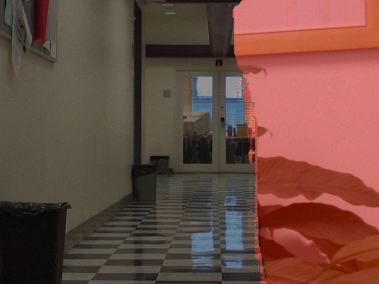}
			\end{overpic}
			\\
			\begin{overpic}[width=\fig_width]{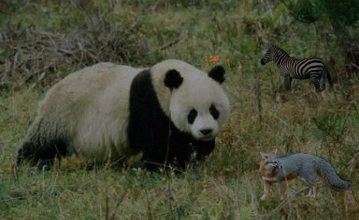}
			\end{overpic}&
			\begin{overpic}[width=\fig_width]{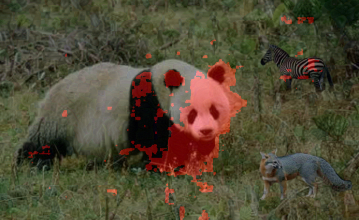}
			\end{overpic}&
			\begin{overpic}[width=\fig_width]{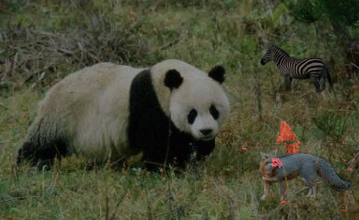}
			\end{overpic}&
			\begin{overpic}[width=\fig_width]{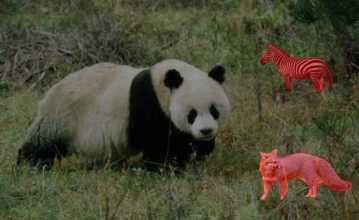}
			\end{overpic}
			\\
			\\
			\begin{overpic}[width=\fig_width]{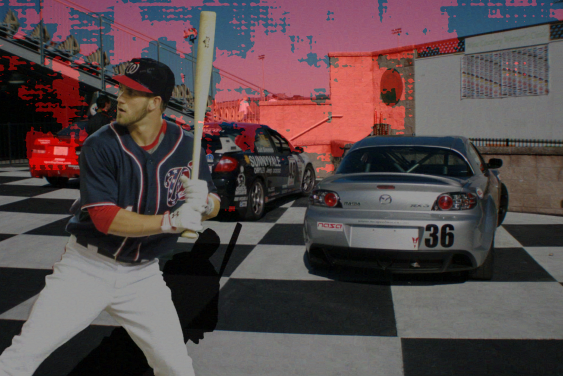}
			\end{overpic}&
			\begin{overpic}[width=\fig_width]{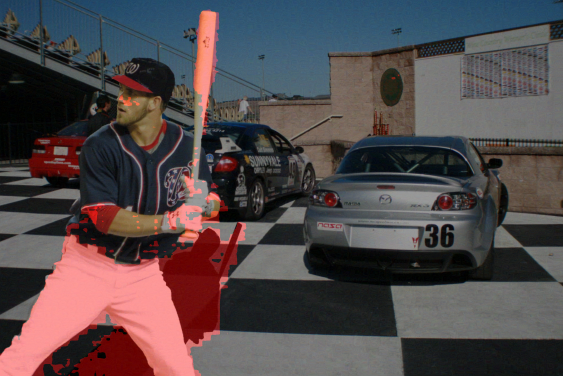}
			\end{overpic}&
			\begin{overpic}[width=\fig_width]{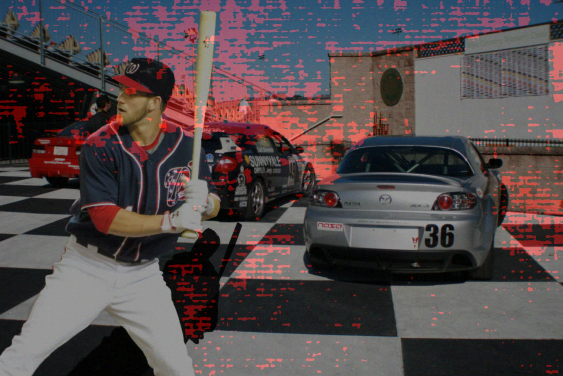}
			\end{overpic}&
			\begin{overpic}[width=\fig_width]{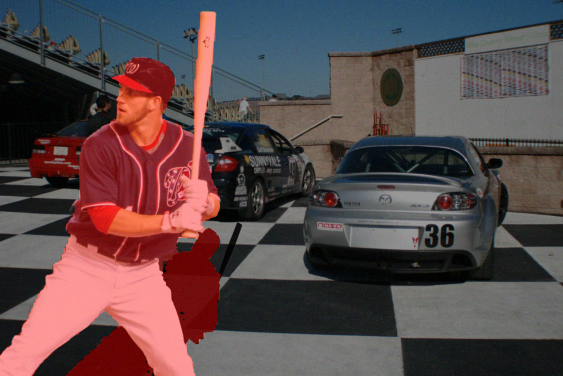}
			\end{overpic}
			\\
			\begin{overpic}[width=\fig_width]{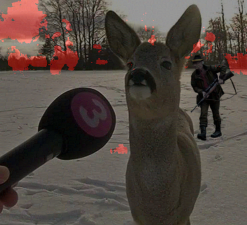}
			\end{overpic}&
			\begin{overpic}[width=\fig_width]{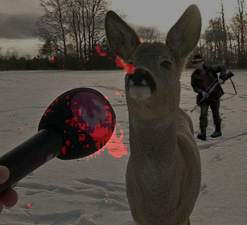}
			\end{overpic}&
			\begin{overpic}[width=\fig_width]{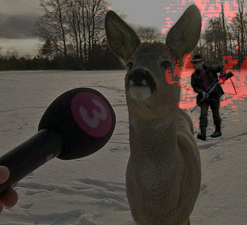}
			\end{overpic}&
			\begin{overpic}[width=\fig_width]{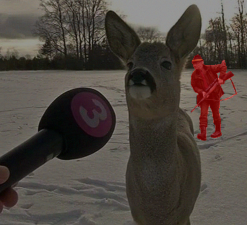}
			\end{overpic}
			\\
		\end{tabular}
	}
	\caption{Comparison of forgery localization before and after  adversarial attack. Red denotes the predicted forged regions.}
	\label{fig:image_forgery_localization_attack} 
\end{figure*}
\begin{table*}[t]
	\centering
	\caption{Image-level Adversary detection performance (ACC) of the adversary detector.}
	\tabcolsep=0.09cm
	\renewcommand{\arraystretch}{0.85}
	\begin{tabular}{c r | c c c c c c c c c c | c}
		\hline
		\multicolumn{2}{c|}{Attack Method} & CASIAv1+ & MISD & Columbia & DSO-1 & Coverage & NIST & CocoGlide & IPM15k & ACDSee & In-the-wild & Average \\
		\hline
		& w/o Attack & 1.000 & 1.000 & 1.000 & 1.000 & 1.000 & 0.988 & 1.000 & 0.999 & 0.996 & 0.995 & 0.998\\
		\hline
		\multirow{4}{*}{\rotatebox{90}{$\varphi=8$}} 
		& MI-FGSM & 1.000 & 1.000 & 1.000 & 1.000 & 1.000 & 1.000 & 1.000 & 1.000 & 1.000 & 1.000 & 1.000 \\
		& PGN & 1.000 & 1.000 & 1.000 & 1.000 & 1.000 & 1.000 & 1.000 & 1.000 & 1.000 & 1.000 & 1.000 \\
		& BSR & 1.000 & 1.000 & 1.000 & 1.000 & 1.000 & 1.000 & 0.999 & 1.000 & 1.000 & 1.000 & 0.999  \\
		& UMI-GRAT & 1.000 & 1.000 & 1.000 & 1.000 & 1.000 & 0.989 & 1.000 & 0.999 & 1.000 & 0.995 & 0.998 \\
		\hline
		\multirow{4}{*}{\rotatebox{90}{$\varphi=12$}} 
		& MI-FGSM & 1.000 & 1.000 & 1.000 & 1.000 & 1.000 & 1.000 & 1.000 & 1.000 & 1.000 & 1.000 & 1.000  \\
		& PGN & 1.000 & 1.000 & 1.000 & 1.000 & 1.000 & 1.000 & 1.000 & 1.000 & 1.000 & 1.000 & 1.000 \\
		& BSR & 1.000 & 1.000 & 1.000 & 1.000 & 1.000 & 1.000 & 0.999 & 1.000 & 1.000 & 1.000 & 0.999 \\
		& UMI-GRAT & 1.000 & 1.000 & 1.000 & 1.000 & 1.000 & 0.998 & 1.000 & 0.999 & 1.000 & 0.995 & 0.999 \\
		\hline    
	\end{tabular}
	\label{tab:adversary_detection}
\end{table*}
\begin{table*}[!t]
	\centering
	\caption{The detailed image forgery localization performance under various intensities and types of adversarial attacks.}
	\tabcolsep=0.05cm
	\renewcommand{\arraystretch}{0.85}
	\subfloat[AutoSAM]{
		\begin{tabular}{c r | c c c c c c c c c c | c c}
			\hline
			\multicolumn{2}{c|}{Attack Method} & CASIAv1+ & MISD & Columbia & DSO-1 & Coverage & NIST & CocoGlide & IPM15k & ACDSee & In-the-wild & \multicolumn{2}{c}{Average} \\
			\hline
			& w/o Attack & 0.740 & 0.763 & 0.916 & 0.464 & 0.628 & 0.436 & 0.545 & 0.640 & 0.479 & 0.651 &  \multicolumn{2}{c}{0.626} \\
			\hline
			\multirow{4}{*}{\rotatebox{90}{$\varphi=8$}} 
			& MI-FGSM & 0.157 & 0.314 & 0.420 & 0.239 & 0.199 & 0.140 & 0.355 & 0.225 & 0.148 & 0.284 & 0.248 & $\downarrow$60.4\% \\
			& PGN & 0.157 & 0.316 & 0.411 & 0.240 & 0.200 & 0.136 & 0.361 & 0.213 & 0.142 & 0.282 & 0.246 & $\downarrow$60.7\%\\
			& BSR & 0.166 & 0.322 & 0.440 & 0.240 & 0.204 & 0.149 & 0.358 & 0.236 & 0.163 & 0.284 & 0.256 & $\downarrow$59.1\%\\
			& UMI-GRAT & 0.159 & 0.321 & 0.441 & 0.241 & 0.205 & 0.183 & 0.359 & 0.228 & 0.165 & 0.293 & 0.260 & $\downarrow$58.5\% \\
			\hline
			\multirow{4}{*}{\rotatebox{90}{$\varphi=12$}} 
			& MI-FGSM & 0.154 & 0.315 & 0.414 & 0.238 & 0.204 & 0.126 & 0.356 & 0.227 & 0.135 & 0.275 & 0.244 & $\downarrow$61.0\%\\
			& PGN & 0.155 & 0.315 & 0.411 & 0.239 & 0.200 & 0.131 & 0.359 & 0.227 & 0.134 & 0.276 & 0.245 & $\downarrow$60.9\%\\
			& BSR & 0.154 & 0.317 & 0.408 & 0.237 & 0.204 & 0.127 & 0.356 & 0.227 & 0.134 & 0.280 & 0.244 &  $\downarrow$61.0\%\\
			& UMI-GRAT & 0.155 & 0.315 & 0.416 & 0.243 & 0.199 & 0.141 & 0.359 & 0.231 & 0.141 & 0.283 & 0.248 & $\downarrow$60.4\%\\
			\hline    
		\end{tabular}
	}
	
	\subfloat[SAFIRE]{
		\begin{tabular}{c r | c c c c c c c c c c | c c}
			\hline
			\multicolumn{2}{c|}{Attack Method} & CASIAv1+ & MISD & Columbia & DSO-1 & Coverage & NIST & CocoGlide & IPM15k & ACDSee & In-the-wild & \multicolumn{2}{c}{Average} \\
			\hline
			& w/o Attack & 0.394 & 0.664 & 0.967 & 0.558 & 0.635 & 0.489 & 0.629 & 0.410 & 0.411 & 0.566 & \multicolumn{2}{c}{0.572} \\
			\hline
			\multirow{4}{*}{\rotatebox{90}{$\varphi=8$}} 
			& MI-FGSM & 0.226 & 0.375 & 0.889 & 0.322 & 0.303 & 0.375 & 0.399 & 0.285 & 0.261 & 0.382 & 0.382 &  $\downarrow$33.2\%\\
			& PGN & 0.215 & 0.370 & 0.856 & 0.299 & 0.279 & 0.365 & 0.398 & 0.276 & 0.255 & 0.392 & 0.371 &  $\downarrow$35.1\%\\
			& BSR & 0.205 & 0.358 & 0.762 & 0.305 & 0.283 & 0.323 & 0.390 & 0.285 & 0.222 & 0.392 & 0.353 &  $\downarrow$38.3\%\\
			& UMI-GRAT & 0.232 & 0.397 & 0.905 & 0.300 & 0.280 & 0.400 & 0.412 & 0.287 & 0.282 & 0.406 & 0.390 &  $\downarrow$31.8\%\\
			\hline
			\multirow{4}{*}{\rotatebox{90}{$\varphi=12$}} 
			& MI-FGSM & 0.213 & 0.370 & 0.850 & 0.299 & 0.288 & 0.359 & 0.396 & 0.294 & 0.249 & 0.404 & 0.372 &  $\downarrow$35.0\%\\
			& PGN & 0.201 & 0.353 & 0.742 & 0.274 & 0.249 & 0.315 & 0.395 & 0.271 & 0.224 & 0.352 & 0.338 & $\downarrow$40.9\%\\
			& BSR & 0.194 & 0.345 & 0.619 & 0.293 & 0.242 & 0.272 & 0.384 & 0.270 & 0.203 & 0.338 & 0.316 & $\downarrow$44.8\%\\
			& UMI-GRAT & 0.221 & 0.377 & 0.839 & 0.305 & 0.264 & 0.358 & 0.397 & 0.292 & 0.252 & 0.382 & 0.369 &  $\downarrow$35.5\%\\
			\hline    
		\end{tabular}
	}
	
	\subfloat[FakeShield]{
		\begin{tabular}{c r | c c c c c c c c c c | c c}
			\hline
			\multicolumn{2}{c|}{Attack Method} & CASIAv1+ & MISD & Columbia & DSO-1 & Coverage & NIST & CocoGlide & IPM15k & ACDSee & In-the-wild & \multicolumn{2}{c}{Average} \\
			\hline
			& w/o Attack & 0.617 & 0.543 & 0.810 & 0.569 & 0.431 & 0.414 & 0.592 & 0.739 & 0.456 & 0.707 & \multicolumn{2}{c}{0.587} \\
			\hline
			\multirow{4}{*}{\rotatebox{90}{$\varphi=8$}} 
			& MI-FGSM & 0.185 & 0.333 & 0.457 & 0.255 & 0.204 & 0.137 & 0.359 & 0.242 & 0.142 & 0.291 & 0.261 &  $\downarrow$55.5\%\\
			& PGN & 0.182 & 0.332 & 0.460 & 0.257 & 0.207 & 0.141 & 0.358 & 0.244 & 0.149 & 0.293 & 0.262 &  $\downarrow$55.4\%\\
			& BSR & 0.195 & 0.331 & 0.491 & 0.287 & 0.215 & 0.179 & 0.360 & 0.286 & 0.170 & 0.328 & 0.284 &  $\downarrow$51.6\%\\
			& UMI-GRAT & 0.177 & 0.344 & 0.466 & 0.268 & 0.205 & 0.140 & 0.366 & 0.258 & 0.140 & 0.317 & 0.268 &  $\downarrow$54.3\%\\
			\hline
			\multirow{4}{*}{\rotatebox{90}{$\varphi=12$}} 
			& MI-FGSM & 0.175 & 0.335 & 0.454 & 0.258 & 0.204 & 0.128 & 0.364 & 0.246 & 0.140 & 0.292 & 0.260 &  $\downarrow$55.7\% \\
			& PGN & 0.182 & 0.342 & 0.461 & 0.271 & 0.212 & 0.133 & 0.369 & 0.253 & 0.147 & 0.296 & 0.267 &  $\downarrow$54.5\%\\
			& BSR & 0.174 & 0.327 & 0.441 & 0.271 & 0.206 & 0.137 & 0.364 & 0.246 & 0.150 & 0.300 & 0.262 &  $\downarrow$55.4\%\\
			& UMI-GRAT & 0.180 & 0.356 & 0.486 & 0.270 & 0.209 & 0.134 & 0.378 & 0.268 & 0.142 & 0.320 & 0.274 &  $\downarrow$53.3\%\\
			\hline    
		\end{tabular}
	}
	
	\subfloat[Ours*]{
		\begin{tabular}{c r | c c c c c c c c c c | >{\columncolor{gray!20}} c >{\columncolor{gray!20}} c}
			\hline
			\multicolumn{2}{c|}{Attack Method} & CASIAv1+ & MISD & Columbia & DSO-1 & Coverage & NIST & CocoGlide & IPM15k & ACDSee & In-the-wild & \multicolumn{2}{c}{Average} \\
			\hline
			& w/o Attack & 0.815 & 0.792 & 0.959 & 0.851 & 0.781 & 0.613 & 0.588 & 0.755 & 0.645 & 0.734 & \multicolumn{2}{>{\columncolor{gray!20}} c}{0.753}\\
			\hline
			\multirow{4}{*}{\rotatebox{90}{$\varphi=8$}} & MI-FGSM & 0.710 & 0.737 & 0.965 & 0.655 & 0.552 & 0.501 & 0.537 & 0.698 & 0.554 & 0.687 & 0.660 & $\downarrow$12.4\%\\
			& PGN & 0.697 & 0.740 & 0.970 & 0.634 & 0.506 & 0.488 & 0.538 & 0.699 & 0.532 & 0.673 & 0.648 & $\downarrow$13.9\%\\
			& BSR & 0.686 & 0.733 & 0.968 & 0.647 & 0.546 & 0.507 & 0.532 & 0.683 & 0.544 & 0.685 & 0.653 & $\downarrow$13.3\%\\
			& UMI-GRAT & 0.732 & 0.724 & 0.968 & 0.600 & 0.532 & 0.505 & 0.541 & 0.696 & 0.561 & 0.672 & 0.653 & $\downarrow$13.3\%\\
			\hline
			\multirow{4}{*}{\rotatebox{90}{$\varphi=12$}} & MI-FGSM & 0.688 & 0.725 & 0.970 & 0.603 & 0.491 & 0.502 & 0.531 & 0.686 & 0.548 & 0.675 & 0.642 & $\downarrow$14.7\%\\
			& PGN & 0.669 & 0.719 & 0.966 & 0.621 & 0.484 & 0.479 & 0.534 & 0.674 & 0.498 & 0.659 & 0.630 & $\downarrow$16.3\%\\
			& BSR & 0.650 & 0.708 & 0.976 & 0.610 & 0.526 & 0.492 & 0.519 & 0.652 & 0.529 & 0.661 & 0.632 & $\downarrow$16.1\%\\
			& UMI-GRAT & 0.684 & 0.690 & 0.960 & 0.581 & 0.500 & 0.490 & 0.529 & 0.690 & 0.519 & 0.654 & 0.630 & $\downarrow$16.3\%\\
			\hline    
		\end{tabular}
	}
	\label{tab:robustness_adversarial_attacks}
\end{table*}
\subsubsection{Forgery Detection.}
As shown in Table~\ref{tab:image_forgery_detection}, our method achieves the highest average detection performance across both real and forged images. Notably, most existing methods are primarily optimized using forged images and tend to overlook the accurate identification of real images. This often leads to over-sensitive behavior, where a substantial portion of real images are misclassified as forged, resulting in high false positive rates. In contrast, our method explicitly incorporates real image detection during training, thereby learning a more balanced and discriminative decision boundary. Compared to the second-best method, we achieve a 0.058 improvement in average ACC, demonstrating enhanced discriminability and reliability in distinguishing between real and forged images.

Fig.~\ref{fig:detection_results} presents the qualitative results of both image-level forgery detection and pixel-level localization for real and forged images. To correctly identify a real image, the predicted mask is expected to be entirely zero across all pixels. However, we can observe that this is difficult to achieve in practice when relying solely on the predicted masks. Even small, spurious activations in the mask can trigger false alarms, resulting in real images being misclassified as forged. By introducing a forgery detector, the image-level forgery scores become significantly more reliable, effectively suppressing the false alarms.

\subsubsection{Forgery Localization.}
As shown in Table~\ref{tab:image_forgery_localization}, our method achieves the best or second-best forgery localization performance across all evaluated datasets. On average, it yields a 0.094 improvement in F1 score over the second-best method, underscoring its superior generalizability across diverse image sources and forgery types.

Fig.~\ref{fig:image_forgery_localization} further provides qualitative comparisons, where we visualize the predicted forgery masks for both real and forged images across different IFDL methods. Existing approaches often exhibit excessive activations or fail to accurately localize forged regions, leading to false alarms and incomplete prediction. In contrast, our method produces sharper and more accurate localization results, with significantly fewer false alarms on real images and finer delineation of forged regions in forged images. 

These results highlight the effectiveness of our method in capturing subtle forgery traces and establishing precise pixel-level decision boundaries, which are critical for trustworthy image forensics in real-world scenarios.

\subsection{Evaluation on the Resistance to Adversarial Attacks}
\begin{table}[t]
	\centering
	\caption{The detection performance of adversary detector under the adversarial attack on itself.}
	\newcommand{\xmark}{\ding{55}}
	\begin{tabular}{c | c c c c}
		\hline
		Input Image & Det. Adv. & Loc \\
		\hline
		$X_{c}$ & 0.998 & 0.753 \\
		$X_{a}$ & 1.000 & 0.660 \\
		$X_{c} + N_{ad}$ & 0.999 & 0.672 \\
		$X_{a} + N_{ad}$ & 0.999 & 0.649 \\
		\hline    
	\end{tabular}
	\label{tab:robustness_adversary_detector}
\end{table}
In this section, we compare the resistance to adversarial attacks of our method with that of existing PERT-based SAM variants under four types of adversarial attacks, including MI-FGSM, PGN, BSR, and UMI-GRAT.
\subsubsection{Adversary Detection}
At first, we evaluate the image-level detection performance of the proposed adversary detector. We regard both the real and forged images as clean images, while those with adversarial noise added are regarded as adversarial images. As shown in Table~\ref{tab:adversary_detection}, the adversary detector achieves near-perfect accuracy on clean images (w/o Attack) across all 10 datasets, with an average ACC of 0.998, confirming its ability to avoid false positives. When applied to adversarial images generated by four different attack methods under two perturbation bounds, the adversary detector consistently maintains high accuracy. Specifically, under $\varphi=8$, the average ACC remains above 0.998 for all attacks, and reaches 1.000 in many datasets. With stronger perturbation ($\varphi=12$), the adversary detector performs even better, possibly because higher-strength adversarial attacks introduce more salient artifacts that are easier to capture. Overall, the proposed adversary detector exhibits excellent image-level discrimination capability between clean and adversarial images, enabling reliable gating control of the adversary experts.
\subsubsection{Adversary Localization}
Secondly, we evaluate the ability to restore pixel-level forgery localization performance against adversarial attacks. As shown in Table~\ref{tab:robustness_adversarial_attacks}, the localization performance (measured by F1) of existing methods drops significantly under four adversarial attacks. To be specific, the average F1 score of AutoSAM drops by at least 0.378 and up to 0.400 across different attack settings. SAFIRE shows a performance drop ranging from 0.291 to 0.337, while FakeShield degrades by 0.292 at minimum and 0.336 at most. In contrast, our method demonstrates significantly stronger robustness, with F1 score dropping by only 0.095 at minimum and 0.123 at most across all attack settings. This corresponds to preserving between 84\% and 87\% of the original performance, clearly outperforming other methods. Remarkably, although our model is trained only with MI-FGSM, it maintains consistently high robustness across diverse adversarial attack methods, highlighting its strong generalization ability. We further provide visualizations in Fig.~\ref{fig:image_forgery_localization_attack}, where our method consistently and accurately localizes forged regions despite the presence of strong adversarial noise. In contrast, other methods yield fragmented, or misleading predictions, highlighting their vulnerability to adversarial attacks.

What's more, it is worth noting that the adversary detector is also derived from an open-source model. This raises a potential concern: attackers might adopt similar strategies to mislead the adversary detector and thereby bypass the adversary experts.  To address this concern, we generate adversarial noise $N_{ad}$ by attacking the upstream adversary detector, and add it to both clean images ($X_c$) and adversarial images ($X_a$) to evaluate whether $N_{ad}$ can mislead the well-trained adversary detector. Table~\ref{tab:robustness_adversary_detector} presents four types of input image, where only $X_{c}$ is labeled as non-adversarial image. Despite the presence of $N_{\text{ad}}$, the adversary detector still identifies almost all adversarial images, with accuracy close to 1.000. The localization performance also remains robust. The F1 stays above 0.64, retaining over 85\% of the original performance (F1 = 0.753). These results demonstrate the reliability of our framework, even when the adversary detector itself is targeted.

\begin{table}[t]
	\centering
	\caption{Ablation study for proposed three core components: forgery experts, adversary detector, and adversary experts. `+' indicates cumulative addition of components on top of the previous configuration.}
	\tabcolsep=0.06cm
	\renewcommand{\arraystretch}{1.3}
	\begin{tabular}{c r | c c | c c | c c}
		\hline
		& \multirow{2}*{Variant} & \multicolumn{2}{c|}{Loc.} & \multicolumn{2}{c|}{Det. Forged} & \multicolumn{2}{c}{Det. Adv.}\\
		& & $X_{c}$ & $X_{a}$ & $X_{c}$ & $X_{a}$ & $X_{c}$ & $X_{a}$\\
		\hline
		& Baseline & 0.510 & 0.255 & - & - & - & -\\
		& + Forgery Experts & 0.753 & 0.246 & 0.817 & 0.144 & - & -\\
		& + Adversary Detector & 0.753 & 0.246 & 0.817 & 0.144 & 0.998 & 1.000\\
		& + Adversary Experts & 0.753 & 0.660 & 0.817 & 1.000 & 0.998 & 1.000\\
		\hline
	\end{tabular}
	\label{tab:ablation_study}
\end{table}
\begin{figure}[t]
	\centering
	\def\fig_width{1.5cm}
	\setlength\tabcolsep{0.7mm}
	\begin{tabular}{c c c c c}
		\begin{overpic}[width=\fig_width]{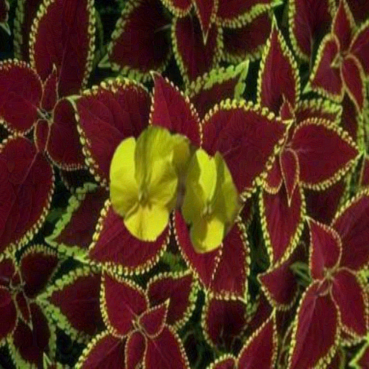}
		\end{overpic}&
		\begin{overpic}[width=\fig_width]{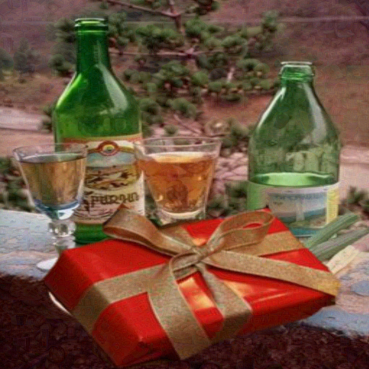}
		\end{overpic}&
		\begin{overpic}[width=\fig_width]{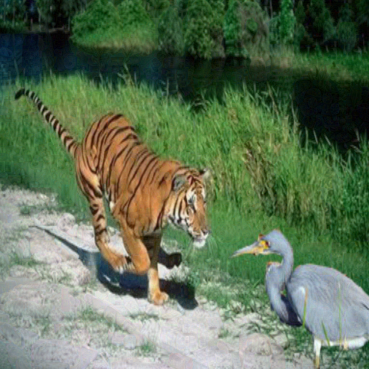}
		\end{overpic}&
		\begin{overpic}[width=\fig_width]{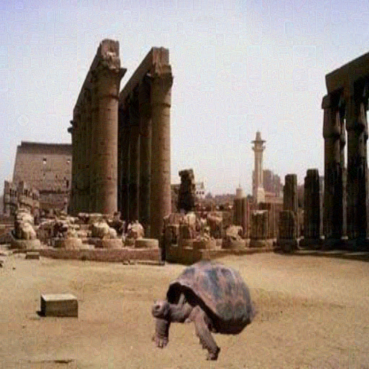}
		\end{overpic}
		\begin{overpic}[width=\fig_width]{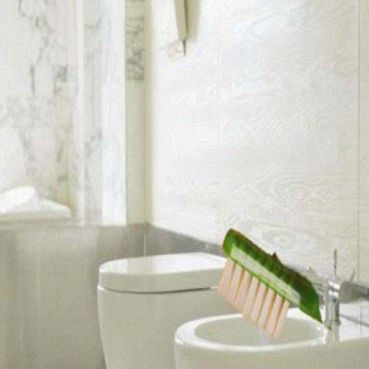}
		\end{overpic}
		\\
		\begin{overpic}[width=\fig_width]{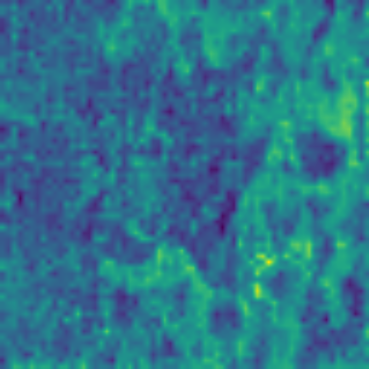}
		\end{overpic}&
		\begin{overpic}[width=\fig_width]{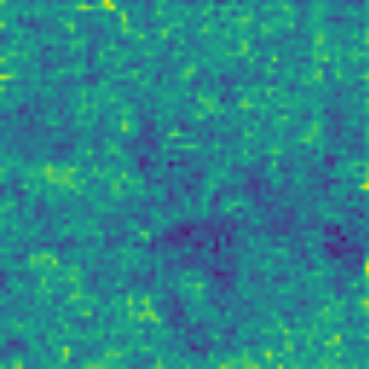}
		\end{overpic}&
		\begin{overpic}[width=\fig_width]{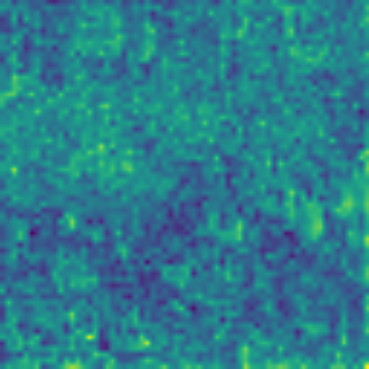}
		\end{overpic}&
		\begin{overpic}[width=\fig_width]{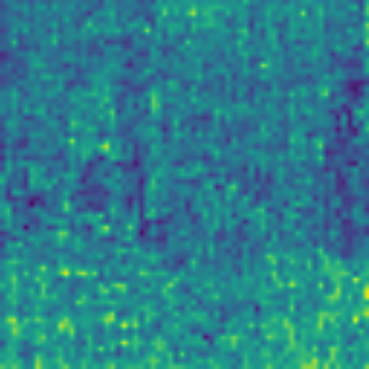}
		\end{overpic}
		\begin{overpic}[width=\fig_width]{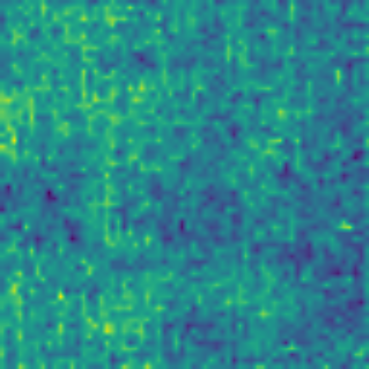}
		\end{overpic}
		\\
		\begin{overpic}[width=\fig_width]{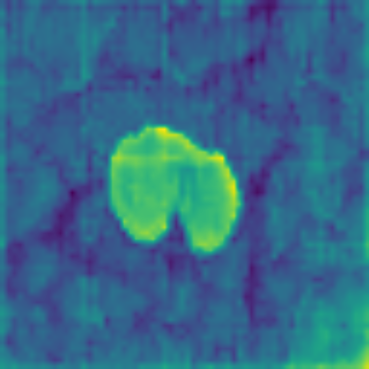}
		\end{overpic}&
		\begin{overpic}[width=\fig_width]{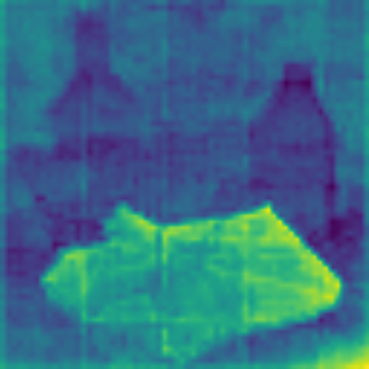}
		\end{overpic}&
		\begin{overpic}[width=\fig_width]{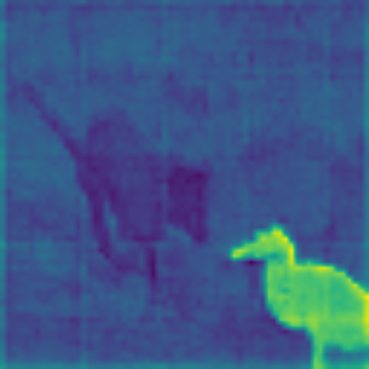}
		\end{overpic}&
		\begin{overpic}[width=\fig_width]{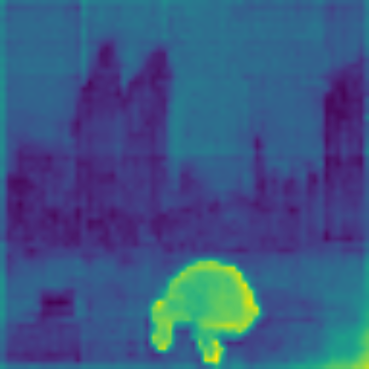}
		\end{overpic}
		\begin{overpic}[width=\fig_width]{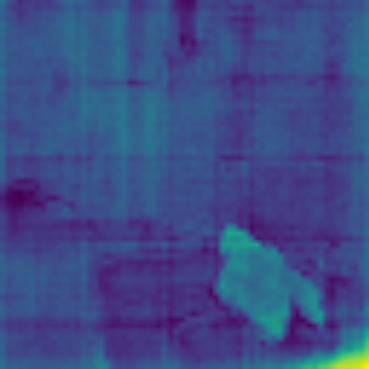}
		\end{overpic}
	\end{tabular}
	\caption{The visualization of feature maps (image embedding) of adversarial images. The first row presents the adversarial images. The second and third rows are the feature maps without and with adversary experts, respectively.}
	\label{fig:ablation_feature}
\end{figure}
\subsection{Ablation Study}
In this section, we present the ablation study evaluating the contribution of the three core components: forgery experts, adversary detector, and adversary experts. Starting from the baseline SAM model (with only the mask decoder fine-tuned), we incrementally add each component to evaluate their respective contributions.

As shown in Table~\ref{tab:ablation_study}, firstly, adding the forgery experts significantly enhances forgery localization performance on clean images, increasing the F1 from 0.510 to 0.753. It also enables effective image-level forgery detection, achieving ACC = 0.817. However, localization performance on adversarial images remains low at this stage (F1 = 0.246).

Secondly, incorporating the adversary detector does not directly improve forgery localization or detection performance, but achieves highly reliable adversarial discrimination, with an average ACC close to 1.000. Importantly, the strong discriminative capability of the adversary detector enables precise activation of the adversary experts, ensuring that clean images are not mistakenly routed through adversary experts—thus preserving the original performance. 

Finally, introducing the adversary experts leads to a substantial improvement in adversarial image localization, with the F1 increasing from 0.246 to 0.660. As shown in Fig.~\ref{fig:ablation_feature}, the feature maps produced with adversary experts (bottom row) exhibit clearer semantic structures and stronger activation over forged regions, effectively suppressing irrelevant background noise. In contrast, the feature maps without adversary experts (middle row) appear disorganized and lack distinct foreground-background separation, indicating that adversarial perturbations significantly distort the original feature space. These results highlight the role of adversary experts in restoring semantic consistency and guiding accurate localization under attack conditions.

These results validate the complementary roles of each module: forgery experts enhance basic IFDL performance, the adversary detector ensures reliable adversary identification, and adversary experts effectively correct feature shifts caused by adversarial noise.

\subsection{Robustness under Common Distortions}
In addition to adversarial attacks, attackers in real-world scenarios may resort to model-agnostic post-processing operations—such as compression, blurring, or resizing—to weaken or conceal forgery traces without relying on any specific model structure. It is essential to evaluate the IFDL model's robustness under varying degrees of image degradation. Therefore, we apply a range of common post-processing operations to the testing datasets, including JPEG compression, Gaussian noise, Gaussian blurring, and resizing operations. Specifically, we apply JPEG compression with quality factors ranging from 90 to 50, Gaussian noise with standard deviations from 5 to 13, Gaussian blur with kernel sizes from 5 to 13, and image resizing with scaling rates from 0.9 to 0.5. These settings simulate increasingly severe real-world distortions.

\begin{figure}[t]
	\centering
	\begin{tabular}{c}
		\hspace{-1.0cm}
		\begin{overpic}[width=0.5\textwidth,keepaspectratio]{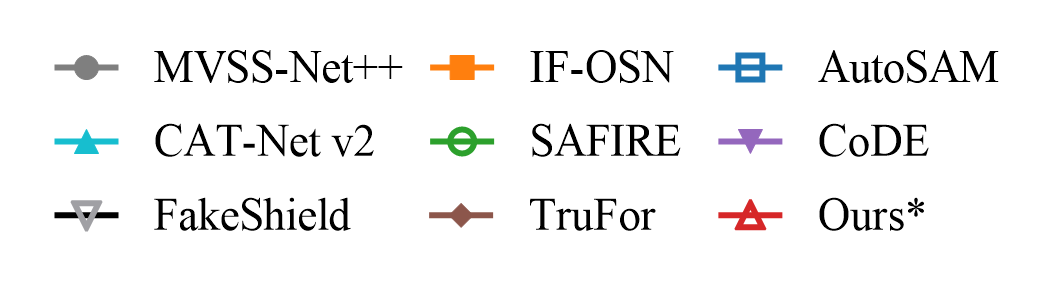}
		\end{overpic}\\\\
		\begin{tabular}{c c c c}
			\hspace{-0.9cm}
			\begin{overpic}[width=0.3\textwidth,keepaspectratio]{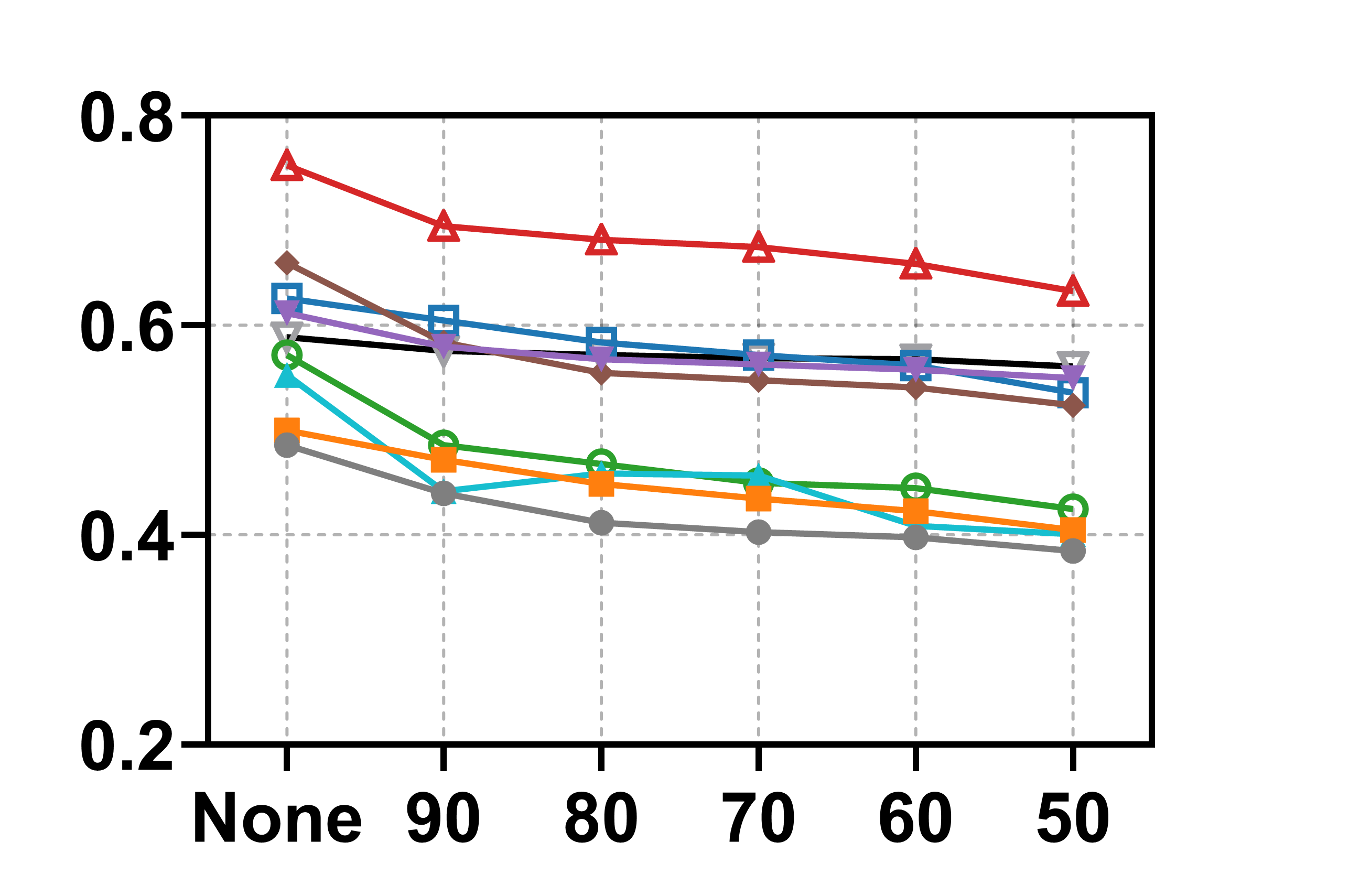}
				\put(22, 62){JPEG Compression}
				\put(5, 65){F1}
				\put(30, -6){Quality Factor}
			\end{overpic} &
			\hspace{-1.3cm}
			\begin{overpic}[width=0.3\textwidth,keepaspectratio]{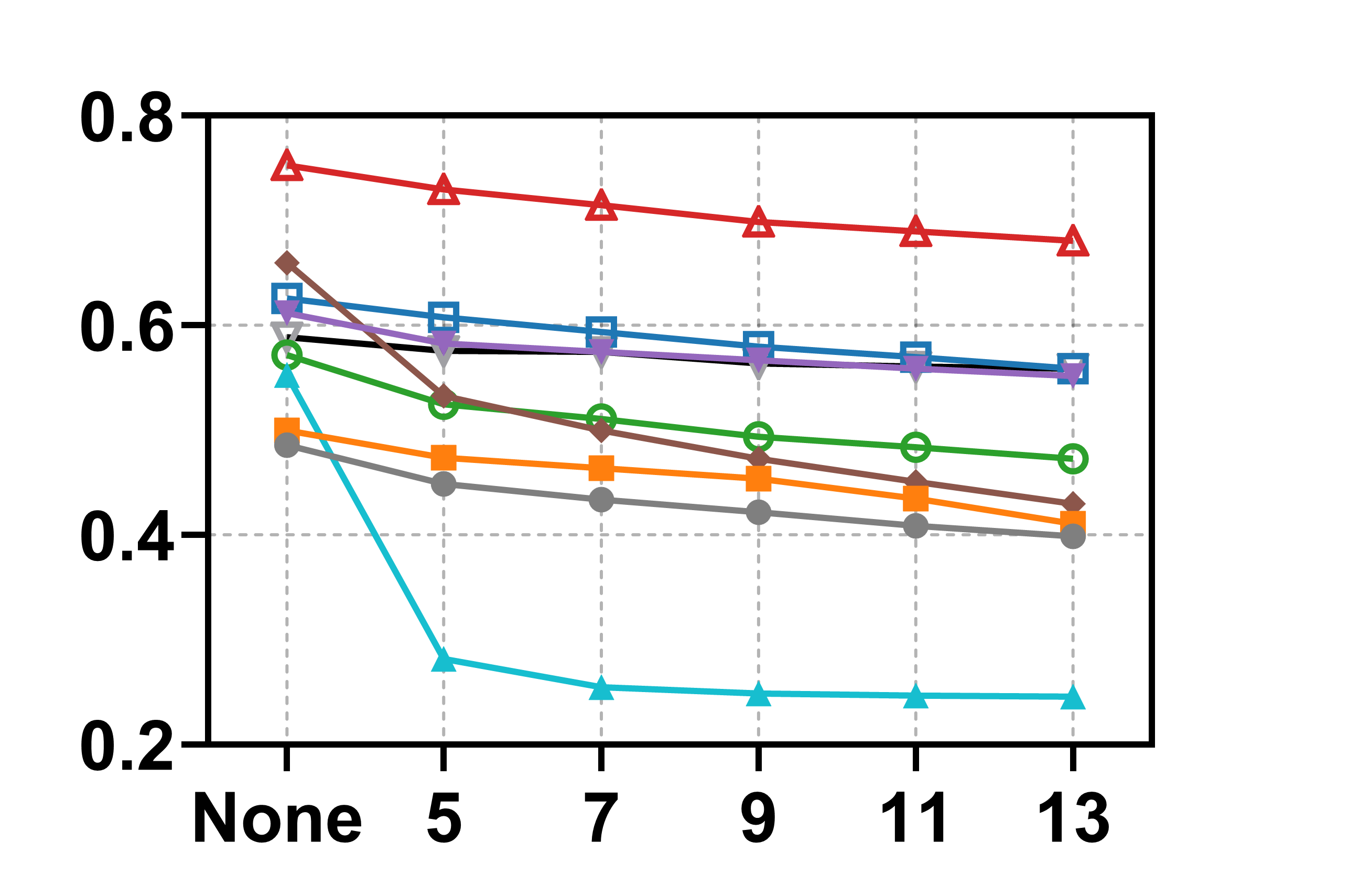}
				\put(28, 62){Gaussian Noise}
				\put(5, 65){F1}
				\put(28, -6){Noise Deviation}
			\end{overpic}
		\end{tabular}
		\\\\\\
		\begin{tabular}{c c c c}
			\hspace{-0.9cm}
			\begin{overpic}[width=0.3\textwidth,keepaspectratio]{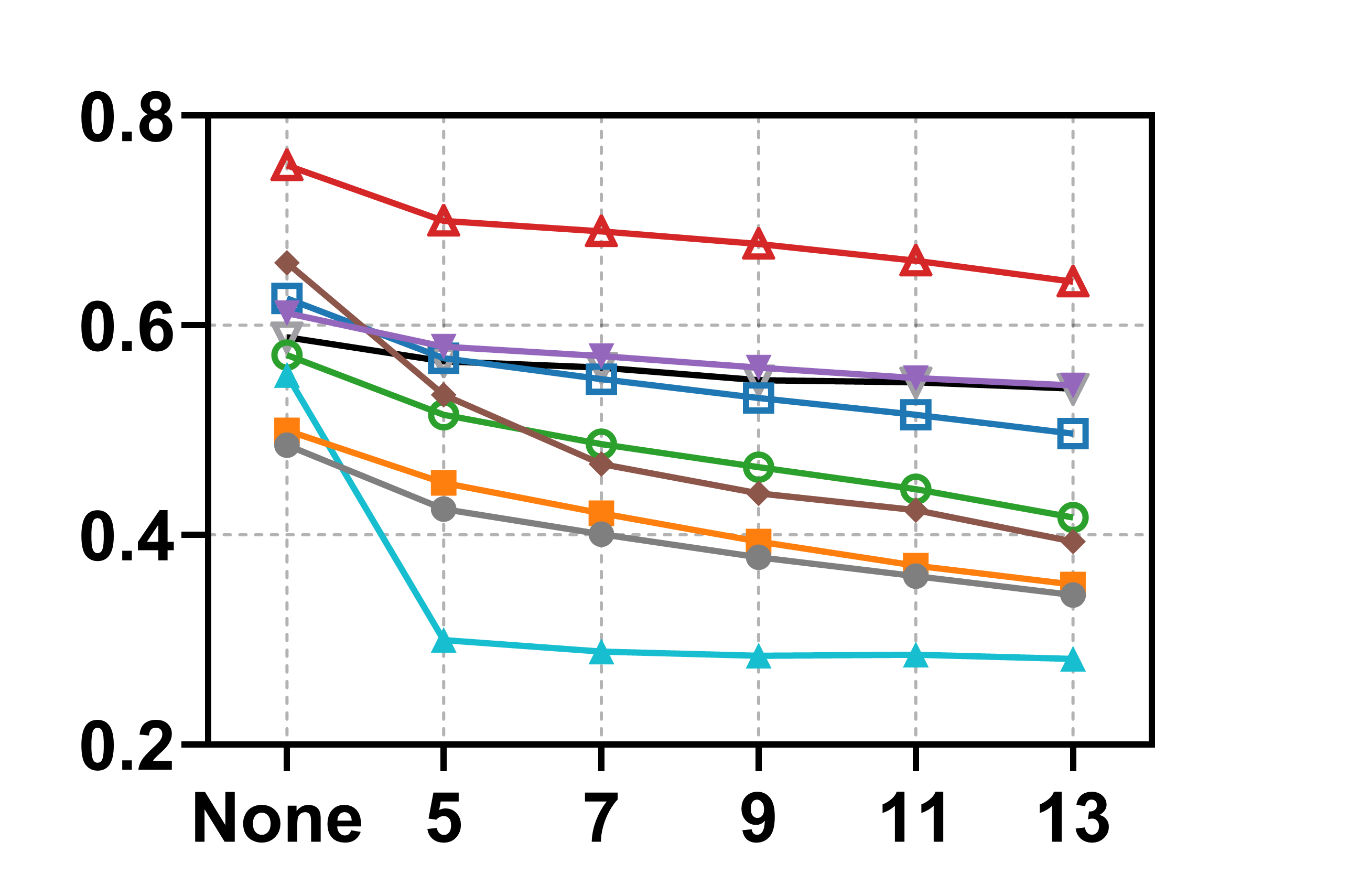}
				\put(30, 62){Gaussian Blur}
				\put(5, 65){F1}
				\put(35, -6){Kernel Size}
			\end{overpic} &
			\hspace{-1.3cm}
			\begin{overpic}[width=0.3\textwidth,keepaspectratio]{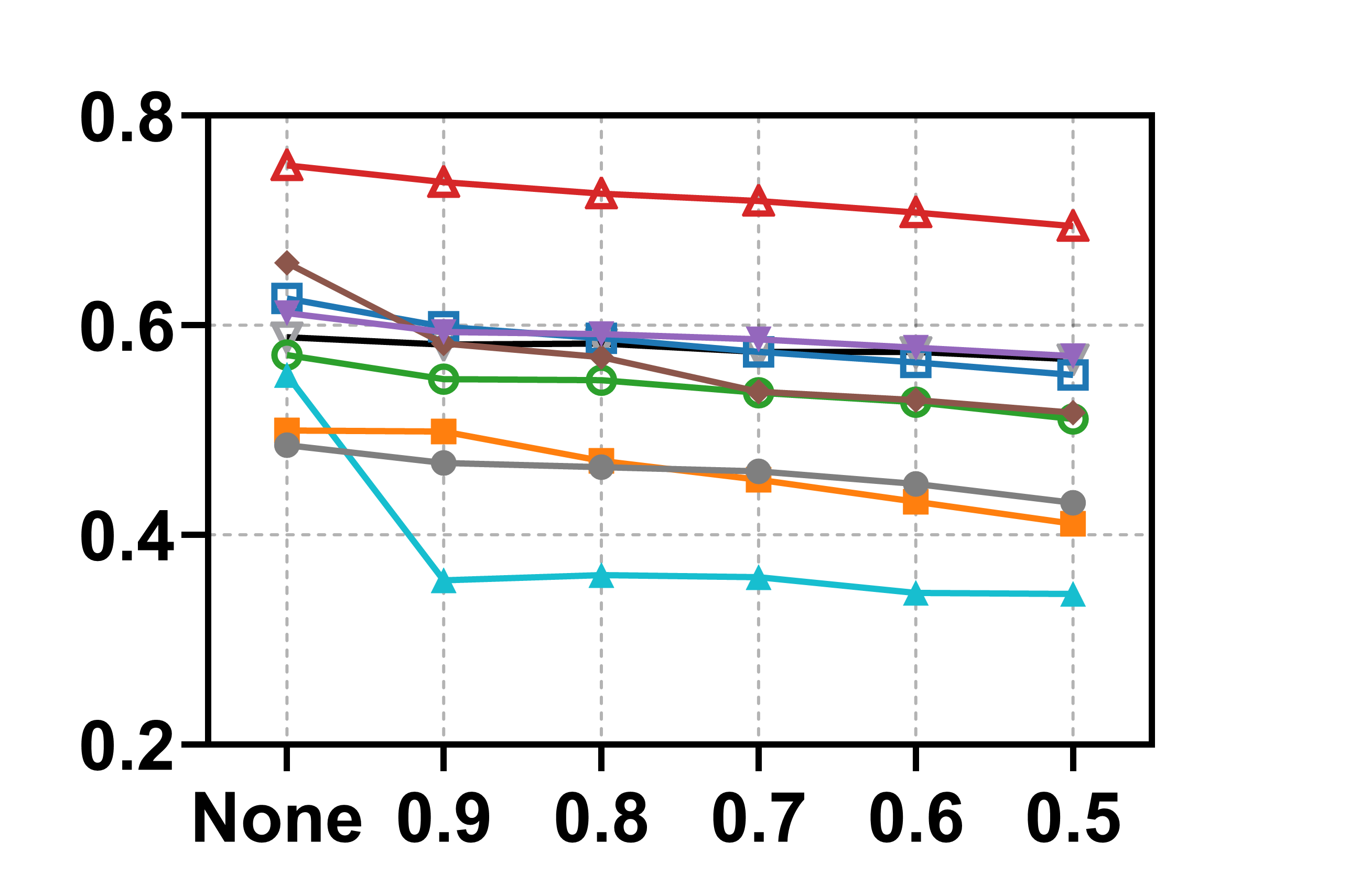}
				\put(40, 62){Resizing}
				\put(5, 65){F1}
				\put(31, -6){Resizing Rate}
			\end{overpic}
		\end{tabular}
		\\\\
	\end{tabular}
	\caption{Robustness comparison under JPEG compression, Gaussian noise, Gaussian blur, and resizing across all test sets.}
	\label{fig:robustness_all_testing_set}
\end{figure}
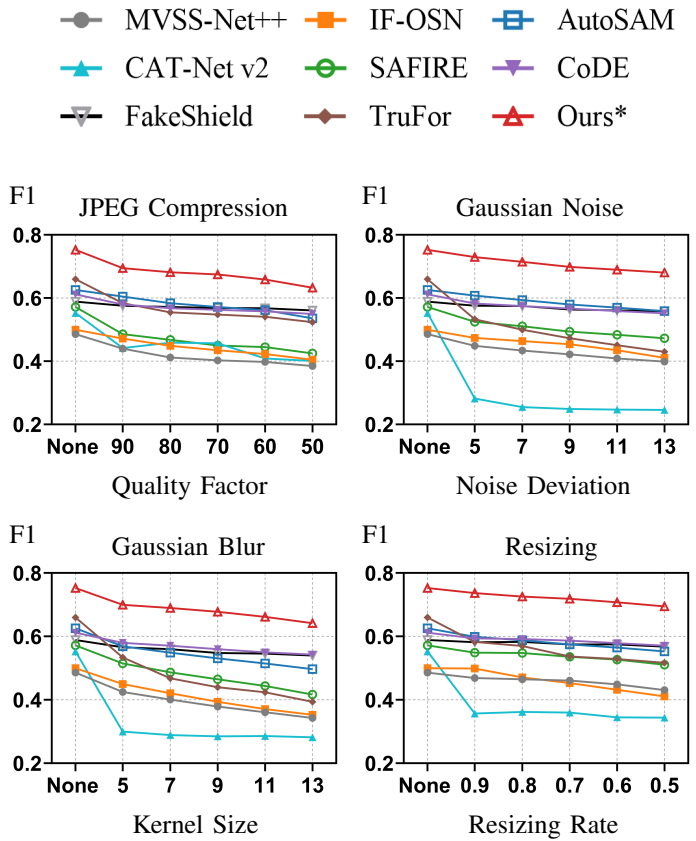

As shown in Fig.~\ref{fig:robustness_all_testing_set}, our method consistently outperforms existing methods across all types of operations and severity levels. This advantage is particularly evident under stronger degradation conditions, such as low-quality JPEG compression or aggressive Gaussian blurring, where many baseline methods tend to suffer significant performance drops. This consistent advantage underscores the effectiveness of our design in preserving critical discriminative cues under challenging conditions.

\subsection{Distortion-based Defense vs. Adversary Experts}
Prior works~\cite{dziugaite2016study, xie2017adversarial} have attempted to mitigate adversarial noise by applying JPEG compression or resizing as image pre-processing operations. In this section, we compare the defense effectiveness of our proposed adversary experts with that of common distortion operations, including JPEG compression (JC), Gaussian noise (GN), Gaussian blur (GB), and resizing (Re). We select the BSR algorithm, which achieves the strongest attack performance, as the baseline. For our method, we evaluate both settings: with and without adversarial experts (Adv. Exp.).

As shown in Table~\ref{tab:defense_comparison}, among the four distortion operations, GB achieves the best defense performance. The other three operations result in only marginal recovery. This may be attributed to the fact that Gaussian blur effectively suppresses high-frequency adversarial noise by smoothing local pixel variations. However, such operations are inherently task-agnostic and are applied indiscriminately to both clean and adversarial images, which inevitably leads to a degradation of the IFDL model's performance on clean images.

In contrast, the proposed adversary experts are selectively activated according to the adversary detector's judgment, ensuring they remain inactive for clean images and thus preserve original performance. Our method, when equipped with adversary experts, achieves significantly better defense performance on adversarial images compared to the four distortion operations applied to other IFDL models.

\begin{table}[t]
	\centering
	\caption{Defense effectiveness under common distortions versus our adversary experts. JC, GN, GB, and Re denote JPEG compression, Gaussian noise, Gaussian blur, and resizing, respectively.}
	\tabcolsep=0.05cm
	\newcommand{\xmark}{\ding{55}}
	\begin{tabular}{c c c | c c c c}
		\hline
		\multicolumn{3}{c|}{Defense} Strategy & AutoSAM & SAFIRE & FakeShield & $\textbf{Ours}^{\star}$\\
		\hline
		\multirow{6}{*}{\rotatebox{90}{BSR}} &
		\multirow{6}{*}{\rotatebox{90}{($\varphi=8$)}} 
		& w/o Defense & 0.256 & 0.353 & 0.284 & 0.259\\
		& & JC (QF=75) & 0.334 & 0.364 & 0.379 & 0.384 \\
		& & GN (ND=9) & 0.296 & 0.381 & 0.370 & 0.354 \\
		& & GB (KS=9) & 0.465 & 0.344 & 0.448 & 0.424 \\
		& & Re (RR=0.7) & 0.336 & 0.362 & 0.314 & 0.330\\
		\rowcolor{gray!20} & & Adv. Exp. & - & - & - & \textbf{0.653}\\
		\hline
		\multirow{6}{*}{\rotatebox{90}{BSR}} &
		\multirow{6}{*}{\rotatebox{90}{($\varphi=12$)}} 
		& w/o Defense & 0.244 & 0.316 & 0.262 & 0.247 \\
		& & JC (QF=75) & 0.275 & 0.329 & 0.317 & 0.311 \\
		& & GN (ND=9) & 0.255 & 0.346 & 0.287 & 0.280 \\
		& & GB (KS=9) & 0.413 & 0.323 & 0.427 & 0.380 \\
		& & Re (RR=0.7) & 0.277 & 0.332 & 0.272 & 0.278 \\
		\rowcolor{gray!20} & & Adv. Exp. & - & - & - & \textbf{0.632}\\
		\hline    
	\end{tabular}
	\label{tab:defense_comparison}
\end{table}

\section{Conclusion}
\label{sect:conclusion}
In this paper, we introduce an unified framework that can not only perform image-level detection of real, forged and adversarial images, but also achieve pixel-level forgery localization considering potential adversarial attacks. Our proposed framework, ForensicsSAM, integrates shared forgery experts, an adversary detector, and adaptive adversary experts into the SAM backbone. These components are jointly optimized through a three-stage training pipeline—clean supervision for IFDL task, adversarial detection, and forgery-relevant feature correction—designed to address the challenge of transferable adversarial attacks targeting the PERT-based SAM variants. This design ensures adversarial robustness without sacrificing performance on clean images. Extensive experimental results demonstrate that our method achieves the state-of-the-art IFDL performance with built-in adversarial robustness. 

Beyond the IFDL task, our proposed framework and training paradigm can be easily extended to other downstream tasks to enable unified adversarial robustness. We believe our work provides valuable insights for designing generalizable and robust systems in the era of large vision foundation models.

\bibliographystyle{IEEEtran}
\bibliography{references}

\end{document}